\documentclass{article}

\usepackage[nonatbib]{neurips_data_2024}

\usepackage[utf8]{inputenc} 
\usepackage[T1]{fontenc}    
\usepackage[colorlinks,
            linkcolor=red,
            anchorcolor=blue,
            citecolor=blue]{hyperref} 
\usepackage{url}            
\usepackage{booktabs}       
\usepackage{amsfonts}       
\usepackage{nicefrac}       
\usepackage{microtype}      
\usepackage[table,xcdraw]{xcolor}
\usepackage{adjustbox}
\usepackage{subfigure}
\usepackage{multirow,multicol}
\usepackage{wrapfig}
\usepackage{cite}

\newcommand{\CLA}[1]{{\color[HTML]{E76254} \textbf{#1}}}

\newcommand{\CLB}[1]{{\color[HTML]{4472c4} \textbf{#1}}}

\title{CMC-Bench: Towards a New Paradigm of Visual Signal Compression}

\author{ \textbf{Chunyi Li}\, $^{1}$ \hspace{9pt} \textbf{Xiele Wu}$^{1}$ \hspace{9pt} \textbf{Haoning Wu}$^{2}$ \hspace{9pt} \textbf{Donghui Feng}$^{1}$ \hspace {9pt} \textbf{Zicheng Zhang}$^{1}$ \hspace {9pt} \\ \textbf{Guo Lu}$^{1}$ \hspace{9pt} \ \textbf{Xiongkuo Min}$^{1}$ \hspace{9pt} \textbf{Xiaohong Liu}$^{1*}$ \hspace{9pt} \textbf{Guangtao Zhai}$^{1*}$ \hspace{9pt} \textbf{Weisi Lin}$^{2}$ \vspace{2pt} \\
$^1$ Shanghai Jiao Tong University \hspace{9pt} $^2$ Nanyang Technological University \vspace{2pt}\\
$^\ast$Corresponding Authors. Project Page: \href{https://github.com/Q-Future/CMC-Bench}{\textit{https://github.com/Q-Future/CMC-Bench}}
}

\begin{document}

\maketitle

\begin{figure}[tbph]
    \centering
    \vspace{-15pt}
    \includegraphics[width=0.95\linewidth]{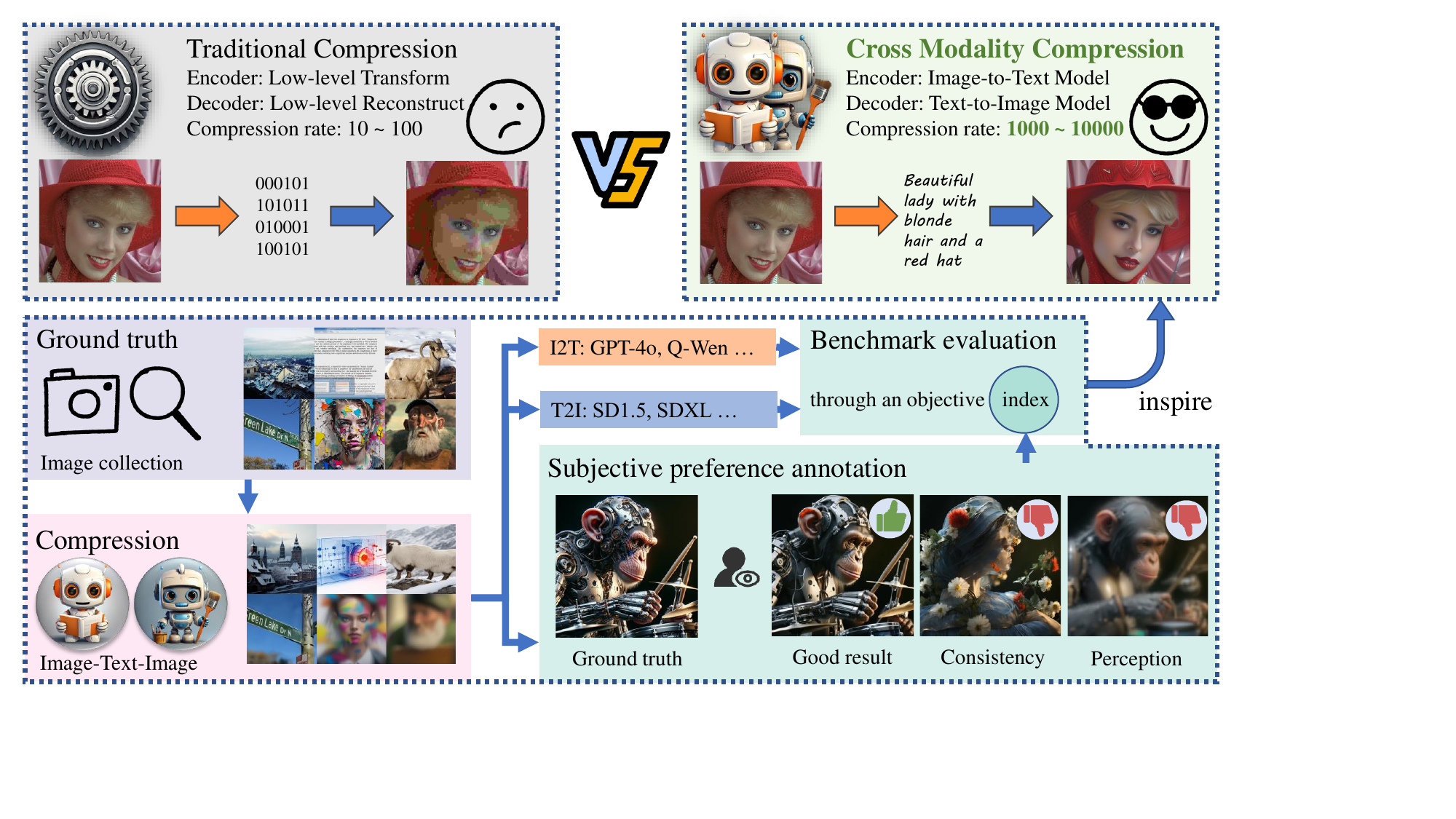}
    \vspace{-6pt}
    \caption{Overview of CMC-Bench. We demonstrate the superiority of Cross Modality Compression over traditional codecs, and subjective and objective evaluations of compression results on Consistency and Perception. This benchmark can motivate it to become the future codec paradigm.}
    \label{fig:spotlight}
\end{figure}

\begin{abstract}
  Ultra-low bitrate image compression is a challenging and demanding topic. With the development of Large Multimodal Models (LMMs), a Cross Modality Compression (CMC) paradigm of Image-Text-Image has emerged. Compared with traditional codecs, this semantic-level compression can reduce image data size to 0.1\% or even lower, which has strong potential applications. However, CMC has certain defects in consistency with the original image and perceptual quality. To address this problem, we introduce CMC-Bench, a benchmark of the cooperative performance of Image-to-Text (I2T) and Text-to-Image (T2I) models for image compression. This benchmark covers 18,000 and 40,000 images respectively to verify 6 mainstream I2T and 12 T2I models, including 160,000 subjective preference scores annotated by human experts. At ultra-low bitrates, this paper proves that the combination of some I2T and T2I models has surpassed the most advanced visual signal codecs; meanwhile, it highlights where LMMs can be further optimized toward the compression task. We encourage LMM developers to participate in this test to promote the evolution of visual signal codec protocols.
\end{abstract}

\section{Introduction}

Visual signal compression aims to minimize image data, playing a crucial role in delivering high-quality image/video services with limited network resources and storage capacity. Since the MPEG-1 \cite{metric:tra-mpeg} standard was introduced, compression rates for visual signals have doubled ~\cite{metric:tra-jpeg,metric:tra-avc,metric:tra-hevc,metric:tra-vvc} every decade. In recent years, traditional image codecs have achieved a 500 times compression rate while ensuring a decent visual experience for humans. However, traditional codecs are approaching the Shannon limit of 1,000 times Compression Rate (CR) in the upcoming next-generation protocols. Fortunately, the rapid development of Large Multimodal Models (LMMs) has opened up possibilities for such Ultra-low-Bitrate (ULB) compression.

\textbf{Why use LMMs for compression?} LMMs support the conversion between multiple modalities, where text consumes much less space than image modalities. By cascading Image-to-Text (I2T) and Text-to-Image (T2I) models, images can be compressed and reconstructed from semantic information. This Cross-Modality Compression (CMC) paradigm operates at the semantic level, which outperforms traditional codecs at the pixel level. It enables easy attainment of ULB, and even the Extreme-low Bitrate (ELB) for CR about 10,000 times.

However, at such low bitrates, CMC presents two significant issues that cannot be overlooked. (1) \textbf{Consistency}: The reconstruction process heavily relies on intermediate textual information. Any omission by the I2T model (encoder) or misunderstanding by the T2I model (decoder) can result in severe distortion. Unlike minor changes in brightness and color, this can lead to a semantic-level inversion of the entire image. (2) \textbf{Perception}: Textual encoding provides a coarse representation of the image, necessitating the T2I model to add details. Insufficient details degrade perceptual quality, while excessive ones compromise faithfulness to the original image. Unfortunately, as the bitrate decreases, the conflict between them ~\cite{intro:tradeoff-1,intro:tradeoff-2} becomes more pronounced. As the consistency and perception failure cases in Figure \ref{fig:spotlight}, these issues jointly limit the application of CMC.

For LMMs, there is a lack of effective evaluation criteria both in terms of consistency and perceptual aspects. Although numerous benchmarks have recently emerged for LMMs, they are primarily designed to assess the performance of either I2T (Image-to-Text) or T2I (Text-to-Image) models working alone, such as captioning/visual question answering for I2T ~\cite{relate:bench-i2t-mm,relate:bench-i2t-seed}, or generation quality/realism for T2I  ~\cite{relate:bench-t2i-hrs,relate:bench-t2i-comp}.
Consequently, we introduce the first joint benchmark called CMC-Bench, aimed at testing the collaborative capabilities of I2T and T2I models. Our contributions include:

\begin{itemize}
    \item A large-scale dataset consists of 58,000 images using the CMC paradigm. 4,000 images among them have 160,000 expert annotations, covering both consistency and perception issues, paving the way for information loss modeling in the I2T and T2I processes.
    \item A comprehensive evaluation standards, consisting of four compression modes under different requirements, along with the two dimensions mentioned above. We validate mainstream models (including 6 I2T and 12 T2I) to explore optimal combinations.
    \item A throughout comparison with traditional codecs. We compared the benchmark winner with existing image codecs, revealing the significant advantages of the CMC image compression paradigm and some remaining drawbacks. We encourage LMM developers (both I2T and T2I) to participate in CMC-Bench to further expand the application of CMC.
\end{itemize}

\section{Related Works}

\textbf{Cross-Modality Compression.} The earliest CMC method \cite{relate:cmc-cmc} emerged in 2021, achieving a compression ratio of almost 10,000 times through text modality. However, as a simple combination of I2T and T2I models, their results often exhibit noticeable differences from the original images. Subsequently, Text+Sketch \cite{relate:cmc-text} employed edge operators and ControlNet \cite{add:controlnet} to refine CMC, but its consistency remained inferior to traditional codecs. The most advanced CMC methods, like M-CMC and MISC ~\cite{relate:cmc-rcmc,relate:cmc-mcmc,relate:cmc-vqgan,relate:cmc-uigc,relate:cmc-onceforall,relate:cmc-misc}, have surpassed advanced codecs like VVC \cite{metric:tra-vvc} in both consistency and perception, indicating the promising future of this paradigm. Nevertheless, there is still room for improvement in these two aspects. All existing CMC methods are from one specific I2T and one T2I model, and the models used are relatively outdated. Considering the rapid development of Generative-AI, how to combine the latest models towards a better CMC becomes an unrevealed question.

\textbf{Benchmark for LMM Evaluation.} Existing benchmarks are mainly designed for T2I and I2T models. For I2T, they usually take a specific image sequence as input, compare the text output by LMM with the ground truth, and use the relevance of the two as a performance indicator. The annotation content includes common sense ~\cite{relate:bench-i2t-mm,relate:bench-i2t-seed} or specific expert fields ~\cite{relate:bench-i2t-qbench,relate:bench-i2t-fakebench,add:a-bench,add:q-instruct}. For T2I, the input is a carefully designed text prompt ~\cite{relate:bench-t2i-hrs,relate:bench-t2i-comp} (e.g. different themes, adjectives, and spatial relationship). They use specific visual encoders to process the output image of LMM and determine its alignment with the text as the generative performance ~\cite{relate:bench-t2i-draw,relate:bench-t2i-dalleval}.  However, as the current CMC paradigm is still immature, there is no pipeline for the joint evaluation of I2T+T2I model.

\begin{table*}[t]
\centering
\caption{Comparison of image compression related dataset with subjective labeling.}
\adjustbox{max width=1.0\textwidth}{
\begin{tabular}{l|ccccccc}
\toprule
Dataset     & Ref & Dis & Ratings & Score & Resolution     & Image type   & Dimension               \\ \midrule
CLIC2021 \cite{dataset:clic}    & 585          & 2,730           & 122,107           & DS       & 768      & NSI          & Consistency             \\
\rowcolor[HTML]{D0D0D0} 
CLIC2022 \cite{dataset:clic}       & 585           & 2,730            & 57,300            & DS       & 768      & NSI          & Consistency             \\ 
NTIRE2022 \cite{relate:dataset-ntire2022}       & 250           & 29,150            & 1,880,000            & MOS       & 288      & NSI          & Consistency, Perception             \\
\rowcolor[HTML]{D0D0D0} 
SCID \cite{dataset:scid}       & 40           & 1,800             & 18,000            & MOS      & 1,280     & SCI          & Consistency             \\
CCT \cite{dataset:cct}       & 72           & 1,320             & 26,400            & MOS      & 1,280     & NSI, SCI          & Consistency             \\
\rowcolor[HTML]{D0D0D0} 
AGIQA-3K \cite{dataset:agiqa-3k}   & -            & 2,982             & 62,622            & MOS      & 512      & AIGI         & Perception              \\
ImageReward \cite{relate:dataset-imagereward} & -            & 136,892           & 136,892           & SS       & 512$\sim$1,024 & AIGI         & Perception              \\ \midrule
\rowcolor[HTML]{CCCCCC} 
\textbf{CMC-Bench}    & \textbf{1,000}         & \textbf{58,000}           & \textbf{160,000}           & \textbf{MOS}      & \textbf{512$\sim$1,024} & \textbf{NSI, SCI, AIGI} & \textbf{Consistency, Perception} \\ \bottomrule
\end{tabular}
}
\vspace{-5mm}
\label{tab:relate}
\end{table*}

\textbf{Benchmark for Image Compression.} Given the significance of visual information compression, several related competitions \cite{dataset:clic,relate:dataset-ntire2022} have been held in recent years. However, these competitions often limit their scope to Natural Scene Images (NSIs). Screen Content Images (SCIs) ~\cite{dataset:scid,dataset:cct}, which are prevalent on the internet, and the emerging AI-Generated Images (AIGIs) ~\cite{dataset:agiqa-3k,relate:dataset-imagereward,dataset:agiqa1k} have received some attention with new datasets, but no existing dataset comprehensively considers them together. Moreover, the performance evaluation of compression algorithms can be challenging, often requiring subjective quality assessments from human viewers to train Image Quality Assessment (IQA) ~\cite{add:aspect-qoe,add:cartoon,add:rr,add:6G,add:advancing,add:gms,add:iscas,add:lmmpcqa} models, which provide objective metrics for compression algorithms. In the context of ULB image compression, both the \textbf{consistency} between the distorted and reference images, as well as the inherent appeal of the distorted image in human \textbf{perception}, need to be annotated. Existing IQA datasets typically annotate only one aspect, while often in a coarse-grained manner through Single Stimulus (SS) or Double Stimulus (DS) comparisons. In contrast, Mean Opinion Score (MOS) derived from multiple subjects offers a more detailed and objective evaluation as shown in Table \ref{tab:relate}.

\section{Dataset Construction}
\subsection{Ground Truth Selection}
\begin{figure*}[t]
    \centering
    \includegraphics[width=0.98\textwidth]{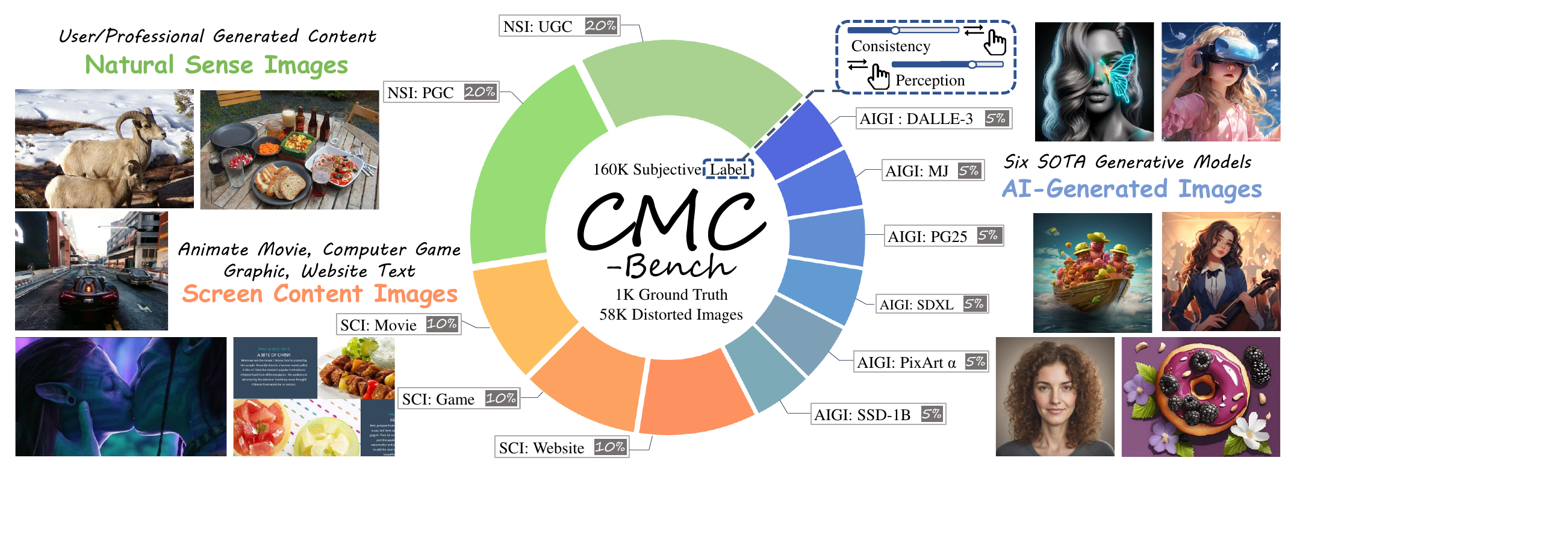}
    \caption{Source data illustration of CMC-Bench from three content types.}
    \label{fig:dataset}
    \vspace{-5mm}
\end{figure*}

To provide a comprehensive and high-quality resource for various applications on the Internet, we carefully curated 1,000 high-quality images without compression distortion as the ground truth of CMC-Bench. Among them, NSIs are the most mainstream content, so we selected 400 images. At the same time, considering that SCIs are more common on screens and AIGIs are increasing on the Internet in the upcoming LMM era, we selected 300 images from each of these two categories. The specific content is as shown in Figure \ref{fig:dataset}.

\textbf{NSI}: A collection of 200 high-quality Professional Generated Content (PGC) released by TV stations and photographers, specifically sampled from the CLIC database \cite{dataset:clic}; and 200 User Generated Content (UGC) by average users, selected from MS-COCO \cite{dataset:mscoco}. To ensure image quality, we employed Q-Align \cite{perception:q-align} to filter out low-quality UGC that might be overexposed.

\textbf{SCI}: Consisting 100 computer graphics from CGIQA-6K \cite{dataset:cgiqa} in animated movies; 100 game renders from CCT and CGIQA-6K ~\cite{dataset:cct,dataset:cgiqa}; and 100 webpages with both images and text from CCT, SCID, and Webpage Saliency datasets ~\cite{dataset:cct,dataset:scid,dataset:webpage}. To maintain frame clarity, we also applied Q-Align \cite{perception:q-align} to address factors like motion blur that affect visual quality.

\textbf{AIGI}: Comprises 50 images each, generated by 6 latest models: DALLE3, MidJourney, PG v25, PixArt $\alpha$, SDXL, and SSD-1B ~\cite{gen:dalle,gen:MJ,gen:Playground25,gen:pixart,gen:xl,gen:ssd-1b}. They have demonstrated exceptional preference in previous subjective ratings ~\cite{database:aigiqa20k,add:ntire2024,add:track2,add:g-refine,add:q-refine}, representing the pinnacle of AIGI capabilities.

\subsection{Compression Mode}

\begin{figure*}[t]
    \centering
    \includegraphics[width=0.98\textwidth]{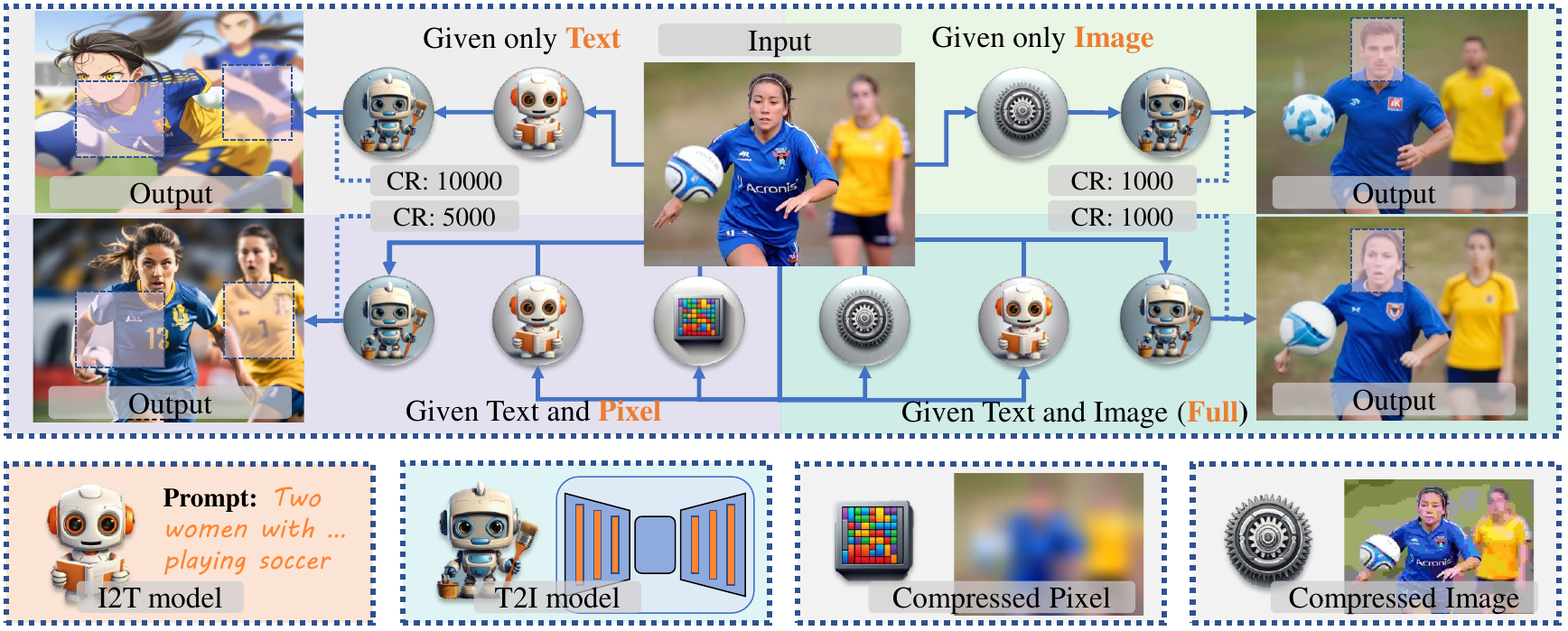}
    \caption{Illustration of 4 working modes of CMC. \textit{Text} mode roughly reconstructs the semantic information, \textit{Pixel} mode slightly improves low-level consistency, \textit{Image} mode provides a similar structure towards ground truth but a different character, and \textit{Full} mode has the best performance.}
    \label{fig:mode}
    \vspace{-5mm}
\end{figure*}

Drawing on previous work in CMC, we categorize CMC into four working modes, as shown in Figure \ref{fig:mode}. Each type employs distinct configurations and is suitable for different scenarios:

\textbf{\textit{Text}}: The I2T model converts images to text and is directly restored by the T2I model. Due to its reliance on the text modality only, this approach achieves a CR of 10,000, ideal for ELB situations.

\textbf{\textit{Pixel}}: Each $64 \times 64$ blocks from ground truth are merged and quantized into one pixel. Beyond the \textit{Text} mode, these pixels initialize the T2I process. The pixel representation is relatively compact, offering a CR of around 5,000, suitable for less rigorous ELB but higher demands on consistency.

\textbf{\textit{Image}}: Traditional codecs are employed to compress the image, which serves as input for the T2I model for enhancement. Unlike the previous two, it omits the time-consuming I2T process by leaving the text input of the T2I model empty. This approach can achieve a CR of 1,000, suitable for ULB bandwidth but with high real-time requirements.

\textbf{\textit{Full}}: Extending the \textit{Image} mode, the T2I is guided by text content, encompassing the full pipeline of I2T, traditional codec, and T2I. It also has a CR of approximately 1,000, suitable for the most demanding performance scenarios.

\subsection{Benchmark Candidates}
\label{sec:setting}

We employ 6 I2T and 12 T2I models across four compression modes. Due to the absence of text, the I2T model is not used in the \textit{Image} mode; while for T2I, among the 4 Image Reconstruction (IR) models requiring an initial image and are not compatible with \textit{Text} and \textit{Pixel} modes. The remaining 8 T2I generative models support all modes. We use one certain T2I, and validate all possible I2T models to verify their performance separately (vice versa for T2I validation). 
For a fair comparison, 
We fixed RealVis \cite{gen:RealVis} to minimize the T2I process distortion, which ensures the performance fluctuation mainly comes from the I2T.
Similarly, we fix I2T as GPT-4o \cite{i2t:gpt4} when validating T2I models.
Each I2T model produces 3,000 images, while restorative and generative models for T2I have 2,000 and 4,000, respectively. A total of 18,000 + 40,000 = 58,000 images are generated.

I2T model: GPT-4o \cite{i2t:gpt4}, LLAVA-1.5 \cite{i2t:llava}, MPlugOwl-2 \cite{i2t:mplugowl}, Qwen \cite{i2t:Qwen-VL}, ShareGPT \cite{i2t:sharegpt4v}, and InstructBLIP \cite{i2t:iblip}. Except for one model \cite{i2t:iblip} for image captioning with default token length, we modify the output length of others to 10$\sim$20 words for a balance between bitrate and performance.

T2I model: Animate \cite{gen:animatediff}, Dreamlike \cite{gen:dream}, PG20 \cite{gen:Playground20}, PG25 \cite{gen:Playground25}, RealVis \cite{gen:RealVis}, SD15 \cite{gen:sd}, SDXL \cite{gen:xl}, and SSD-1B \cite{gen:ssd-1b} as generative model; DiffBIR \cite{ir:diffbir}, InstructPix \cite{ir:instructpix2pix}, PASD \cite{ir:pasd}, and StableSR \cite{ir:stablesr} as IR model. A higher denoising strength indicates a more obvious modification on the starting point. To balance the consistency and perception indicators, we set the strength of \textit{Full} and \textit{Image} modes as 0.5, the \textit{Pixel} mode as 0.8, and the \textit{Text} mode as the default 1.

Traditional codec: For \textit{Full} and \textit{Image} mode, we utilize the most advanced traditional codec VVC \cite{metric:tra-vvc} to provide a reference image. 
Towards 1,000 times compression, we take its nearest bitrate that meets the ULB requirement, where the Quantizer Parameter (QP) is 53.

\subsection{Human Preference Annotation}
\label{sec:subject}

Referring from previous large-scale subjective annotation \cite{dataset:clic} methods, we do not perform coarse-grained labeling on the entire dataset considering the limitation on annotator numbers. Instead, we fine-grain the annotations on 4,000 images to ensure multiple ratings for each image. Note that, as the benchmark indicator should be adjusted on subjective data, we did not directly select subsets from the 58,000 test images. Instead, we generated new images to prevent prior exposure to the content being evaluated.
Given the greater impact of T2I models on CMC tasks than I2T models, we follow the T2I paradigm described in Section \ref{sec:setting}. The I2T model is fixed as GPT-4o \cite{i2t:gpt4} and combined with 12 different T2I models, compressing 100 ground truth into 4,000 distorted images. To ensure quality diversity, we randomly assigned strength from 0.2 to 0.9 rather than a fixed value. Each distorted image is paired with its corresponding ground truth and shown to 20 experienced participants who provided ratings on consistency and perception dimensions.
Each image is then summarized into two overall scores from 0 to 5, combining all participants' feedback. For a detailed description of the experimental setup and data process, please refer to the supplementary.


\section{Experiment}

\subsection{Evaluation Indicator Settings}

All 6 I2T and 12 T2I LMMs are verified and tested by different parameters and fixed them towards an optimal situation according to Section \ref{sec:setting}, while the internal model weight remains zero-shot to ensure fairness in ranking. The specific parameters verification is provided in the supplementary. Taking the reference and distorted image pairs as input, We use TOPIQ \cite{quality:topiq}, the most advanced IQA metric in Full-Reference (FR) and No-Reference (NR) configuration to characterize consistency and perception. The average score of 1,000 ground truth images is reported as the final performance. Combining these two issues towards 4 working modes, the models are evaluated by 8 indicators for generative T2I, 6 indicators (exclude \textit{Image} mode) for I2T, and 4 indicators (exclude \textit{Pixel} and \textit{Text} mode) for T2I restorative models. A weighted average of $2\times$ FR indicators and $1\times$ NR indicators is given as the overall score for ranking since the TOPIQ-FR has a smaller floating range than TOPIQ-NR. As restorative models only support ULB compression in \textit{Full} and \textit{Image} mode, but not ELB compression in \textit{Pixel} and \textit{Text} mode, the overall score of the T2I model is ranked under ULB and ELB respectively.
\begin{wraptable}{r}{0.5\textwidth}
    \centering
    \vspace{-5mm}
    \caption{Correlation between objective IQA evaluation and subjective human preference.}
    \label{tab:iqa}
    \adjustbox{max width=0.5\textwidth}{
    \begin{tabular}{l|cc|l|cc}
    \toprule
    Consistency       & $\sigma \uparrow$  & $\kappa \uparrow$  & Perception       & $\sigma \uparrow$  & $\kappa \uparrow$  \\ \midrule
    AHIQ     & 0.844 & 0.645 & CLIPIQA  & 0.825 & 0.623 \\
    \rowcolor[HTML]{D0D0D0} 
    DISTS    & 0.795 & 0.599 & CNNIQA   & 0.584 & 0.414 \\
    LPIPS    & 0.583 & 0.406 & DBCNN    & 0.833 & 0.640 \\
    \rowcolor[HTML]{D0D0D0} 
    PieAPP   & 0.433 & 0.294 & HyperIQA & 0.730 & 0.534 \\
    \textbf{TOPIQ} & \textbf{0.943} & \textbf{0.792} & \textbf{TOPIQ} & \textbf{0.901} & \textbf{0.738} \\ \bottomrule
    \end{tabular}
    }
    \vspace{-3mm}
\end{wraptable}
In addition to TOPIQ, we also used four cutting-edge FR-IQA (AHIQ \cite{quality:ahiq}, DISTS \cite{quality:dists}, LPIPS \cite{quality:lpips}, PieAPP \cite{quality:pieapp}) and NR-IQA (CLIPIQA \cite{quality:CLIPIQA}, CNNIQA \cite{quality:CNNIQA}, DBCNN \cite{quality:DBCNN}, HyperIQA \cite{quality:HyperIQA}) algorithms to objectively score the distorted images in terms of consistency and perception. The higher the Spearman ($\sigma$) and Kendall Rank-order Correlation Coefficient ($\kappa$), the better correlation between the objective and subjective scores. All models are trained on 80\% of the distorted images in Section 3.4 and tested on the remaining 20\%. Experiments in Table \ref{tab:iqa} show that the correlation between the fine-tuned TOPIQ \cite{quality:topiq} and the subjective score is outstanding with $\sigma$ beyond 0.9 in both dimensions, making it appropriate performance indicators reflecting human preference for compressed images.

\begin{figure*}[t]
    \centering
    \subfigure{\includegraphics[width=0.48\textwidth]{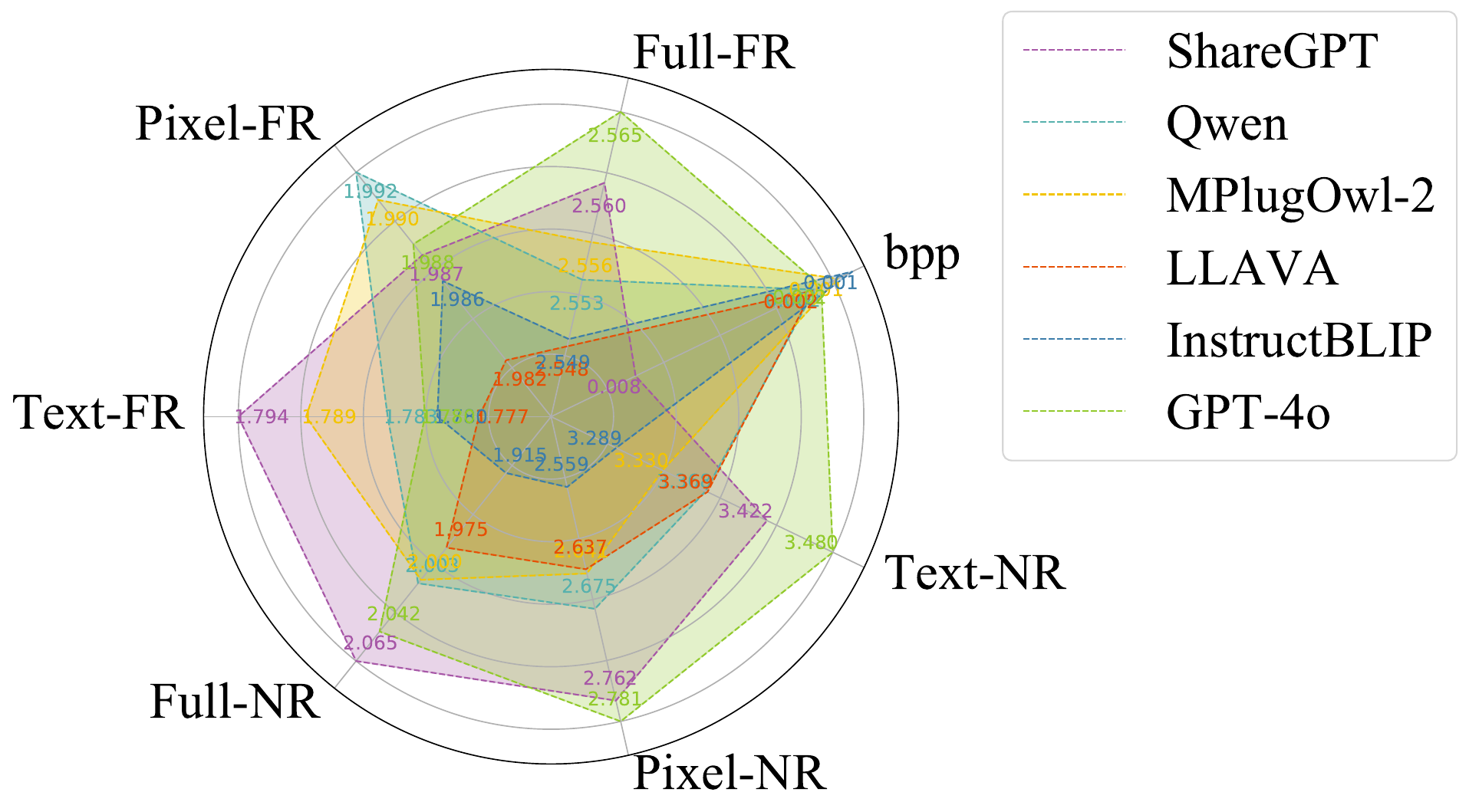}}\hskip.2em
    \subfigure{\includegraphics[width=0.48\textwidth]{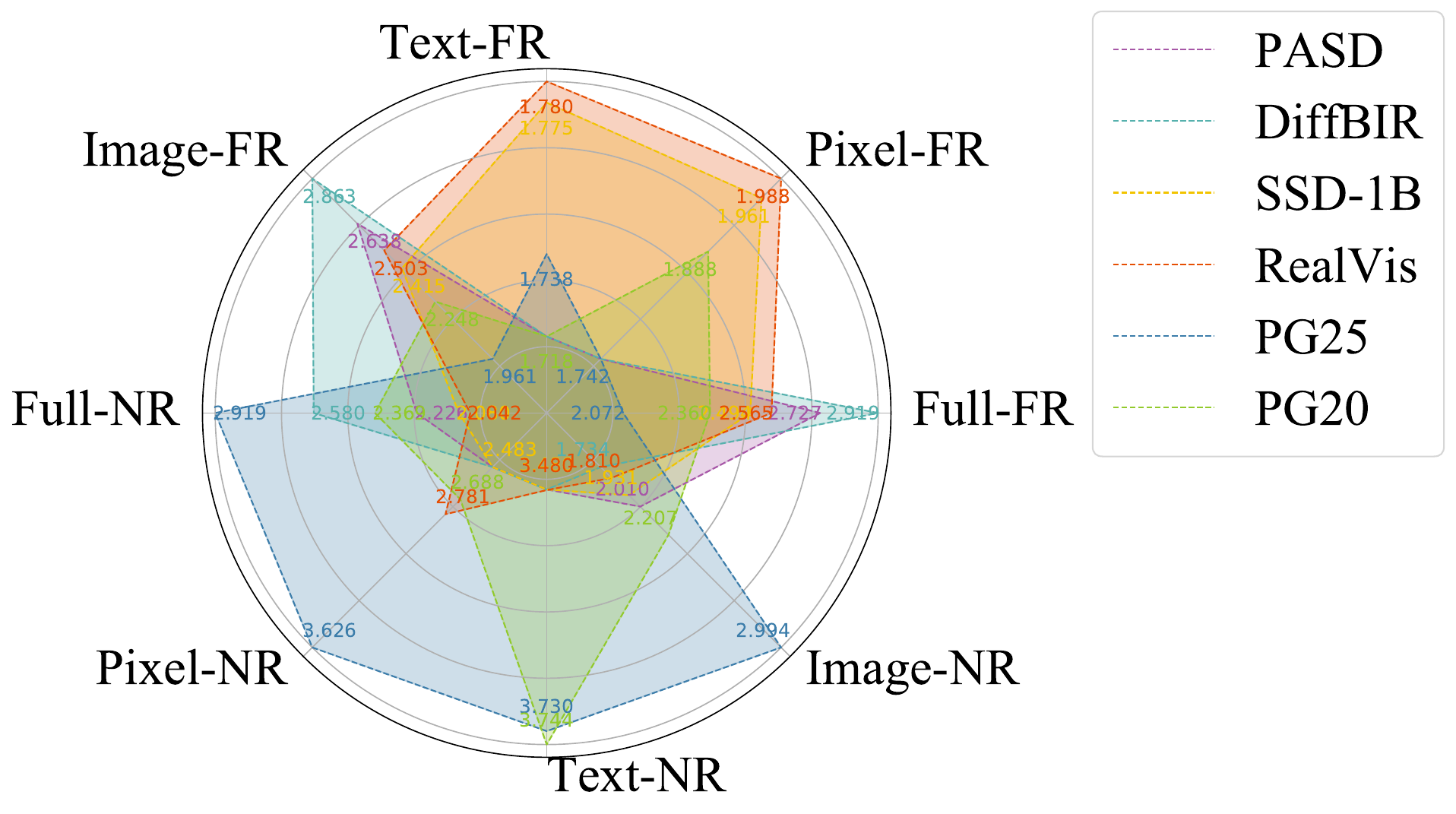}}\hskip.2em
    \vspace{-8pt}
    \caption{A radar map illustrates the collaboration of mainstream I2T (left) and T2I (right) LMMs. The model are tested as \{6 different I2Ts + RealVis \cite{gen:RealVis}\} and \{GPT-4o \cite{i2t:gpt4} + 12 different T2Is\}. }
    \label{fig:radar}
\end{figure*}

\subsection{Benchmark Result and Discussion}

Figure \ref{fig:radar} shows the performance of 6 I2T models as encoders and 12 I2T models as decoders in image compression. For I2T, considering the different lengths of intermediate text, we show the bit-per-pixel (bpp) of each model together with the performance index, where ULB and ELB correspond to 0.024 and 0.0024 bpps respectively.

\begin{table*}[t]
\centering
\vspace{-2mm}
\caption{Benchmark result in \textit{Full} and \textit{Image} modes for 8 T2I generative and 4 restorative models at ultra-low bitrate (1,000 times compression). NSI/SCI/AIGI stands for the compressed image types. FR and NR indicate consistency and perception scores. [Key: \CLA{Best}; \CLB{Second Best}]}
\adjustbox{max width=1.0\textwidth}{
\begin{tabular}{l|ccc|ccc|ccc|ccc|c}
\toprule
Index               & \multicolumn{3}{c|}{Full-FR $\uparrow$} & \multicolumn{3}{c|}{Full-NR $\uparrow$} & \multicolumn{3}{c|}{Image-FR $\uparrow$} & \multicolumn{3}{c|}{Image-NR $\uparrow$} & \multicolumn{1}{c}{\multirow{2}{*}{Overall $\uparrow$}} \\
T2I Model               & NSI     & SCI     & AIGI    & NSI     & SCI     & AIGI    & NSI      & SCI     & AIGI    & NSI      & SCI     & AIGI    & \multicolumn{1}{c}{} \\ \midrule
DiffBIR     & \CLA{3.052}&\CLA{2.785}&\CLA{2.877}           & \CLA{2.778}&\CLB{2.380}&2.517            & \CLA{2.899}&\CLA{2.804}&\CLA{2.873}           & 1.847&1.644&1.674            & \CLA{2.647}                       \\
\rowcolor[HTML]{D0D0D0} 
PASD        & \CLB{2.796}&\CLB{2.621}&\CLB{2.741}           & 2.339&1.932&2.367            & \CLB{2.652}&\CLB{2.583}&2.675           & \CLB{2.056}&1.818&2.141            & \CLB{2.494}                       \\
RealVis     & 2.617&2.475&2.584           & 1.914&1.808&2.445            & 2.509&2.441&2.558           & 1.686&1.675&2.110            & 2.331                       \\
\rowcolor[HTML]{D0D0D0} 
PG25        & 2.145&1.922&2.123           & \CLB{2.730}&\CLA{2.582}&\CLA{3.509}            & 1.952&1.895&2.040           & \CLA{2.750}&\CLA{2.852}&\CLA{3.459}            & 2.330                       \\
SSD-1B      & 2.515&2.407&2.554           & 1.905&1.878&2.516            & 2.386&2.355&2.512           & 1.758&1.783&2.309            & 2.305                       \\
\rowcolor[HTML]{D0D0D0} 
PG20        & 2.435&2.245&2.376           & 2.200&2.072&\CLB{2.893}            & 2.263&2.174&2.301           & 2.011&\CLB{1.992}&\CLB{2.683}            & 2.299                       \\
StableSR    & 2.599&2.591&2.688           & 1.401&1.373&1.549            & 2.576&2.582&\CLB{2.679}           & 1.392&1.367&1.541            & 2.222                       \\
\rowcolor[HTML]{D0D0D0} 
Dreamlike   & 2.570&2.421&2.509           & 1.760&1.659&1.958            & 2.413&2.376&2.482           & 1.446&1.471&1.645            & 2.194                       \\
SD15        & 2.607&2.379&2.444           & 1.787&1.652&1.877            & 2.464&2.333&2.436           & 1.538&1.497&1.644            & 2.190                       \\
\rowcolor[HTML]{D0D0D0} 
SDXL        & 2.436&2.330&2.484           & 1.606&1.610&1.862            & 2.333&2.275&2.442           & 1.480&1.524&1.698            & 2.129                       \\
Animate     & 2.293&2.213&2.392           & 1.743&1.703&2.129            & 2.223&2.210&2.334           & 1.519&1.600&1.757            & 2.094                       \\
\rowcolor[HTML]{D0D0D0} 
InstructPix & 2.082&2.207&2.190           & 1.854&1.579&1.679            & 2.249&2.388&2.432           & 1.204&1.240&1.227            & 1.989 \\ \bottomrule                   
\end{tabular}
}
\vspace{-3mm}
\label{tab:t2i-high}
\end{table*}
\begin{table*}[t]
\centering
\caption{Benchmark result in \textit{Pixel} and \textit{Text} mode for 8 T2I generative models at extremely-low bitrate (10,000 times compression). [Key: \CLA{Best}; \CLB{Second Best}]}
\adjustbox{max width=1.0\textwidth}{
\begin{tabular}{l|ccc|ccc|ccc|ccc|c}
\toprule
Index               & \multicolumn{3}{c|}{Pixel-FR $\uparrow$} & \multicolumn{3}{c|}{Pixel-NR $\uparrow$} & \multicolumn{3}{c|}{Text-FR $\uparrow$} & \multicolumn{3}{c|}{Text-NR $\uparrow$} & \multicolumn{1}{c}{\multirow{2}{*}{Overall $\uparrow$}} \\
T2I Model               & NSI     & SCI     & AIGI    & NSI     & SCI     & AIGI    & NSI      & SCI     & AIGI    & NSI      & SCI     & AIGI    & \multicolumn{1}{c}{} \\ \midrule
PG25      & 1.789&1.641&1.779           & \CLA{3.542}&\CLA{3.425}&\CLA{3.939}            & 1.798&1.634&1.762           & \CLB{3.646}&\CLA{3.628}&\CLB{3.944}            & \CLA{2.386}                       \\
\rowcolor[HTML]{D0D0D0} 
RealVis   & \CLA{2.041}&\CLB{1.872}&\CLA{2.033}           & \CLB{2.591}&\CLB{2.502}&3.316            & \CLA{1.868}&\CLA{1.668}&\CLB{1.777}           & 3.428&3.295&3.734            & \CLB{2.300}                       \\
PG20      & 1.901&1.812&1.948           & 2.472&2.325&\CLB{3.338}            & 1.772&1.619&1.745           & \CLA{3.675}&\CLB{3.617}&\CLA{3.963}            & 2.274                       \\
\rowcolor[HTML]{D0D0D0} 
SSD-1B    & 1.990&1.864&\CLB{2.019}           & 2.265&2.271&2.984            & \CLB{1.852}&\CLB{1.661}&\CLA{1.787}           & 3.409&3.285&3.760             & 2.239                       \\
Animate   & 1.828&1.743&1.902           & 2.306&2.159&2.875            & 1.750&1.615&1.712           & 3.485&3.296&3.717            & 2.163                       \\
\rowcolor[HTML]{D0D0D0} 
Dreamlike & 1.986&\CLA{1.877}&1.991           & 2.195&2.129&2.623            & 1.779&1.620&1.705           & 3.233&2.917&3.302            & 2.132                       \\
SDXL      & 1.923&1.824&1.980           & 1.830&1.879&2.255            & 1.822&1.633&1.762           & 3.358&3.224&3.708            & 2.118                       \\
\rowcolor[HTML]{D0D0D0} 
SD15      & \CLB{2.000}&1.856&1.951           & 2.165&1.948&2.314            & 1.760&1.609&1.654           & 2.683&2.364&2.498            & 1.988  \\ \bottomrule                     
\end{tabular}
}
\vspace{-3mm}
\label{tab:t2i-low}
\end{table*}
\begin{table*}[t]
\centering
\caption{Benchmark result in \textit{Full} and \textit{Pixel} mode for 6 I2T models. [Key: \CLA{Best}; \CLB{Second Best}].}
\adjustbox{max width=1.0\textwidth}{
\begin{tabular}{l|ccc|ccc|ccc|ccc|c}
\toprule
Index               & \multicolumn{3}{c|}{Full-FR $\uparrow$} & \multicolumn{3}{c|}{Full-NR $\uparrow$} & \multicolumn{3}{c|}{Pixel-FR $\uparrow$} & \multicolumn{3}{c|}{Pixel-NR $\uparrow$} & \multicolumn{1}{c}{\multirow{2}{*}{Overall $\uparrow$}} \\
I2T Model               & NSI     & SCI     & AIGI    & NSI     & SCI     & AIGI    & NSI      & SCI     & AIGI    & NSI      & SCI     & AIGI    & \multicolumn{1}{c}{} \\ \midrule
GPT-4o       & \CLA{2.617}   & 2.475   & \CLA{2.584}   & \CLB{1.914}   & \CLB{1.808}   & \CLB{2.445}   & \CLA{2.041}    & 1.872   & 2.033   & \CLA{2.591}    & \CLB{2.502}   & \CLA{3.316}   & \CLA{2.439}                                        \\
\rowcolor[HTML]{D0D0D0} 
ShareGPT    & \CLB{2.607}   & \CLA{2.479}   & 2.577   & \CLA{1.925}   & \CLA{1.870}    & \CLA{2.446}   & 2.032    & 1.872   & \CLA{2.042}   & \CLB{2.543}    & \CLA{2.556}   & \CLB{3.259}   & \CLB{2.432}                                        \\
Qwen        & 2.592   & 2.473   & \CLB{2.581}   & 1.894   & 1.799   & 2.353   & 2.034    & \CLB{1.890}    & \CLB{2.036}   & 2.531    & 2.364   & 3.176   & 2.396                                        \\
\rowcolor[HTML]{D0D0D0} 
MPlugOwl-2   & 2.605   & 2.477   & 2.568   & 1.910    & 1.808   & 2.314   & \CLB{2.035}    & \CLA{1.892}   & 2.028   & 2.504    & 2.391   & 3.075   & 2.384                                        \\
LLAVA-1.5       & 2.599   & 2.465   & 2.565   & 1.880    & 1.799   & 2.276   & 2.025    & 1.876   & 2.028   & 2.498    & 2.420    & 3.041   & 2.381                                        \\
\rowcolor[HTML]{D0D0D0} 
InstructBLIP & 2.589   & 2.473   & 2.571   & 1.842   & 1.736   & 2.192   & 2.027    & 1.882   & 2.035   & 2.424    & 2.339   & 2.961   & 2.346   \\ \bottomrule                                    
\end{tabular}
}
\vspace{-3mm}
\label{tab:i2t}
\end{table*}

For 6 indicators in I2T LMMs, while GPT-4o \cite{i2t:gpt4} does not perform well on Text-FR, it significantly outperforms on Full-FR. This suggests that although its generated text carries limited information, it has a strong orthogonal relationship with the low-level details of the image. This semantic information effectively compensates for the information loss after image compression. In addition, the given text facilitates the subsequent T2I model in decoding high-quality images, and its performance across various NR indicators is also commendable. In comparison, MPlugOwl-2 \cite{i2t:mplugowl} and InstructBLIP \cite{i2t:iblip} can effectively encode images into text, but their results are still inferior to GPT-4o. The only viable competitor is ShareGPT \cite{i2t:sharegpt4v}, but it has a bpp of around 0.008, which is significantly larger than the other 5 models. This data size exceeds ELB and occupies one-third of the available ULB space. Considering multiple factors, GPT-4o remains the most suitable I2T model as the CMC encoder.

For 8 indicators in T2I LMMs, 2 restorative models ~\cite{ir:diffbir,ir:pasd} exhibit overwhelming consistency in \textit{Full} and \textit{Image} modes with acceptable perception results, enabling faithful image reconstruction close to the ground truth. However, its applicability is limited for the other 2 modes, particularly under the strict ELB conditions. The performance of the remaining models falls into two distinct extremes, where RealVis \cite{gen:RealVis} shows high consistency but PG25 \cite{gen:Playground25} demonstrates high perception. Given that it is feasible to enhance a compressed low-quality image with high fidelity to the original, while correcting a completely different high-quality image with low fidelity remains challenging, we opt to prioritize consistency by assigning it a higher weight. Consequently, considering the strong performance and wide versatility of RealVis, it is relatively more suitable than the CMC decoder.


To delve into the compression capability of I2T and T2I LMMs with different content on various modes, we present the T2I leaderboard under ULB and ELB conditions in Table \ref{tab:t2i-high} and Table \ref{tab:t2i-low}, respectively, and showcase the performance of I2T models on \textit{Full} and \textit{Pixel} modes (\textit{Text} mode attached in supplementary) in Table \ref{tab:i2t}, with a discussion of content-specific analysis. A horizontal comparison among different modes in Tables 3 and 4 reveals that the \textit{Full} mode has a clear advantage over the \textit{Image} mode in terms of consistency and perception, indicating the significance of the text provided by the I2T model for T2I decoding. This text guidance not only enhances consistency but provides a clear target for the T2I process, thus also boosting perception. In contrast, the \textit{Pixel} mode sacrifices perception for consistency compared to the \textit{Text} mode. This is because the more control conditions added, the less room for creative freedom the model has, leading to a decrease in image aesthetics. However, for models that already have high perception scores \cite{gen:Playground25,gen:Playground20} in the \textit{Text} mode, the trade-off of improving overall performance is acceptable.

Among NSI, SCI, and AIGI, different LMMs excel at different content. For instance, as shown in Table \ref{tab:t2i-high} and Table \ref{tab:t2i-low}, PG25 \cite{gen:Playground25}, trained on internet data, performs better in AIGI tasks; conversely, RealVis aims at image naturalness, manifesting its superior reconstruction capability in NSI. Regardless of the model employed, we observe that NSI generally yields higher consistency scores, while AIGI has higher perception scores. However, SCI stands out from the others, with the compression results of the same model lagging behind in both perception and consistency. This deficiency is relevant to certain words \cite{dataset:webpage} (even long paragraphs) within SCI, making I2T models unable to re-encode them into text, while the text generation capabilities of recent T2I models are still limited.
Besides, although the performance disparities among I2T models are not as significant as those in T2I models, Table \ref{tab:i2t} also clearly illustrates the limitation in SCI, indicating room for further optimization.

\begin{figure*}[t]
    \centering
    \subfigure[\textit{Full} and \textit{Image} modes for 8 T2I generative and 4 restorative models]{\includegraphics[width=1.0\textwidth]{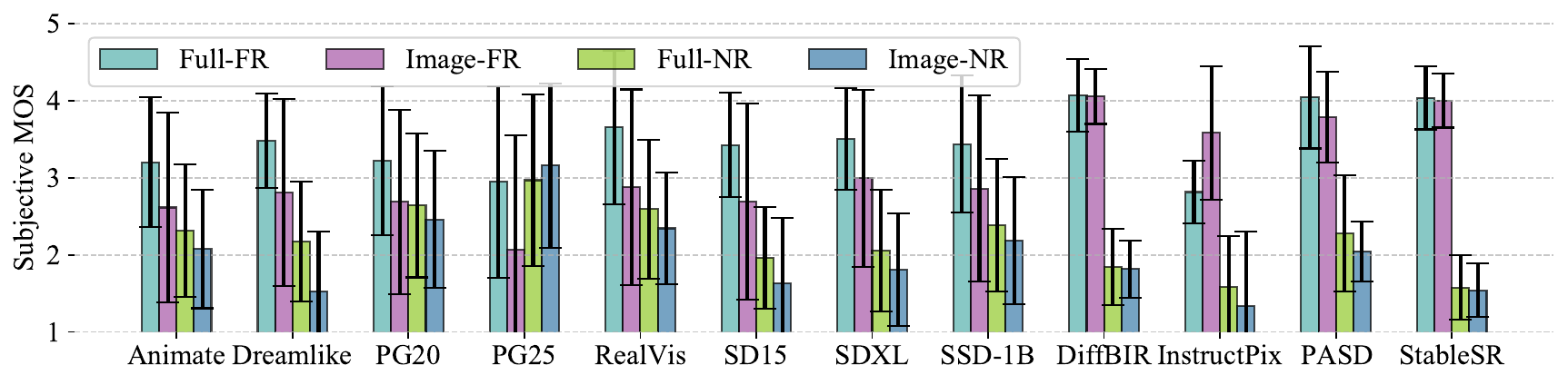}}\vspace{-4mm}
    \subfigure[\textit{Pixel} and \textit{Text} modes for 8 T2I generative models]{\includegraphics[width=0.66\textwidth]{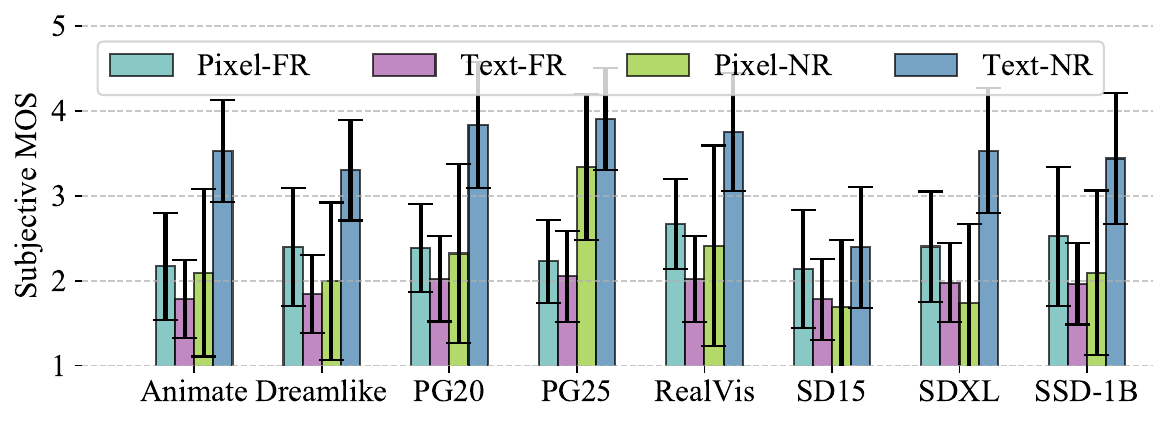}}
    \subfigure[Subjective leaderboard]{
    \begin{minipage}{.31\linewidth}
        \centering
        \vspace{-32mm}
        \resizebox{0.8\columnwidth}{!}{
        \begin{tabular}{l|l}
        \toprule
            ULB   & ELB   \\ \midrule
            PASD (1)        & PG25 (1)      \\
            DiffBIR (2)     & RealVis (2)   \\
            RealVis (3)     & PG20 (3)      \\
            StableSR (4)    & SSD-1B (4)    \\
            PG25 (5)        & SDXL (5)      \\
            PG20 (6)        & Animate (6)   \\
            SSD-1B (7)      & Dreamlike (7) \\
            SDXL (8)        & SD15 (8)      \\ \bottomrule
        \end{tabular}}
    \end{minipage}\hfill}\hskip.2em
    \vspace{-4pt}
    \caption{Illustration of subjective preference in terms of Mean Opinion Score.}
    \vspace{-4mm}
    \label{fig:subject-bar}
\end{figure*}

\subsection{Subjective Data Analysis}

Figure \ref{fig:subject-bar} presents the subjective preference for images decoded by 12 T2I models under ULB and 8 models under ELB. For ULB, the 3 restorative models ~\cite{ir:diffbir,ir:pasd,ir:stablesr} exhibit slightly higher consistency compared to generative models, where PG25 achieves the highest perception score against all others. It is worth noting that the restorative models are more robust. The upper and lower bounds of the scores in each dimension seldom surpass 1.0, whereas the randomness of the generative models notably deteriorates. As the bitrate further decreases to ELB, consistency scores of all models decline, while perception scores have slight improvement. In summary, apart from Animate \cite{gen:animatediff} specifically for cartoon styles, and InstructPix \cite{ir:instructpix2pix} that significantly alters images, all other models demonstrate potential applications in CMC. Additionally, by averaging all scores, we find that the models ranking based on subjective scores aligns closely with the objective ones shown in Table \ref{tab:t2i-high} and Table \ref{tab:t2i-low}. This finding validates the reasonability of our previous experiments and highlights that, compared to perception, humans tend to focus more on consistency when viewing compressed images.

\subsection{Compare with Traditional Codecs}
\begin{figure}[t]
    \centering
    \subfigure{\includegraphics[width=0.32\textwidth]{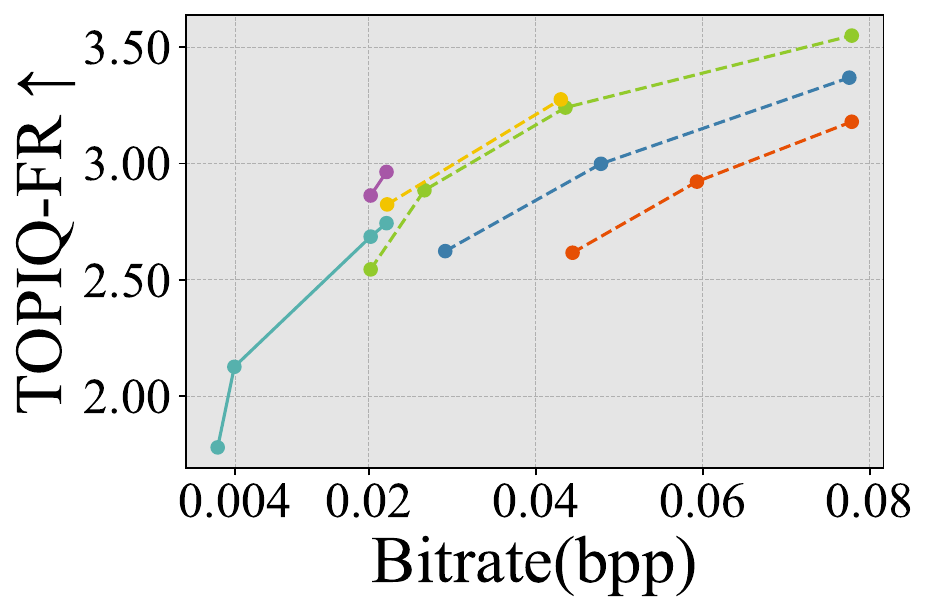}
    }\vspace{-1.5mm}
    \subfigure{\includegraphics[width=0.32\textwidth]{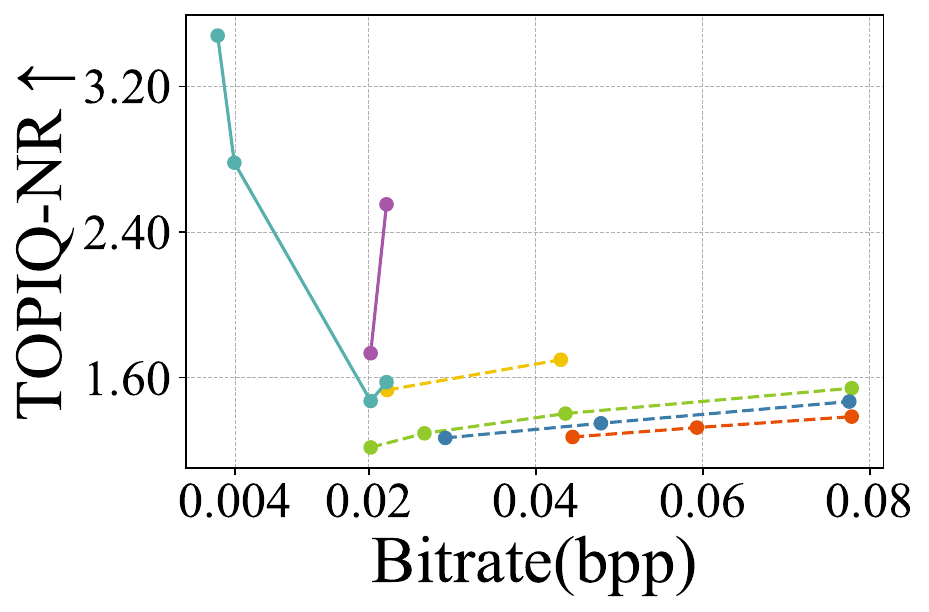}
    }\vspace{-1.5mm}
    \subfigure{\includegraphics[width=0.32\textwidth]{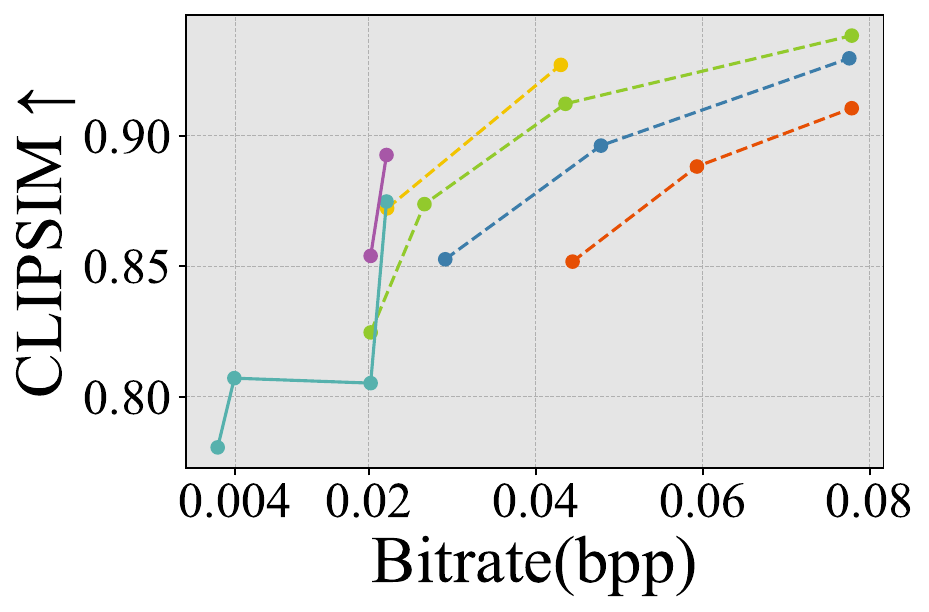}
    }\vspace{-1.5mm}
    \subfigure{\includegraphics[width=0.32\textwidth]{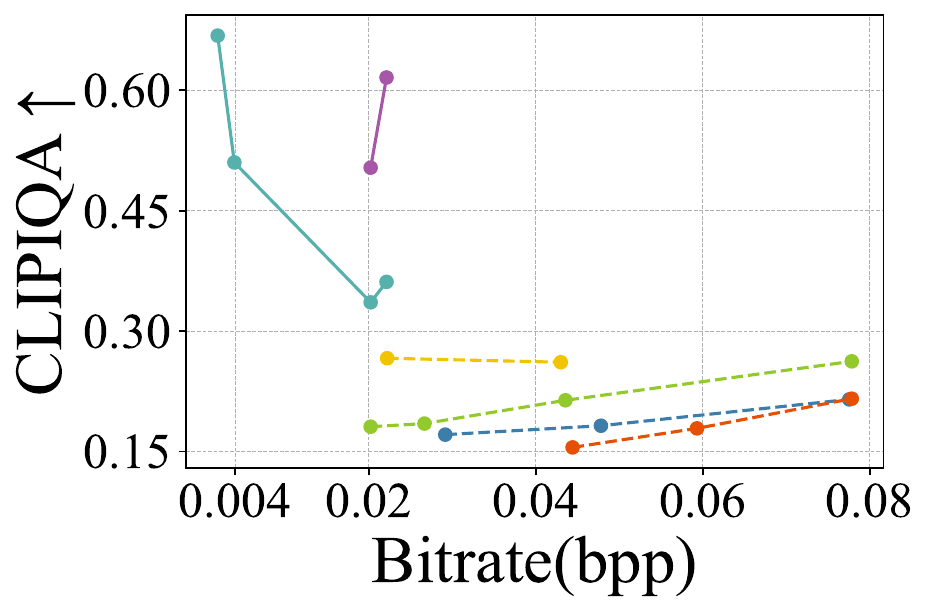}
    }\vspace{-1.5mm}
    \subfigure{\includegraphics[width=0.32\textwidth]{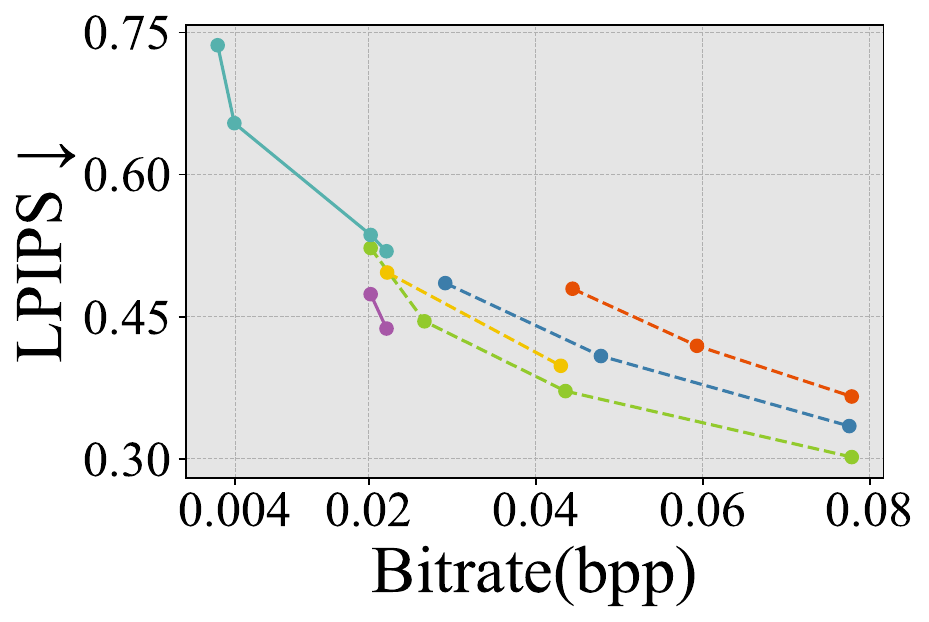}
    }\vspace{-1.5mm}
    \subfigure{\includegraphics[width=0.32\textwidth]{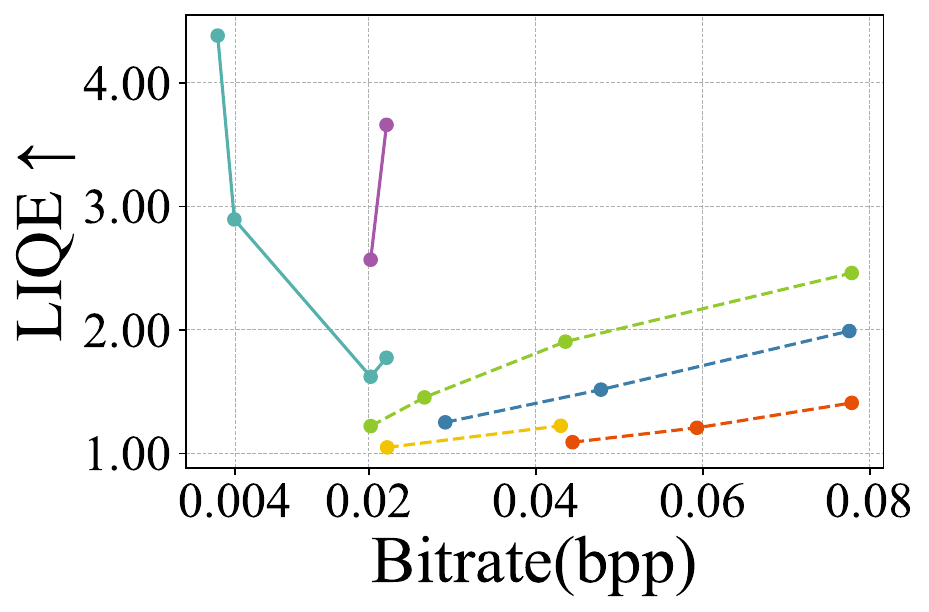}
    }\vspace{-1.5mm} 
    \subfigure{\includegraphics[width=0.32\textwidth]{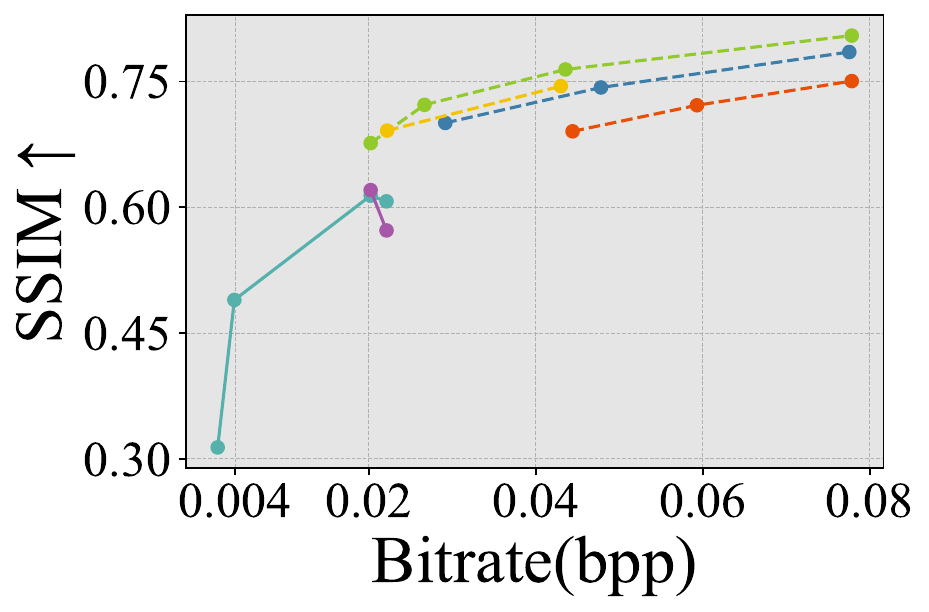}
    }\vspace{-1.5mm} 
    \subfigure{\includegraphics[width=0.32\textwidth]{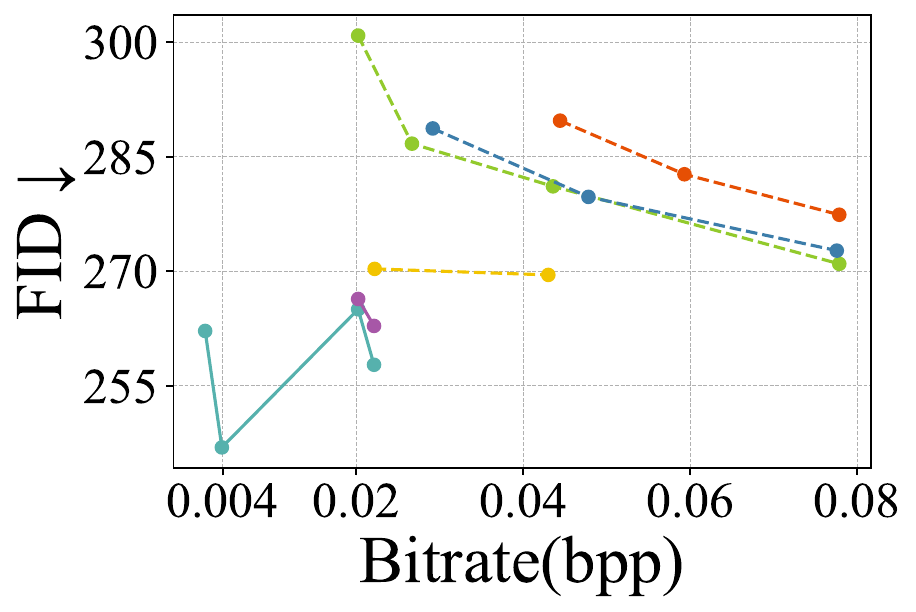}
    }\vspace{-1.5mm}  
    \subfigure{
    \centering
    \begin{minipage}{0.32\linewidth}
        \centering
        \vspace{-30mm}
        \hspace{4mm}
        \includegraphics[width=0.7\textwidth]{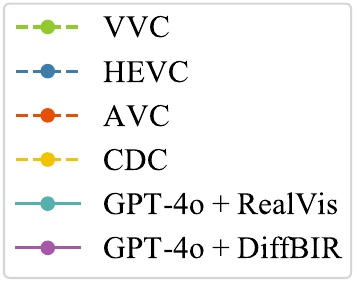}
    \end{minipage}
    
    }
    \vspace{-8pt}
    \caption{Comparison of CMC-Bench winners against existing image codecs, evaluated by 4 consistency and 4 perception metrics indicated by \colorbox{lightgray!60!white}{marked} and plain background. The combination of I2T and T2I models generally exceeds the existing codecs under the same bitrate.}
    \vspace{-4mm}
    \label{fig:curve}
\end{figure}

To validate the practicality of the CMC paradigm, we select 2 outstanding combinations of I2T and T2I models from CMC-Bench, and compare them with 3 mainstream codecs: AVC \cite{metric:tra-avc}, HEVC \cite{metric:tra-hevc}, and VVC \cite{metric:tra-vvc} at I-Frame mode, and the latest semantic codec pipeline CDC \cite{metric:tra-cdc}. Given the superiority of GPT-4o \cite{i2t:gpt4} as the encoder, we initially pair it with the top-ranked decoder DiffBIR \cite{ir:diffbir}. Considering applications on different modes, excluding the reconstructive model, we also assess its performance with the third-tanked decoder RealVis \cite{gen:RealVis}. These two approaches with four bitrates correspond to \textit{Text}, \textit{Pixel}, \textit{Image}, \textit{Full} modes are shown in Figure \ref{fig:curve}. To comprehensively compare the two paradigms across different dimensions, we add 3 Consistency metrics: CLIPSIM \cite{quality:clipsim}, LPIPS \cite{quality:lpips}, and SSIM \cite{quality:ssim}; and 3 Perception metrics: CLIPIQA \cite{quality:CLIPIQA}, LIQE \cite{quality:LIQE}, and FID \cite{quality:fid}. Models ranked higher prioritize semantic information, while those lower focus on pixel-level consistency.

Both CMC paradigms demonstrate an advance in terms of most metrics. Given that SSIM is purely pixel-based, the performance drop due to generative compression is expected. The lead in perception is particularly notable, as it surpasses traditional codecs at extremely low bitrates. However, the advantage in consistency is relatively smaller, achieving a reduction of around 30\% in bitrate compared to traditional methods at 0.02 bpp. The DiffBIR decoder generally shows better performance, while RealVis fits A wider range of bitrates.
In summary, based on the above analyses, we believe that CMC holds a certain advantage over traditional encoding. However, for implementing LMMs into the next generation of visual signal codecs, further optimization is required in the following aspects:

\CLA{Enhanced T2I models}: Both encoders and decoders are crucial in CMC, but decoders are more decisive. Future T2I models should possess more sophisticated control mechanisms, ensuring high-quality generation while maintaining consistency with reference images and text.

\CLA{Better adaption to SCI}: the compression performance of SCI is inferior to NSI and AIGI,  necessitating LMMs with specialized understanding and generating mechanisms to handle SCI.

\CLA{Wider bitrate range}: Although leading in ULB and ELB, the margin of consistency improvement is not as pronounced as perception. Future efforts should focus on CMC at higher bitrates, incorporating more control information to aid in reconstructing the original image, ultimately achieving superiority across all bitrates and dimensions as compared to traditional codecs.

\section{Conclusion}

We construct CMC-Bench, a benchmark for assessing the collaborative functioning of I2T and T2I models in image compression. Anticipating the bitrate requirements for codecs in the next decade, we proposed four collaboration modes among LMMs, along with two indicators of consistency and perception. By employing 6 mainstream I2T and 12 T2I models, we collected 58,000 distorted images through CMC with 160,000 human subjective annotations to train objective metrics for comprehensive evaluation. Our assessment demonstrates that even without dedicated training for compression tasks, combinations of several advanced I2T and T2I models have already surpassed traditional codecs in multiple aspects. However, there is still a long way to go before LMMs can directly become the future codecs paradigm. We sincerely hope that CMC-Bench will inspire future LMMs to perform better compression towards the evolution of visual signal codecs.


{\small
\bibliographystyle{unsrt}
\bibliography{neurips_data_2024}

\begin{thebibliography}{10}

\bibitem{metric:tra-mpeg}
Didier Le~Gall.
\newblock Mpeg: A video compression standard for multimedia applications.
\newblock {\em Communications of the ACM}, 34(4):46--58, 1991.

\bibitem{metric:tra-jpeg}
Athanassios~N. Skodras, Charilaos~A. Christopoulos, and Touradj Ebrahimi.
\newblock The jpeg 2000 still image compression standard.
\newblock {\em IEEE Signal Process. Mag.}, 18:36--58, 2001.

\bibitem{metric:tra-avc}
Thomas Wiegand, Gary~J. Sullivan, Gisle Bj{\o}ntegaard, and Ajay Luthra.
\newblock Overview of the h.264/avc video coding standard.
\newblock {\em IEEE Trans. Circuits Syst. Video Technol.}, 13:560--576, 2003.

\bibitem{metric:tra-hevc}
Gary~J. Sullivan, Jens-Rainer Ohm, Woojin Han, and Thomas Wiegand.
\newblock Overview of the high efficiency video coding (hevc) standard.
\newblock {\em IEEE Transactions on Circuits and Systems for Video Technology}, 22:1649--1668, 2012.

\bibitem{metric:tra-vvc}
Benjamin Bross, Ye-Kui Wang, Yan Ye, Shan Liu, Jianle Chen, Gary~J. Sullivan, and Jens-Rainer Ohm.
\newblock Overview of the versatile video coding (vvc) standard and its applications.
\newblock {\em IEEE Transactions on Circuits and Systems for Video Technology}, 31:3736--3764, 2021.

\bibitem{intro:tradeoff-1}
Yochai Blau and Tomer Michaeli.
\newblock The perception-distortion tradeoff.
\newblock In {\em Proceedings of the IEEE conference on computer vision and pattern recognition}, pages 6228--6237, 2018.

\bibitem{intro:tradeoff-2}
Yochai Blau and Tomer Michaeli.
\newblock Rethinking lossy compression: The rate-distortion-perception tradeoff.
\newblock In {\em International Conference on Machine Learning}, pages 675--685. PMLR, 2019.

\bibitem{relate:bench-i2t-mm}
Yuan Liu, Haodong Duan, Yuanhan Zhang, Bo~Li, Songyang Zhang, Wangbo Zhao, Yike Yuan, Jiaqi Wang, Conghui He, Ziwei Liu, Kai Chen, and Dahua Lin.
\newblock Mmbench: Is your multi-modal model an all-around player?, 2024.

\bibitem{relate:bench-i2t-seed}
Bohao Li, Rui Wang, Guangzhi Wang, Yuying Ge, Yixiao Ge, and Ying Shan.
\newblock Seed-bench: Benchmarking multimodal llms with generative comprehension, 2023.

\bibitem{relate:bench-t2i-hrs}
Eslam~Mohamed Bakr, Pengzhan Sun, Xiaogian Shen, Faizan~Farooq Khan, Li~Erran Li, and Mohamed Elhoseiny.
\newblock Hrs-bench: Holistic, reliable and scalable benchmark for text-to-image models.
\newblock In {\em Proceedings of the IEEE/CVF International Conference on Computer Vision}, pages 20041--20053, 2023.

\bibitem{relate:bench-t2i-comp}
Kaiyi Huang, Kaiyue Sun, Enze Xie, Zhenguo Li, and Xihui Liu.
\newblock T2i-compbench: A comprehensive benchmark for open-world compositional text-to-image generation.
\newblock {\em Advances in Neural Information Processing Systems}, 36:78723--78747, 2023.

\bibitem{relate:cmc-cmc}
Jiguo Li, Chuanmin Jia, Xinfeng Zhang, Siwei Ma, and Wen Gao.
\newblock Cross modal compression: Towards human-comprehensible semantic compression.
\newblock In {\em Proceedings of the 29th ACM international conference on multimedia}, pages 4230--4238, 2021.

\bibitem{relate:cmc-text}
Eric Lei, Yiğit~Berkay Uslu, Hamed Hassani, and Shirin~Saeedi Bidokhti.
\newblock Text + sketch: Image compression at ultra low rates, 2023.

\bibitem{add:controlnet}
Lvmin Zhang, Anyi Rao, and Maneesh Agrawala.
\newblock Adding conditional control to text-to-image diffusion models.
\newblock In {\em IEEE International Conference on Computer Vision (ICCV)}, 2023.

\bibitem{relate:cmc-rcmc}
Junlong Gao, Jiguo Li, Chuanmin Jia, Shanshe Wang, Siwei Ma, and Wen Gao.
\newblock Cross modal compression with variable rate prompt.
\newblock {\em IEEE Transactions on Multimedia}, 26:3444--3456, 2024.

\bibitem{relate:cmc-mcmc}
Junlong Gao, Chuanmin Jia, Zhimeng Huang, Shanshe Wang, Siwei Ma, and Wen Gao.
\newblock Rate-distortion optimized cross modal compression with multiple domains.
\newblock {\em IEEE Transactions on Circuits and Systems for Video Technology}, pages 1--1, 2024.

\bibitem{relate:cmc-vqgan}
Qi~Mao, Tinghan Yang, Yinuo Zhang, Zijian Wang, Meng Wang, Shiqi Wang, Libiao Jin, and Siwei Ma.
\newblock Extreme image compression using fine-tuned vqgans.
\newblock In {\em 2024 Data Compression Conference (DCC)}, pages 203--212, 2024.

\bibitem{relate:cmc-uigc}
Naifu Xue, Qi~Mao, Zijian Wang, Yuan Zhang, and Siwei Ma.
\newblock Unifying generation and compression: Ultra-low bitrate image coding via multi-stage transformer, 2024.

\bibitem{relate:cmc-onceforall}
Anqi Li, Yuxi Liu, Huihui Bai, Feng Li, Runmin Cong, Meng Wang, and Yao Zhao.
\newblock Once-for-all: Controllable generative image compression with dynamic granularity adaption, 2024.

\bibitem{relate:cmc-misc}
Chunyi Li, Guo Lu, Donghui Feng, Haoning Wu, Zicheng Zhang, Xiaohong Liu, Guangtao Zhai, Weisi Lin, and Wenjun Zhang.
\newblock Misc: Ultra-low bitrate image semantic compression driven by large multimodal model, 2024.

\bibitem{relate:bench-i2t-qbench}
Haoning Wu, Zicheng Zhang, Erli Zhang, Chaofeng Chen, Liang Liao, Annan Wang, Chunyi Li, Wenxiu Sun, Qiong Yan, Guangtao Zhai, and Weisi Lin.
\newblock Q-bench: A benchmark for general-purpose foundation models on low-level vision, 2024.

\bibitem{relate:bench-i2t-fakebench}
Yixuan Li, Xuelin Liu, Xiaoyang Wang, Shiqi Wang, and Weisi Lin.
\newblock Fakebench: Uncover the achilles' heels of fake images with large multimodal models, 2024.

\bibitem{add:a-bench}
Zicheng Zhang, Haoning Wu, Chunyi Li, Yingjie Zhou, Wei Sun, Xiongkuo Min, Zijian Chen, Xiaohong Liu, Weisi Lin, and Guangtao Zhai.
\newblock A-bench: Are lmms masters at evaluating ai-generated images?, 2024.

\bibitem{add:q-instruct}
Haoning Wu, Zicheng Zhang, Erli Zhang, Chaofeng Chen, Liang Liao, Annan Wang, Kaixin Xu, Chunyi Li, Jingwen Hou, Guangtao Zhai, et~al.
\newblock Q-instruct: Improving low-level visual abilities for multi-modality foundation models, 2023.

\bibitem{relate:bench-t2i-draw}
Chitwan Saharia, William Chan, Saurabh Saxena, Lala Li, Jay Whang, Emily~L Denton, Kamyar Ghasemipour, Raphael Gontijo~Lopes, Burcu Karagol~Ayan, Tim Salimans, et~al.
\newblock Photorealistic text-to-image diffusion models with deep language understanding.
\newblock {\em Advances in neural information processing systems}, 35:36479--36494, 2022.

\bibitem{relate:bench-t2i-dalleval}
Jaemin Cho, Abhay Zala, and Mohit Bansal.
\newblock Dall-eval: Probing the reasoning skills and social biases of text-to-image generation models.
\newblock In {\em Proceedings of the IEEE/CVF International Conference on Computer Vision}, pages 3043--3054, 2023.

\bibitem{dataset:clic}
Johannes Ball{\'e}, Philip~A Chou, David Minnen, Saurabh Singh, Nick Johnston, Eirikur Agustsson, Sung~Jin Hwang, and George Toderici.
\newblock Nonlinear transform coding.
\newblock {\em IEEE Journal of Selected Topics in Signal Processing}, 15(2):339--353, 2020.

\bibitem{relate:dataset-ntire2022}
Jinjin Gu, Haoming Cai, Chao Dong, Jimmy~S Ren, Radu Timofte, Yuan Gong, Shanshan Lao, Shuwei Shi, Jiahao Wang, Sidi Yang, et~al.
\newblock Ntire 2022 challenge on perceptual image quality assessment.
\newblock In {\em Proceedings of the IEEE/CVF conference on computer vision and pattern recognition}, pages 951--967, 2022.

\bibitem{dataset:scid}
Zhangkai Ni, Lin Ma, Huanqiang Zeng, Jing Chen, Canhui Cai, and Kai-Kuang Ma.
\newblock Esim: Edge similarity for screen content image quality assessment.
\newblock {\em IEEE Transactions on Image Processing}, 26(10):4818--4831, 2017.

\bibitem{dataset:cct}
Xiongkuo Min, Kede Ma, Ke~Gu, Guangtao Zhai, Zhou Wang, and Weisi Lin.
\newblock Unified blind quality assessment of compressed natural, graphic, and screen content images.
\newblock {\em IEEE Transactions on Image Processing}, 26(11):5462--5474, 2017.

\bibitem{dataset:agiqa-3k}
Chunyi Li, Zicheng Zhang, Haoning Wu, Wei Sun, Xiongkuo Min, Xiaohong Liu, Guangtao Zhai, and Weisi Lin.
\newblock Agiqa-3k: An open database for ai-generated image quality assessment.
\newblock {\em IEEE Transactions on Circuits and Systems for Video Technology}, 2023.

\bibitem{relate:dataset-imagereward}
Jiazheng Xu, Xiao Liu, Yuchen Wu, Yuxuan Tong, Qinkai Li, Ming Ding, Jie Tang, and Yuxiao Dong.
\newblock Imagereward: Learning and evaluating human preferences for text-to-image generation.
\newblock {\em Advances in Neural Information Processing Systems}, 36, 2024.

\bibitem{dataset:agiqa1k}
Zicheng Zhang, Chunyi Li, Wei Sun, Xiaohong Liu, Xiongkuo Min, and Guangtao Zhai.
\newblock A perceptual quality assessment exploration for aigc images.
\newblock In {\em IEEE International Conference on Multimedia and Expo Workshops (ICMEW)}, pages 440--445, 2023.

\bibitem{add:aspect-qoe}
Chunyi Li, May Lim, Abdelhak Bentaleb, and Roger Zimmermann.
\newblock A real-time blind quality-of-experience assessment metric for http adaptive streaming.
\newblock In {\em IEEE International Conference on Multimedia and Expo}, 2023.

\bibitem{add:cartoon}
Chunyi Li, Zicheng Zhang, Wei Sun, Xiongkuo Min, and Guangtao Zhai.
\newblock A full-reference quality assessment metric for cartoon images.
\newblock In {\em IEEE 24th International Workshop on Multimedia Signal Processing}, 2022.

\bibitem{add:rr}
Zicheng Zhang, Yingjie Zhou, Chunyi Li, Kang Fu, Wei Sun, Xiaohong Liu, Xiongkuo Min, and Guangtao Zhai.
\newblock A reduced-reference quality assessment metric for textured mesh digital humans.
\newblock In {\em IEEE Int. Conf. Acoust., Speech, and Signal Processing}, 2024.

\bibitem{add:6G}
Zicheng Zhang, Yingjie Zhou, Long Teng, Wei Sun, Chunyi Li, Xiongkuo Min, Xiao-Ping Zhang, and Guangtao Zhai.
\newblock Quality-of-experience evaluation for digital twins in 6g network environments.
\newblock {\em IEEE Transactions on Broadcasting}, 2024.

\bibitem{add:advancing}
Zicheng Zhang, Wei Sun, Yingjie Zhou, Haoning Wu, Chunyi Li, Xiongkuo Min, Xiaohong Liu, Guangtao Zhai, and Weisi Lin.
\newblock Advancing zero-shot digital human quality assessment through text-prompted evaluation, 2023.

\bibitem{add:gms}
Zicheng Zhang, Wei Sun, Houning Wu, Yingjie Zhou, Chunyi Li, Xiongkuo Min, Guangtao Zhai, and Weisi Lin.
\newblock Gms-3dqa: Projection-based grid mini-patch sampling for 3d model quality assessment, 2023.

\bibitem{add:iscas}
Chunyi Li, Zicheng Zhang, Haoning Wu, Kaiwei Zhang, Lei Bai, Xiaohong Liu, Guangtao Zhai, and Weisi Lin.
\newblock Paps-ovqa: Projection-aware patch sampling for omnidirectional video quality assessment.
\newblock In {\em IEEE Int. Sym. Circuits and Systems}, 2024.

\bibitem{add:lmmpcqa}
Zicheng Zhang, Haoning Wu, Yingjie Zhou, Chunyi Li, Wei Sun, Chaofeng Chen, Xiongkuo Min, Xiaohong Liu, Weisi Lin, and Guangtao Zhai.
\newblock Lmm-pcqa: Assisting point cloud quality assessment with lmm, 2024.

\bibitem{dataset:mscoco}
Tsung-Yi Lin, Michael Maire, Serge Belongie, James Hays, Pietro Perona, Deva Ramanan, Piotr Doll{\'a}r, and C~Lawrence Zitnick.
\newblock Microsoft coco: Common objects in context.
\newblock In {\em Computer Vision--ECCV 2014: 13th European Conference, Zurich, Switzerland, September 6-12, 2014, Proceedings, Part V 13}, pages 740--755. Springer, 2014.

\bibitem{perception:q-align}
Haoning Wu, Zicheng Zhang, Weixia Zhang, Chaofeng Chen, Liang Liao, Chunyi Li, Yixuan Gao, Annan Wang, Erli Zhang, Wenxiu Sun, et~al.
\newblock Q-align: Teaching lmms for visual scoring via discrete text-defined levels, 2023.

\bibitem{dataset:cgiqa}
Zicheng Zhang, Wei Sun, Yingjie Zhou, Jun Jia, Zhichao Zhang, Jing Liu, Xiongkuo Min, and Guangtao Zhai.
\newblock Subjective and objective quality assessment for in-the-wild computer graphics images.
\newblock {\em ACM Transactions on Multimedia Computing, Communications and Applications}, 20(4):1--22, 2023.

\bibitem{dataset:webpage}
Chengyao Shen and Qi~Zhao.
\newblock Webpage saliency.
\newblock In {\em Computer Vision--ECCV 2014: 13th European Conference, Zurich, Switzerland, September 6-12, 2014, Proceedings, Part VII 13}, pages 33--46. Springer, 2014.

\bibitem{gen:dalle}
Aditya Ramesh, Prafulla Dhariwal, Alex Nichol, Casey Chu, and Mark Chen.
\newblock Hierarchical text-conditional image generation with clip latents, 2022.

\bibitem{gen:MJ}
David Holz.
\newblock Midjourney.
\newblock \url{https://www.midjourney.com}, 2023.

\bibitem{gen:Playground25}
Daiqing Li, Aleks Kamko, Ehsan Akhgari, Ali Sabet, Linmiao Xu, and Suhail Doshi.
\newblock Playground v2.5: Three insights towards enhancing aesthetic quality in text-to-image generation, 2024.

\bibitem{gen:pixart}
Junsong Chen, Jincheng Yu, Chongjian Ge, Lewei Yao, Enze Xie, Yue Wu, Zhongdao Wang, James Kwok, Ping Luo, Huchuan Lu, and Zhenguo Li.
\newblock Pixart-$\alpha$: Fast training of diffusion transformer for photorealistic text-to-image synthesis, 2023.

\bibitem{gen:xl}
Robin Rombach, Andreas Blattmann, and Björn Ommer.
\newblock Text-guided synthesis of artistic images with retrieval-augmented diffusion models, 2022.

\bibitem{gen:ssd-1b}
Yatharth Gupta, Vishnu~V. Jaddipal, Harish Prabhala, Sayak Paul, and Patrick~Von Platen.
\newblock Progressive knowledge distillation of stable diffusion xl using layer level loss, 2024.

\bibitem{database:aigiqa20k}
Chunyi Li, Tengchuan Kou, Yixuan Gao, Yuqin Cao, Wei Sun, Zicheng Zhang, Yingjie Zhou, Zhichao Zhang, Weixia Zhang, Haoning Wu, Xiaohong Liu, Xiongkuo Min, and Guangtao Zhai.
\newblock Aigiqa-20k: A large database for ai-generated image quality assessment, 2024.

\bibitem{add:ntire2024}
Xiaohong Liu, Xiongkuo Min, Guangtao Zhai, Chunyi Li, Tengchuan Kou, Wei Sun, Haoning Wu, Yixuan Gao, Yuqin Cao, Zicheng Zhang, Xiele Wu, Radu Timofte, et~al.
\newblock {NTIRE} 2024 quality assessment of {AI}-generated content challenge.
\newblock In {\em Proceedings of the IEEE/CVF Conference on Computer Vision and Pattern Recognition (CVPR) Workshops}, 2024.

\bibitem{add:track2}
Tengchuan Kou, Xiaohong Liu, Zicheng Zhang, Chunyi Li, Haoning Wu, Xiongkuo Min, Guangtao Zhai, and Ning Liu.
\newblock Subjective-aligned dataset and metric for text-to-video quality assessment, 2024.

\bibitem{add:g-refine}
Chunyi Li, Haoning Wu, Hongkun Hao, Zicheng Zhang, Tengchaun Kou, Chaofeng Chen, Lei Bai, Xiaohong Liu, Weisi Lin, and Guangtao Zhai.
\newblock G-refine: A general quality refiner for text-to-image generation, 2024.

\bibitem{add:q-refine}
Chunyi Li, Haoning Wu, Zicheng Zhang, Hongkun Hao, Kaiwei Zhang, Lei Bai, Xiaohong Liu, Xiongkuo Min, Weisi Lin, and Guangtao Zhai.
\newblock Q-refine: A perceptual quality refiner for ai-generated image, 2024.

\bibitem{gen:RealVis}
Civital.
\newblock Realvisxl-v4.0.
\newblock \url{https://realvis.art/}, 2024.

\bibitem{i2t:gpt4}
OpenAI.
\newblock Gpt-4 technical report, 2023.

\bibitem{i2t:llava}
Haotian Liu, Chunyuan Li, Qingyang Wu, and Yong~Jae Lee.
\newblock Visual instruction tuning, 2023.

\bibitem{i2t:mplugowl}
Qinghao Ye, Haiyang Xu, Guohai Xu, Jiabo Ye, Ming Yan, Yiyang Zhou, Junyang Wang, Anwen Hu, Pengcheng Shi, Yaya Shi, Chaoya Jiang, Chenliang Li, Yuanhong Xu, Hehong Chen, Junfeng Tian, Qian Qi, Ji~Zhang, and Fei Huang.
\newblock mplug-owl: Modularization empowers large language models with multimodality, 2023.

\bibitem{i2t:Qwen-VL}
Jinze Bai, Shuai Bai, Shusheng Yang, Shijie Wang, Sinan Tan, Peng Wang, Junyang Lin, Chang Zhou, and Jingren Zhou.
\newblock Qwen-vl: A versatile vision-language model for understanding, localization, text reading, and beyond, 2023.

\bibitem{i2t:sharegpt4v}
Lin Chen, Jisong Li, Xiaoyi Dong, Pan Zhang, Conghui He, Jiaqi Wang, Feng Zhao, and Dahua Lin.
\newblock Sharegpt4v: Improving large multi-modal models with better captions, 2023.

\bibitem{i2t:iblip}
Wenliang Dai, Junnan Li, Dongxu Li, Anthony Meng~Huat Tiong, Junqi Zhao, Weisheng Wang, Boyang Li, Pascale Fung, and Steven Hoi.
\newblock Instructblip: Towards general-purpose vision-language models with instruction tuning, 2023.

\bibitem{gen:animatediff}
Yuwei Guo, Ceyuan Yang, Anyi Rao, Zhengyang Liang, Yaohui Wang, Yu~Qiao, Maneesh Agrawala, Dahua Lin, and Bo~Dai.
\newblock Animatediff: Animate your personalized text-to-image diffusion models without specific tuning.
\newblock {\em International Conference on Learning Representations}, 2024.

\bibitem{gen:dream}
dreamlike art.
\newblock dreamlike-photoreal-2.0.
\newblock \url{https://dreamlike.art}, 2023.

\bibitem{gen:Playground20}
PlaygroundAI.
\newblock playground-v2-1024px-aesthetic.
\newblock \url{https://playground.com}, 2023.

\bibitem{gen:sd}
Robin Rombach, Andreas Blattmann, Dominik Lorenz, Patrick Esser, and Bj{\"o}rn Ommer.
\newblock High-resolution image synthesis with latent diffusion models.
\newblock In {\em Proceedings of the IEEE/CVF conference on computer vision and pattern recognition}, pages 10684--10695, 2022.

\bibitem{ir:diffbir}
Xinqi Lin, Jingwen He, Ziyan Chen, Zhaoyang Lyu, Bo~Dai, Fanghua Yu, Wanli Ouyang, Yu~Qiao, and Chao Dong.
\newblock Diffbir: Towards blind image restoration with generative diffusion prior, 2024.

\bibitem{ir:instructpix2pix}
Tim Brooks, Aleksander Holynski, and Alexei~A Efros.
\newblock Instructpix2pix: Learning to follow image editing instructions.
\newblock In {\em Proceedings of the IEEE/CVF Conference on Computer Vision and Pattern Recognition}, pages 18392--18402, 2023.

\bibitem{ir:pasd}
Tao Yang, Rongyuan Wu, Peiran Ren, Xuansong Xie, and Lei Zhang.
\newblock Pixel-aware stable diffusion for realistic image super-resolution and personalized stylization, 2024.

\bibitem{ir:stablesr}
Jianyi Wang, Zongsheng Yue, Shangchen Zhou, Kelvin C.~K. Chan, and Chen~Change Loy.
\newblock Exploiting diffusion prior for real-world image super-resolution, 2023.

\bibitem{quality:topiq}
Chaofeng Chen, Jiadi Mo, Jingwen Hou, Haoning Wu, Liang Liao, Wenxiu Sun, Qiong Yan, and Weisi Lin.
\newblock Topiq: A top-down approach from semantics to distortions for image quality assessment.
\newblock {\em IEEE Transactions on Image Processing}, 33:2404--2418, 2024.

\bibitem{quality:ahiq}
Shanshan Lao, Yuan Gong, Shuwei Shi, Sidi Yang, Tianhe Wu, Jiahao Wang, Weihao Xia, and Yujiu Yang.
\newblock Attentions help cnns see better: Attention-based hybrid image quality assessment network, 2022.

\bibitem{quality:dists}
Keyan Ding, Kede Ma, Shiqi Wang, and Eero~P Simoncelli.
\newblock Image quality assessment: Unifying structure and texture similarity.
\newblock {\em IEEE transactions on pattern analysis and machine intelligence}, 44(5):2567--2581, 2020.

\bibitem{quality:lpips}
Richard Zhang, Phillip Isola, Alexei~A Efros, Eli Shechtman, and Oliver Wang.
\newblock The unreasonable effectiveness of deep features as a perceptual metric.
\newblock In {\em Proceedings of the IEEE conference on computer vision and pattern recognition}, pages 586--595, 2018.

\bibitem{quality:pieapp}
Ekta Prashnani, Hong Cai, Yasamin Mostofi, and Pradeep Sen.
\newblock Pieapp: Perceptual image-error assessment through pairwise preference.
\newblock In {\em Proceedings of the IEEE Conference on Computer Vision and Pattern Recognition}, pages 1808--1817, 2018.

\bibitem{quality:CLIPIQA}
Jianyi Wang, Kelvin~CK Chan, and Chen~Change Loy.
\newblock Exploring clip for assessing the look and feel of images.
\newblock In {\em Proceedings of the AAAI Conference on Artificial Intelligence}, volume~37, pages 2555--2563, 2023.

\bibitem{quality:CNNIQA}
Le~Kang, Peng Ye, Yi~Li, and David Doermann.
\newblock Convolutional neural networks for no-reference image quality assessment.
\newblock In {\em Proceedings of the IEEE conference on computer vision and pattern recognition}, pages 1733--1740, 2014.

\bibitem{quality:DBCNN}
Weixia Zhang, Kede Ma, Jia Yan, Dexiang Deng, and Zhou Wang.
\newblock Blind image quality assessment using a deep bilinear convolutional neural network.
\newblock {\em IEEE Transactions on Circuits and Systems for Video Technology}, 30(1):36--47, 2018.

\bibitem{quality:HyperIQA}
Shaolin Su, Qingsen Yan, Yu~Zhu, Cheng Zhang, Xin Ge, Jinqiu Sun, and Yanning Zhang.
\newblock Blindly assess image quality in the wild guided by a self-adaptive hyper network.
\newblock In {\em Proceedings of the IEEE/CVF Conference on Computer Vision and Pattern Recognition}, pages 3667--3676, 2020.

\bibitem{metric:tra-cdc}
Ruihan Yang and Stephan Mandt.
\newblock Lossy image compression with conditional diffusion models.
\newblock In A.~Oh, T.~Naumann, A.~Globerson, K.~Saenko, M.~Hardt, and S.~Levine, editors, {\em Advances in Neural Information Processing Systems}, volume~36, pages 64971--64995. Curran Associates, Inc., 2023.

\bibitem{quality:clipsim}
Alec Radford, Jong~Wook Kim, Chris Hallacy, Aditya Ramesh, Gabriel Goh, Sandhini Agarwal, Girish Sastry, Amanda Askell, Pamela Mishkin, Jack Clark, et~al.
\newblock Learning transferable visual models from natural language supervision.
\newblock In {\em International conference on machine learning}, pages 8748--8763. PMLR, 2021.

\bibitem{quality:ssim}
Zhou Wang.
\newblock Image quality assessment: from error visibility to structural similarity.
\newblock {\em IEEE transactions on image processing}, 13(4):600--612, 2004.

\bibitem{quality:LIQE}
Weixia Zhang, Guangtao Zhai, Ying Wei, Xiaokang Yang, and Kede Ma.
\newblock Blind image quality assessment via vision-language correspondence: A multitask learning perspective.
\newblock In {\em Proceedings of the IEEE/CVF Conference on Computer Vision and Pattern Recognition}, pages 14071--14081, 2023.

\bibitem{quality:fid}
Martin Heusel, Hubert Ramsauer, Thomas Unterthiner, Bernhard Nessler, and Sepp Hochreiter.
\newblock Gans trained by a two time-scale update rule converge to a local nash equilibrium.
\newblock {\em Advances in neural information processing systems}, 30, 2017.

\bibitem{other:itu}
I.~T. Union.
\newblock Methodology for the subjective assessment of the quality of television pictures.
\newblock {\em ITU-R Recommendation BT. 500-11}, 2002.

\end{thebibliography}
}

\newpage
\appendix

\renewcommand {\thetable} {A\arabic{table}}
\renewcommand {\thefigure} {A\arabic{figure}}
\setcounter{table}{0}
\setcounter{figure}{0}

\section{Appendix}

In this section, we briefly describe the content of the checklist requirements. Considering that our experiments tried a variety of parameter configurations, the conclusions under different configurations are also stated here, including specific ablation data.

\subsection{Limitations and Social Impact}
\label{app:limitation}

\textbf{Limitation 1}: Although we have considered most of the mainstream I2T and T2I models in CMC-Bench (till March 2024), the number of models is still insufficient to fully characterize the performance of all current LMMs on CMC. Taking the open-source T2 model as an example, more than 20,000 models have been released on huggingface (till May 2024). Although we cannot run all models, the capabilities of some relatively unpopular or more advanced LMMs in the future need to be further updated on CMC-Bench.

\textbf{Limitation 2}: CMC-Bench is currently designed for the performance verification of image compression, not video compression. Considering that the temporal information of videos is relatively complex, the current LMMs are only applicable to image compression, which makes it difficult to ensure consistency with the reference when generating videos. However, as LMMs gradually apply to video compression in the future, CMC-Bench will also be evaluated at the video level.

\textbf{Social Impact}: Through the CMC paradigm, the size of the image can be compressed by 1,000 times, and even 10,000 times in extreme cases. This will effectively promote image communication between a large number of terminals under limited bandwidth, thereby realizing multi-device collaboration in the Internet of Things and semantic communication. Considering that traditional codecs have encountered bottlenecks after three decades of development and the compression rate is gradually approaching the Shannon limit, we believe that LMM will effectively achieve semantic-level compression and thus become the future evolution direction of visual information codec protocol.

\subsection{Subjective Annotation Settings}
\label{app:subjective}

\begin{figure}[tbph]
    \centering
    \includegraphics[width=0.8\linewidth]{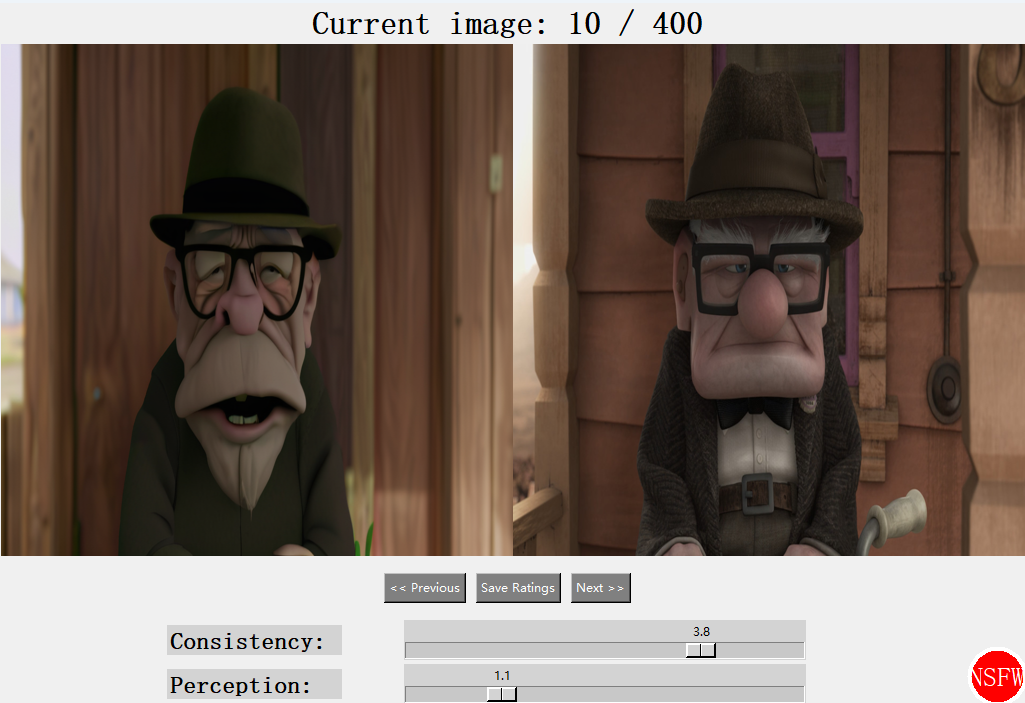}
    \caption{Subjective annotation interface presenting with the
 distorted (left) and reference (right) image. Each viewer is asked to provide (1) a Consistency score between two images from 0 to 5; (2) a Perception score of the distorted image; (3) an NSFW flag when they feel offended.}
    \label{fig:interface}
\end{figure}

Compliant with the ITU-R BT.500-13 \cite{other:itu} standard, we invited 20 viewers (11 male, 9 female) in this subjective experiment with normal lighting levels. Images are presented on the iMac display together with the ground truth in random order on the screen, with a resolution of up to 4096 $\times$ 2304. Both ground truth and distorted images are accessible for subjective. Considering the consistency between the reference and distorted image, and the perceptual quality of the only distorted image, subjects were asked to give two scores within the range of [0, 5], where each one-point interval stands for poor, bad, fair, good, or excellent quality. The user interface is shown in Figure \ref{fig:interface}.

Each user, in accordance with the Helsinki Declaration, provides informed consent for their data to be used in experiments. To prevent NSFW content, we implement three preventive measures: (1) Conduct a thorough manual screening of the ground truth; (2) Utilize the SD safety checker \cite{gen:sd} during decoding; (3) Incorporate an `offensive' flag in the annotation process, allowing viewers to report NSFW content if encountered. The data confirms that the ground truth is safe, with approximately 0.2\% of distorted images receiving reports, which is generally acceptable.

\begin{figure}[t]
    \centering
    \vspace{-2pt}
    \includegraphics[width=0.8\linewidth]{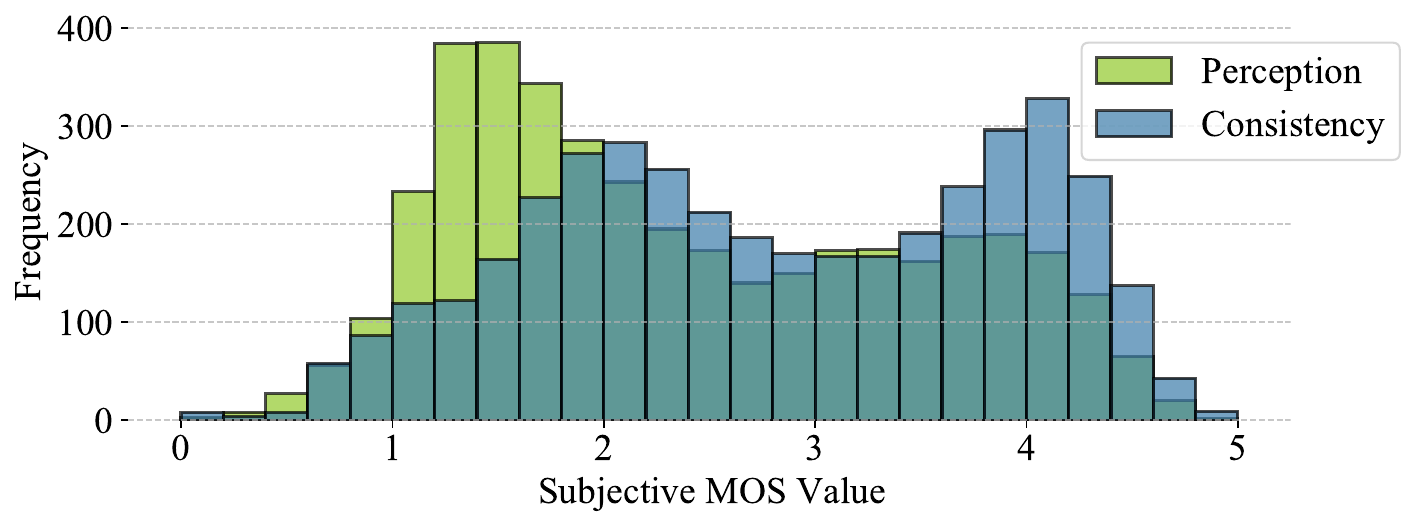}
    \vspace{-2pt}
    \caption{Mean Opinion Score distribution of the Consistency and the Perception dimension.}
    \vspace{-4mm}
    \label{fig:subjective-mos}
\end{figure}

In case of visual fatigue, we split the database into $g \in [0,10]$ groups including $M=400$ images each, while limiting the experiment time to an hour. After collecting every viewer's quality ratings, we compute the Spearman Rank-order Correlation Coefficient (SRoCC) between them and the global average and remove the outliers with SRoCC lower than 0.6. Then we normalize the average score $s$ for between each session to avoid inter-session scoring differences as:
\begin{equation}
s_{ij}(g) = r_{ij}(g) - \frac{1}{M}\sum_{i=0}^{g\cdot M -1} r_{ij} + 2.5,
\end{equation}
where $(i,j)$ represent the index of the image and viewer and $r$ stands for raw score. We observed a fairly even distribution of subjective scores on both dimensions and bar graphs for each score range are provided in the Supplementary. Then subjective scores are converted to Z-scores $z_{ij}$ by:
\begin{equation}
z_{ij}=\frac{s_{ij}-\mu_j}{\sigma_j},
\end{equation}
where $\mu_j=\frac{1}{N}\sum_{i=0}^{N-1} s_{ij}$, $\sigma_j=\sqrt{\frac{1}{N-1}\sum_{i=0}^{N-1}(s_{ij}-\mu_i)^2}$ and $N=10$ is the number of subjects, which is finally reported as MOS, namely golden user annotations. The distribution of Consistency and the Perception MOS is shown in Figure \ref{fig:subjective-mos}, which proves that extremely low and high scores are rare, and most scores are between 1 and 4. The Perception score is concentrated in the medium-low area, while the Consistency dimension tends to be moderately high.

\subsection{Experimental Platform}
\label{app:resource}

For 4,000 labeled image pairs, we trained five FR and five NR quality indicators for 50 epochs using Adam optimizer on a local NVIDIA GeForce RTX 4090. Among which 80/20 for training/testing. We take MSE loss with a learning rate at $2\times10^{-5}$. The TOPIQ-FR and TOPIQ-NR are set as objective indicators for Consistency/Perception. Noted these 4,000 training data images are not included in the source data for objective evaluation for a fair comparison.

The LMMs are validated on a server with four NVIDIA RTX A6000, using I2T in different output lengths, and T2I in different strengths, combining through \textit{Text}/\textit{Pixel}/\textit{Image}/\textit{Full} modes.

\subsection{Image-to-Text Model Configuration}
\label{app:i2t}

Towards different output lengths, we applied different prompts as the input for I2T models. The prompt follows previous CMC templates \cite{relate:cmc-misc}, with length [5,10,20,50]. For example, to describe Figure \ref{fig:example-nsi} in different lengths, the input prompt format and output text from GPT-4o \cite{i2t:gpt4} are:

\begin{wrapfigure}{r}{0.42\textwidth}
\centering
\vspace{-5mm}
\includegraphics[width=0.98\linewidth]{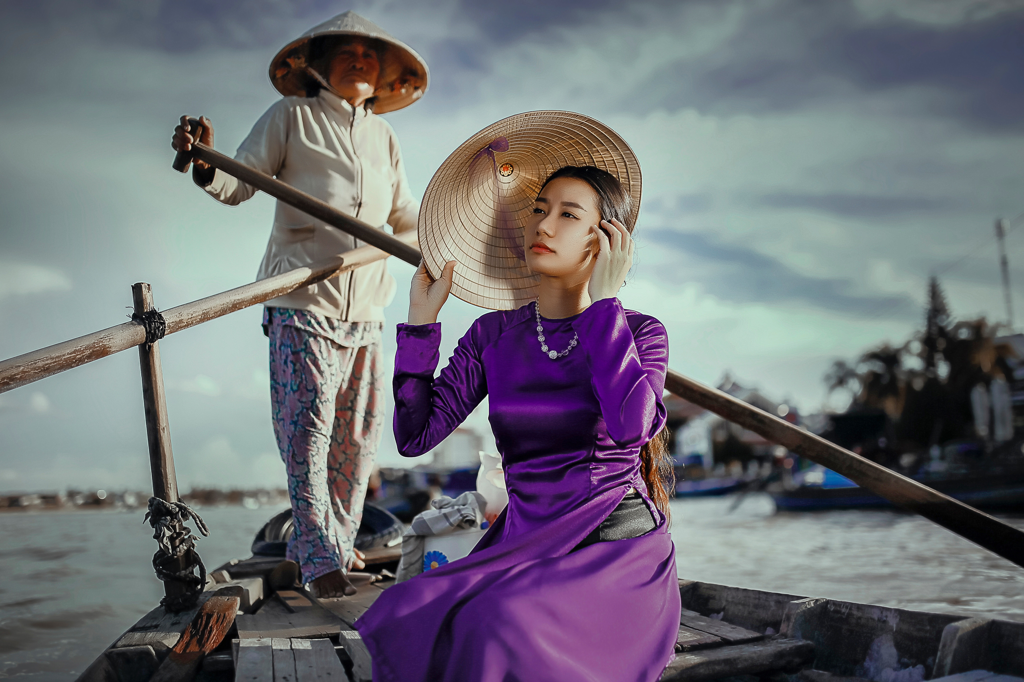}
\caption{Example ground truth.}
\label{fig:example-nsi}
\vspace{-10mm}
\end{wrapfigure}

Q(5): \textit{`Generate an informative paragraph in 5 words based on the image.'}
\\
A(5): \textit{`Traditional attire on serene river.'}

Q(10): \textit{`Generate an informative paragraph in 10 words based on the image. Considering Light, Color.'}
\\
A(10): \textit{`Two individuals navigate a wooden boat in traditional attire, serene.'}

Q(20): \textit{`Generate an informative paragraph in 20 words based on the image. Considering Light, Color, Clarity.'}
\\
A(20): \textit{`A woman in traditional attire and a conical hat, seated on a wooden boat, with another person rowing on water.'}

Q(50): \textit{`Generate an informative paragraph in 50 words based on the image. Considering Light, Color, Clarity, Dense Caption, and Region Semantic.'}
\\
A(50): \textit{`A woman in a purple dress sits near the front of a boat, her hand adjusting a straw hat. Behind her, a person in a beige top and floral pants stands near the boat's wooden oars. The setting is a calm, expansive body of water under a cloudy sky.'}

\begin{figure}[t]
    \centering
    \subfigure{\includegraphics[width=0.32\textwidth]{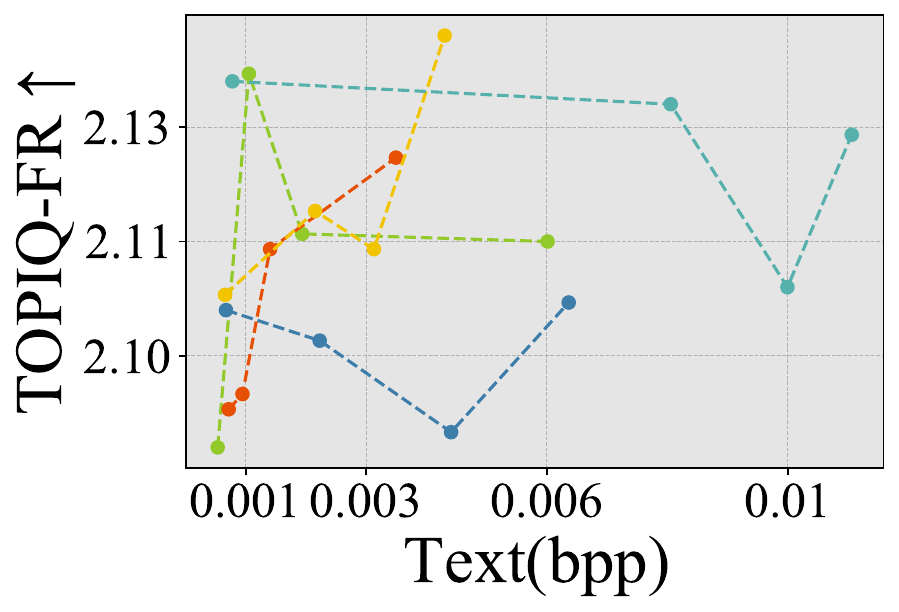}
    }\vspace{-1.5mm}
    \subfigure{\includegraphics[width=0.32\textwidth]{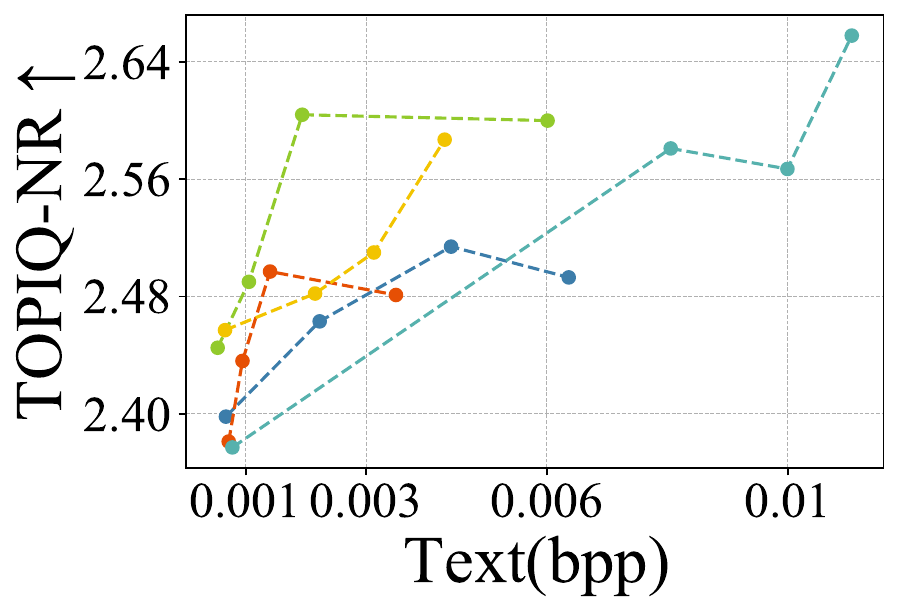}
    }\vspace{-1.5mm}
    \subfigure{\includegraphics[width=0.32\textwidth]{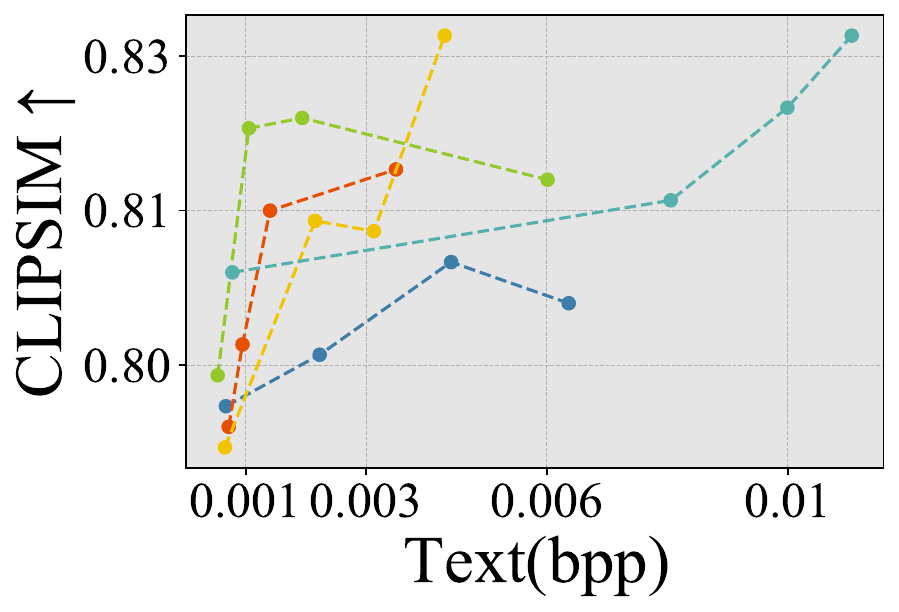}
    }\vspace{-1.5mm}
    \subfigure{\includegraphics[width=0.32\textwidth]{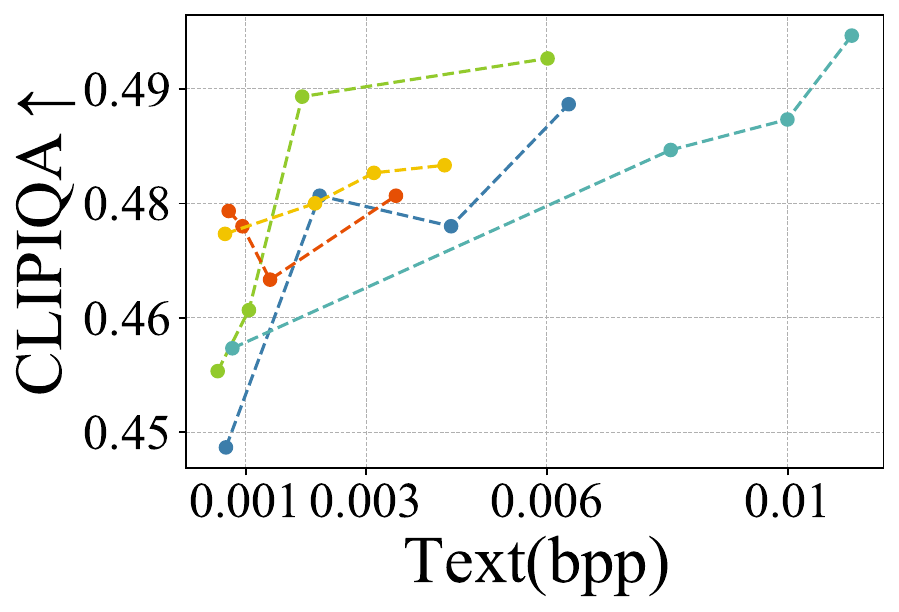}
    }\vspace{-1.5mm}
    \subfigure{\includegraphics[width=0.32\textwidth]{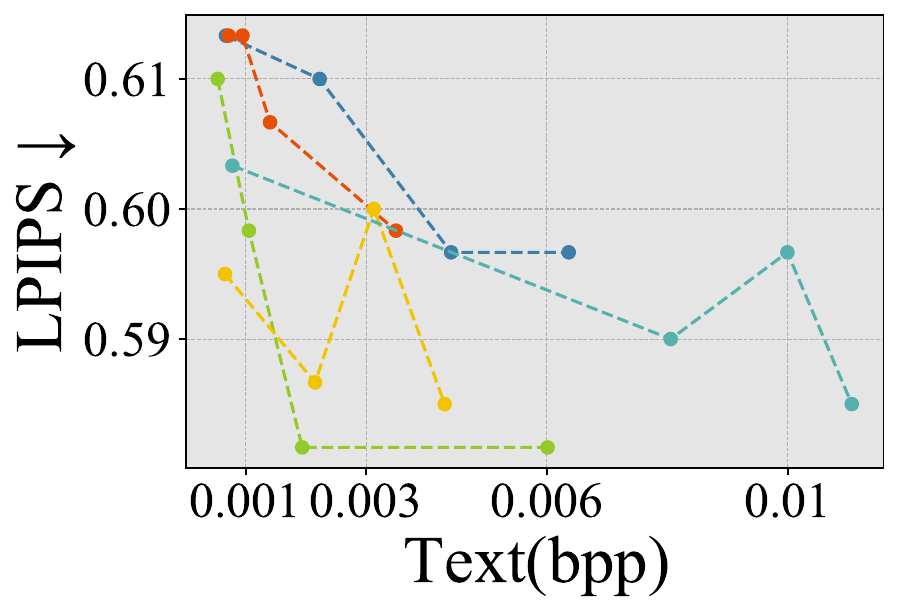}
    }\vspace{-1.5mm}
    \subfigure{\includegraphics[width=0.32\textwidth]{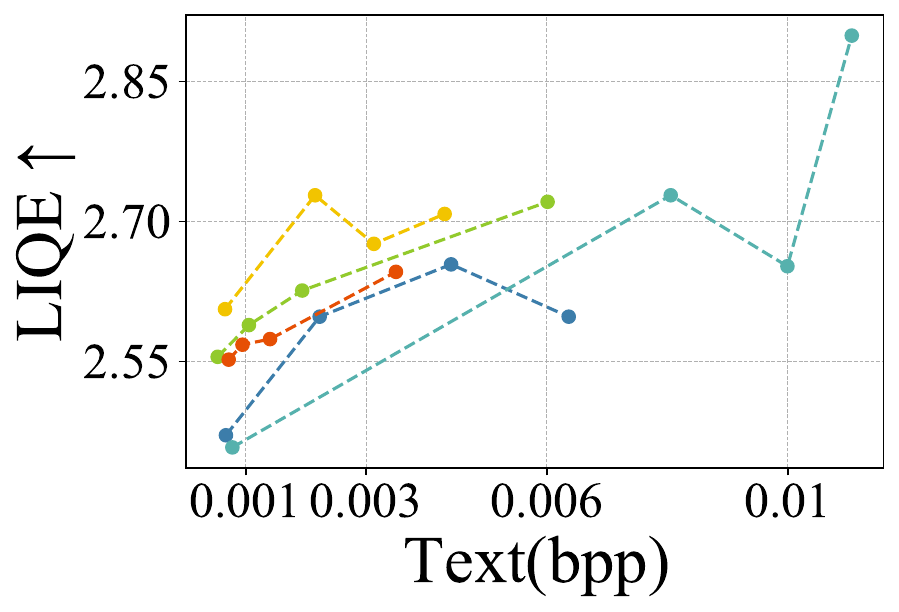}
    }\vspace{-1.5mm} 
    \subfigure{\includegraphics[width=0.32\textwidth]{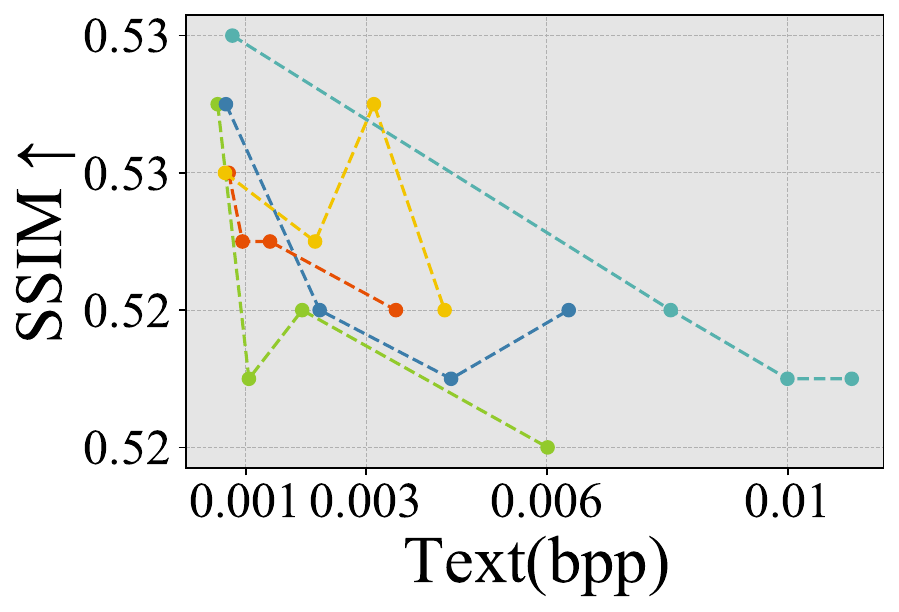}
    }\vspace{-1.5mm} 
    \subfigure{\includegraphics[width=0.32\textwidth]{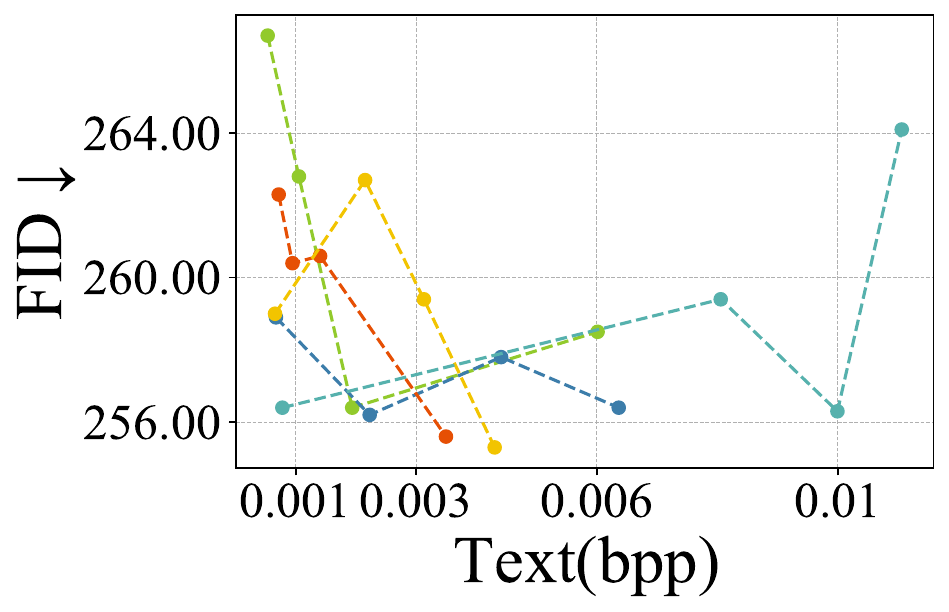}
    }\vspace{-1.5mm}  
    \subfigure{
    \centering
    \begin{minipage}{0.32\linewidth}
        \centering
        \vspace{-30mm}
        \hspace{4mm}
        \includegraphics[width=0.7\textwidth]{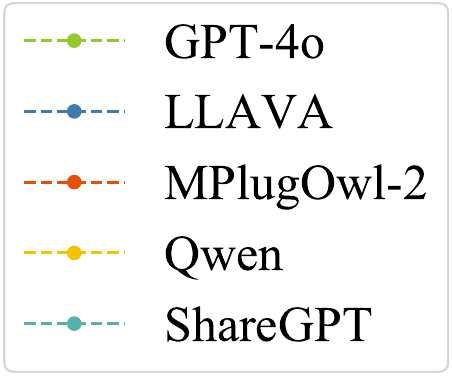}
    \end{minipage}
    
    }
    \vspace{-4pt}
    \caption{Setting I2T model with different output lengths for CMC task in \textit{Full} mode, evaluated by 4
consistency and 4 perception metrics indicated by \colorbox{lightgray!60!white}{marked} and plain background. A moderate output length can realize a satisfying performance while saving bitrate.}
    \vspace{-4mm}
    \label{fig:i2t-valid}
\end{figure}

To explore how much information the above four output lengths can represent. We use different I2T models, and combine them with the most effective T2I model (RealVis \cite{gen:RealVis}) under the above four output lengths, and use four Consistency and four Perception indicators for analysis, as shown in Figure \ref{fig:i2t-valid}, where the four datapoints of each curve represent four output lengths. The experimental results show that most I2T models can dynamically adjust the output length, except that InstructBLIP for image annotation cannot input prompt, and ShareGPT is not sensitive to the specified output length. Overall, when inputting Q(5), the reconstruction effect is relatively poor because of the short output; when inputting Q(50), the overly long paragraph from the I2T model cannot be understood by the T2I model, so the performance is not significantly improved while wasting bitrate. By observing the trend of all curves, we find that when the bpp of the text is between 0.002-0.003, the balance between performance and bitrate can be achieved. Therefore, for each model, we choose the output length closest to this bitrate, that is, Q(20) for GPT-4o \cite{i2t:gpt4}, MPlugOwl-2 \cite{i2t:mplugowl}; Q(10) for LLAVA \cite{i2t:llava}, Qwen \cite{i2t:Qwen-VL}, ShareGPT \cite{i2t:sharegpt4v}; and the default length for InstructBLIP \cite{i2t:iblip}.

\subsection{Text-to-Image Model Configuration}
\label{app:t2i}

In \textit{Text} mode, since there is no reference image as a starting point, the denoising strength is undisputedly 1. In the other three modes, we adjusted different intensities with a granularity of 0.1. For \textit{Full} and \textit{Image} modes that provide a reference image, a high denoising strength will waste the information of this reference, so we verified the performance under strength from 0.2 to 0.8; for \textit{Pixel} mode, since the pixel provides less information than the compressed image, we increased the strength and range from 0.4 to 0.99 (as strength=1 will ignore the reference). 

\begin{figure}[t]
    \centering
    \subfigure{\includegraphics[width=0.32\textwidth]{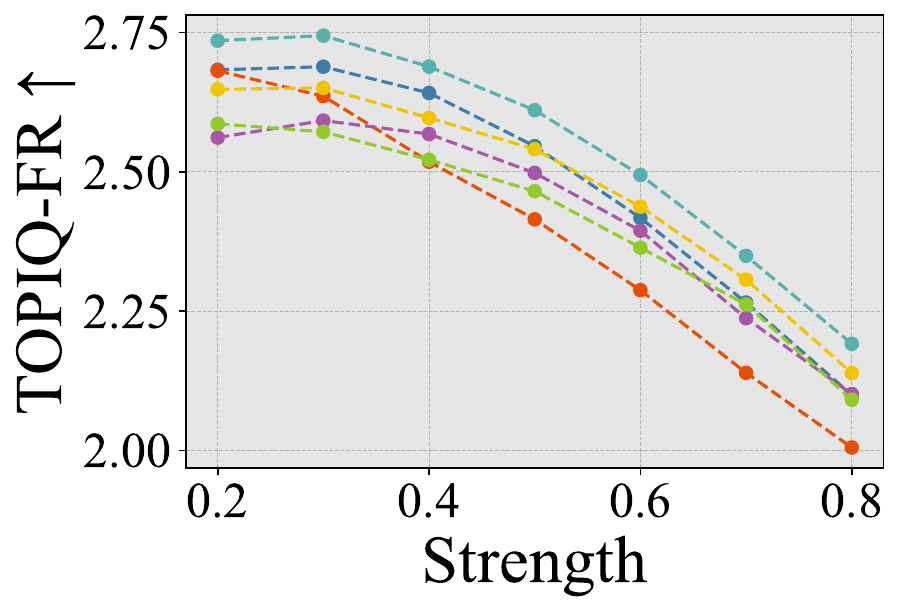}
    }\vspace{-1.5mm}
    \subfigure{\includegraphics[width=0.32\textwidth]{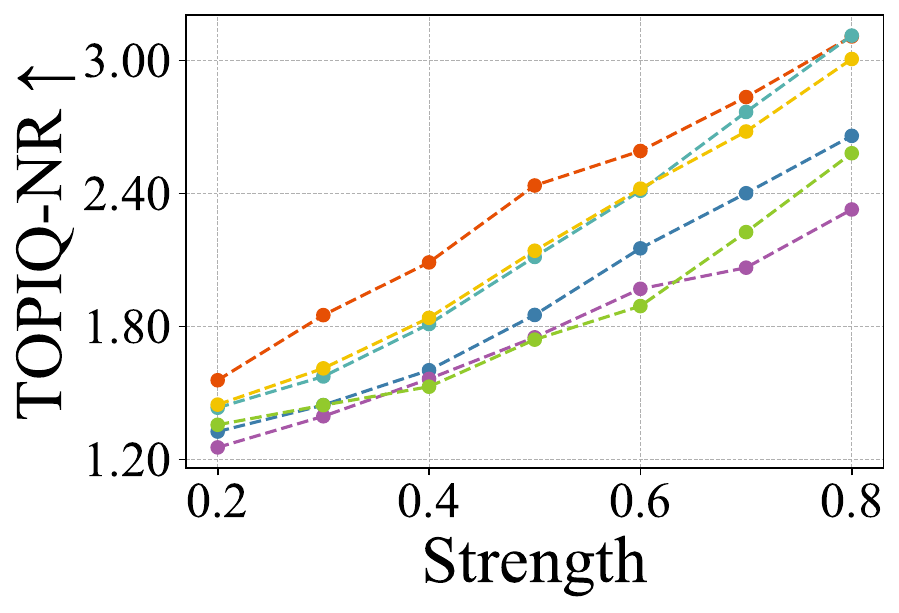}
    }\vspace{-1.5mm}
    \subfigure{\includegraphics[width=0.32\textwidth]{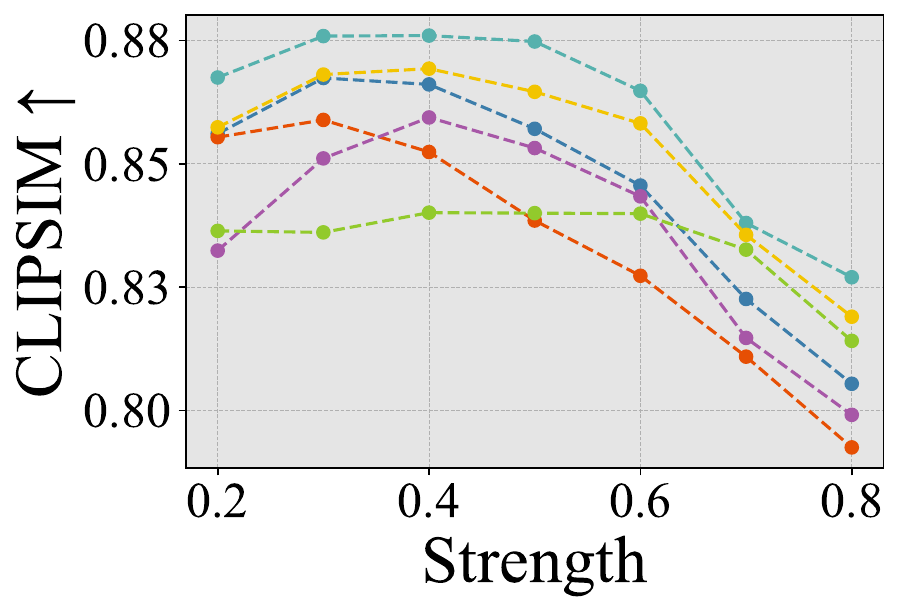}
    }\vspace{-1.5mm}
    \subfigure{\includegraphics[width=0.32\textwidth]{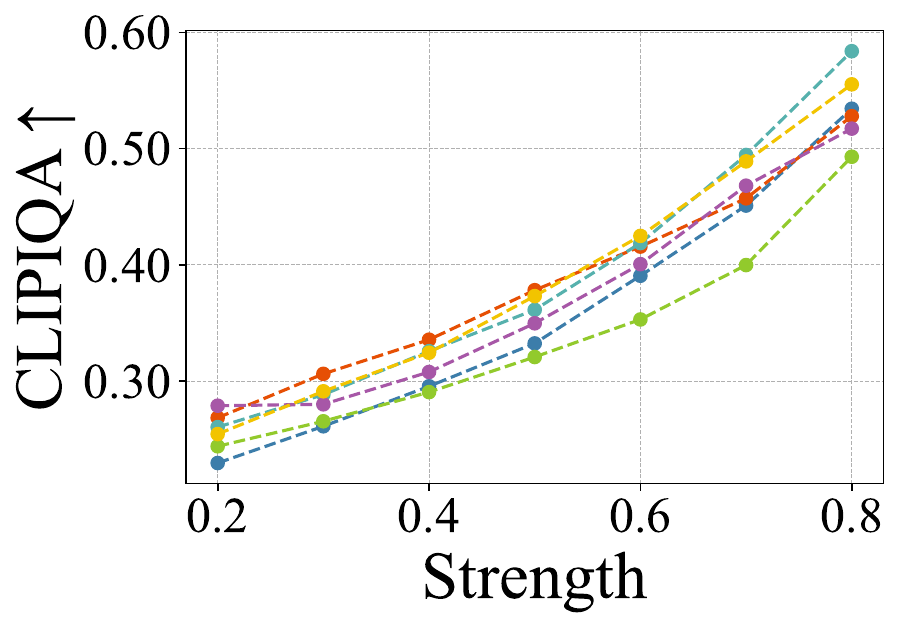}
    }\vspace{-1.5mm}
    \subfigure{\includegraphics[width=0.32\textwidth]{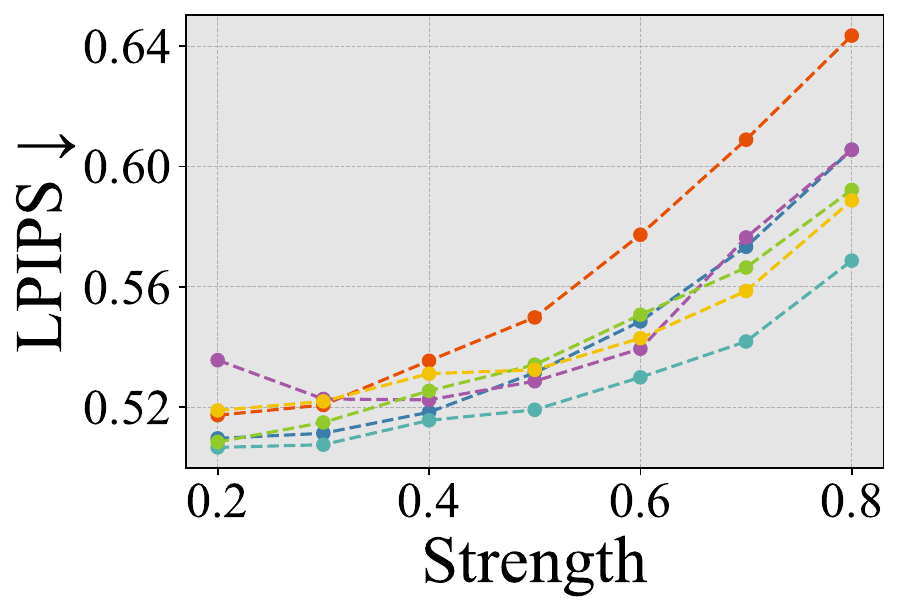}
    }\vspace{-1.5mm}
    \subfigure{\includegraphics[width=0.32\textwidth]{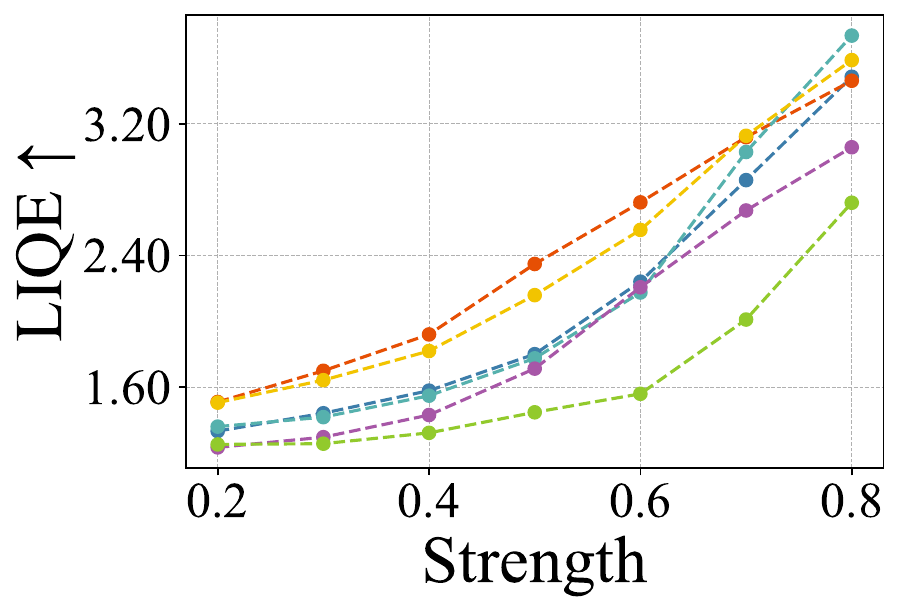}
    }\vspace{-1.5mm} 
    \subfigure{\includegraphics[width=0.32\textwidth]{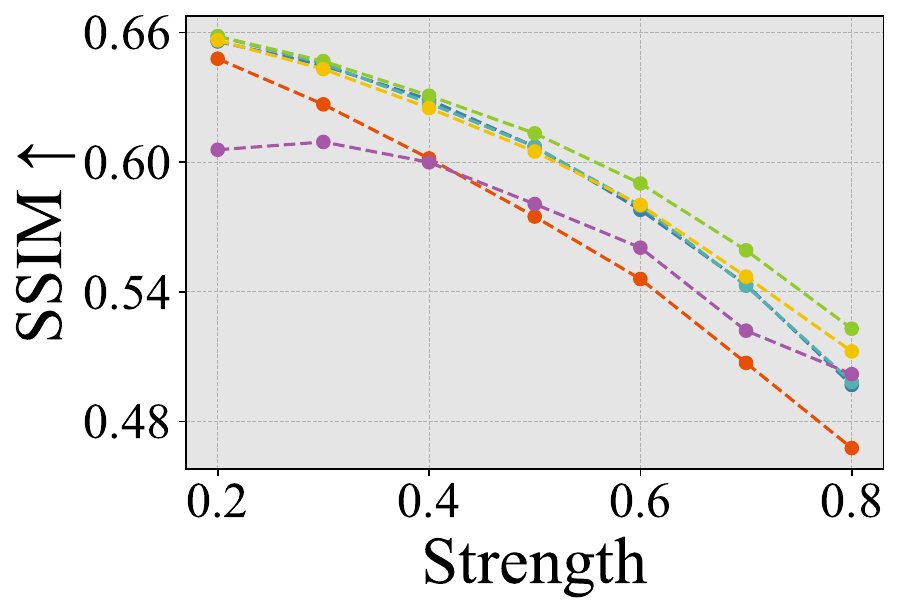}
    }\vspace{-1.5mm} 
    \subfigure{\includegraphics[width=0.32\textwidth]{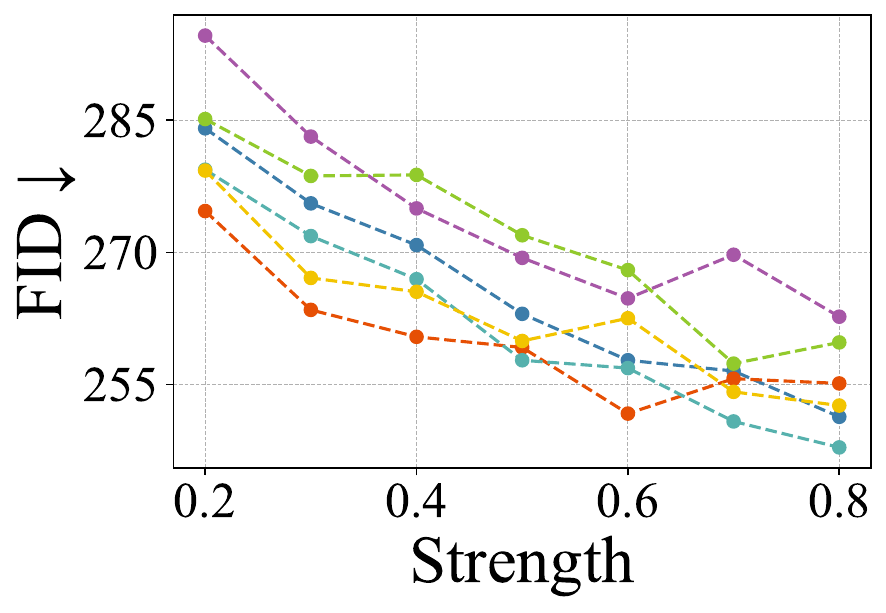}
    }\vspace{-1.5mm}  
    \subfigure{
    \centering
    \begin{minipage}{0.32\linewidth}
        \centering
        \vspace{-28mm}
        \hspace{4mm}
        \includegraphics[width=0.58\textwidth]{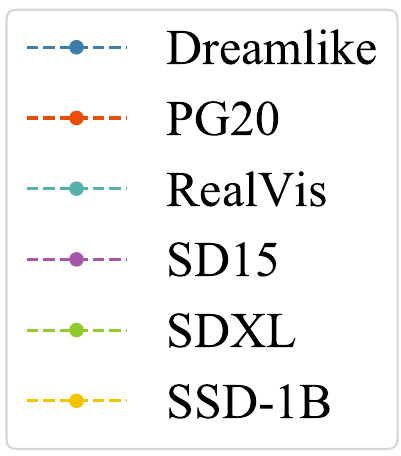}
    \end{minipage}
    
    }
    \vspace{-8pt}
    \caption{Setting T2I model with different strength for CMC task in \textit{Full} mode, evaluated by 4
consistency and 4 perception metrics indicated by \colorbox{lightgray!60!white}{marked} and plain background. A strength of 0.5 can reach a balance between consistency and perception.}
    \label{fig:t2i-valid-full}
\end{figure}

\begin{figure}[t]
    \centering
    \subfigure{\includegraphics[width=0.32\textwidth]{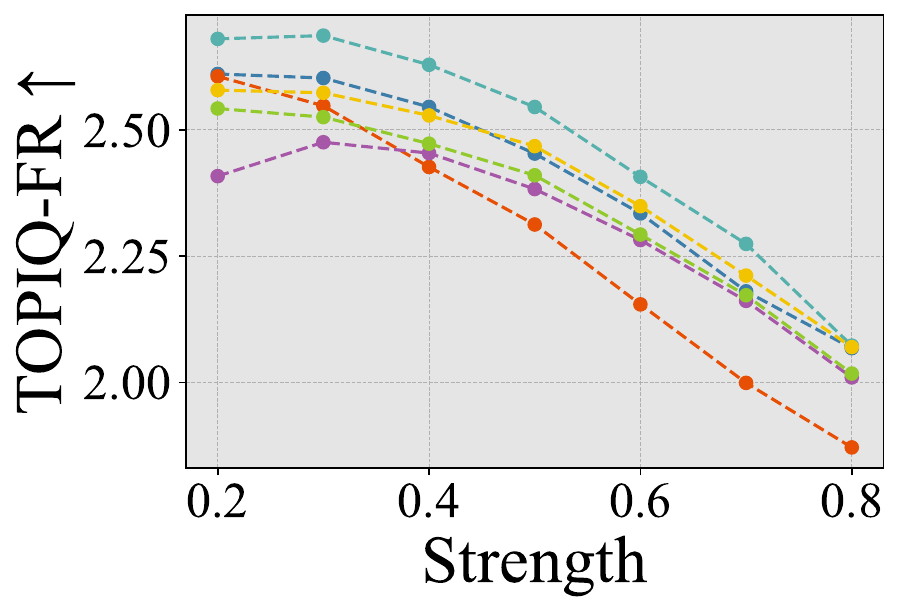}
    }\vspace{-1.5mm}
    \subfigure{\includegraphics[width=0.32\textwidth]{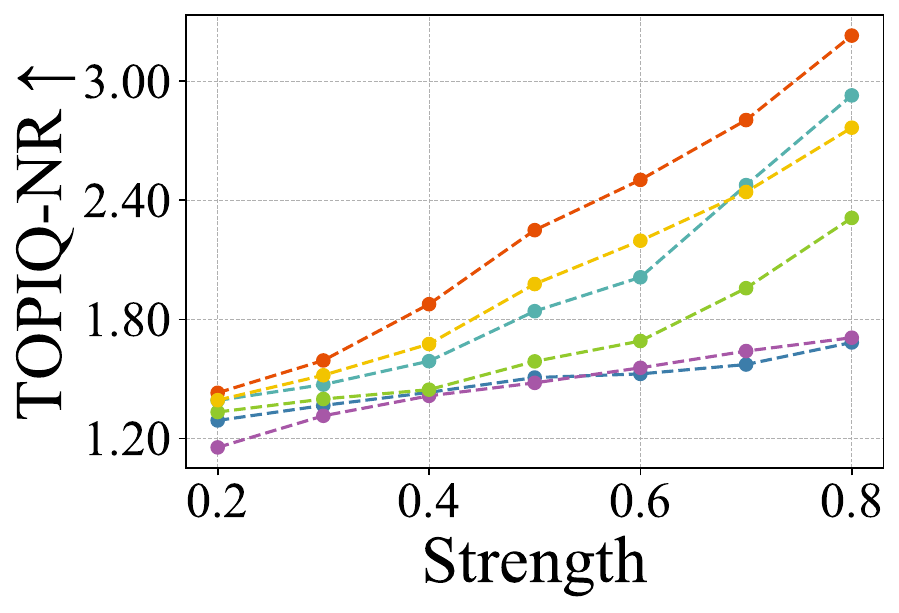}
    }\vspace{-1.5mm}
    \subfigure{\includegraphics[width=0.32\textwidth]{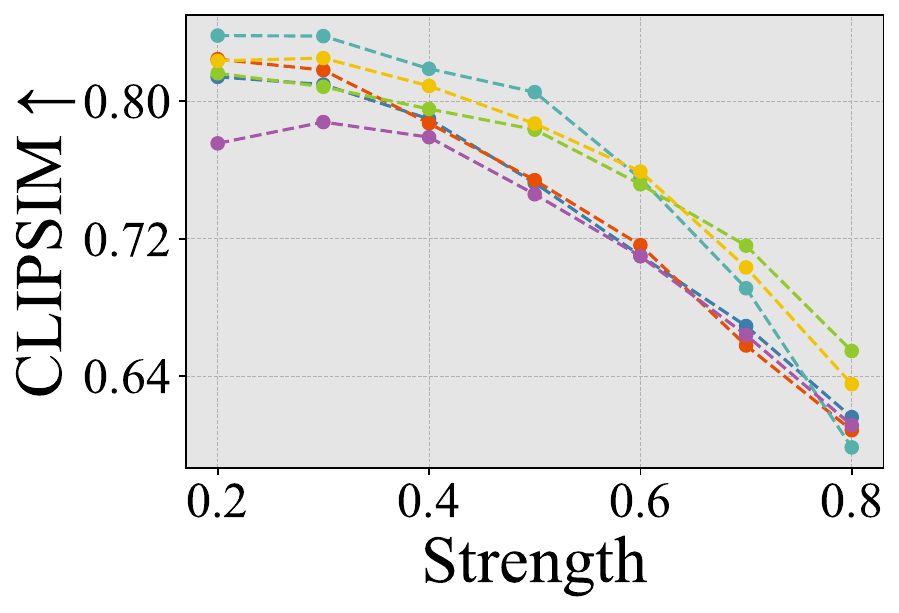}
    }\vspace{-1.5mm}
    \subfigure{\includegraphics[width=0.32\textwidth]{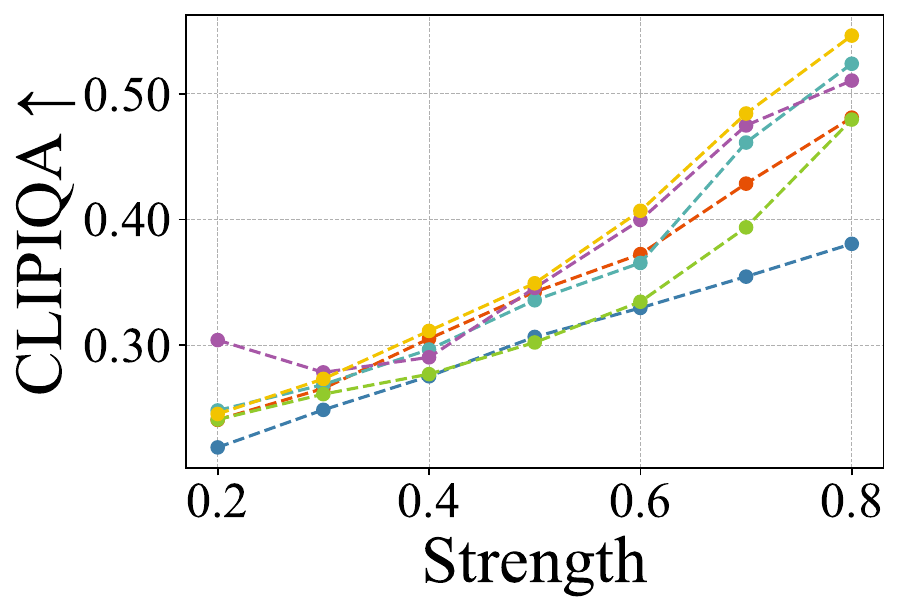}
    }\vspace{-1.5mm}
    \subfigure{\includegraphics[width=0.32\textwidth]{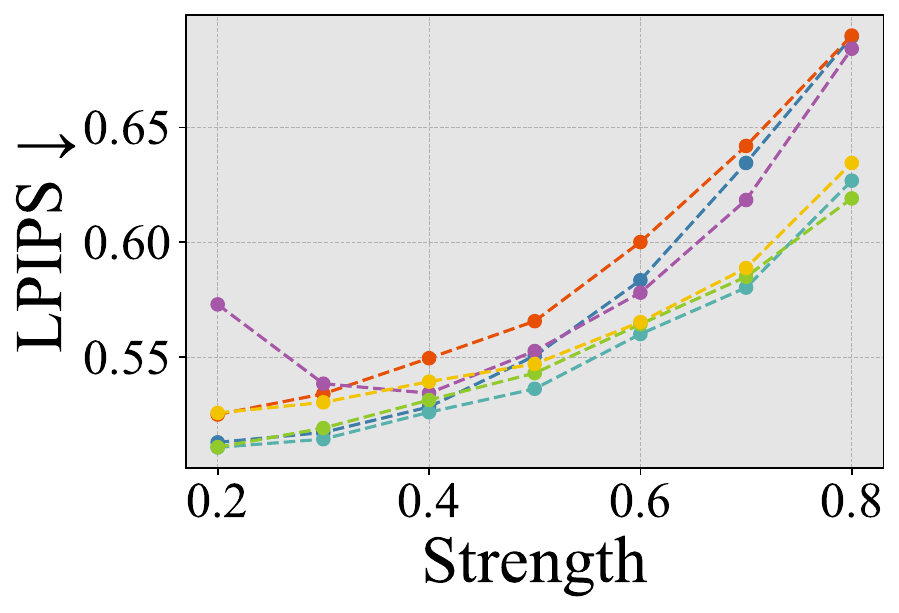}
    }\vspace{-1.5mm}
    \subfigure{\includegraphics[width=0.32\textwidth]{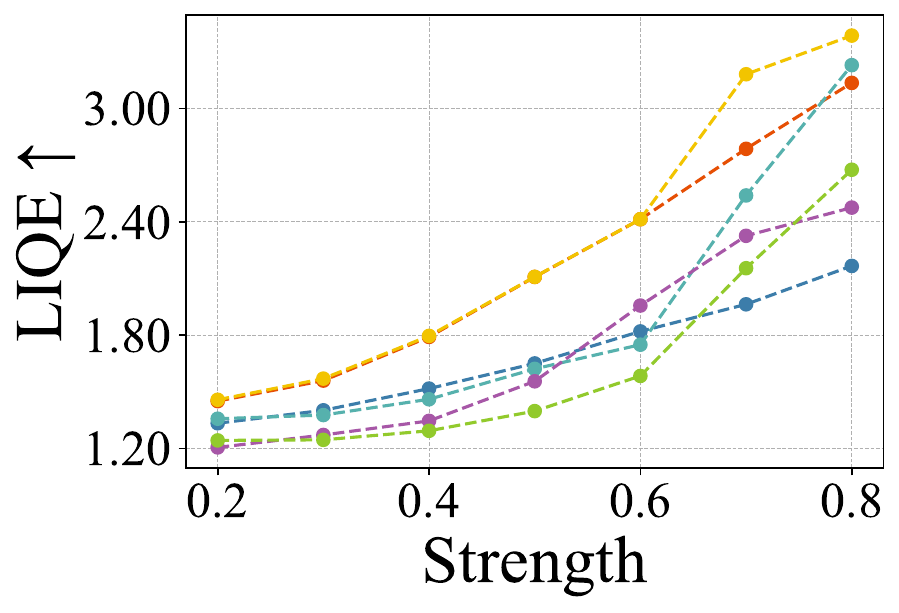}
    }\vspace{-1.5mm} 
    \subfigure{\includegraphics[width=0.32\textwidth]{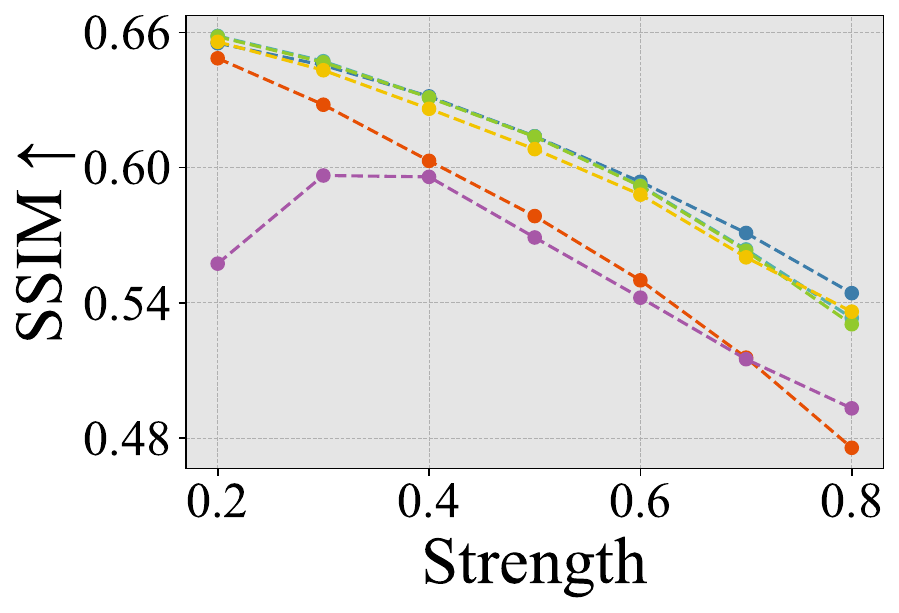}
    }\vspace{-1.5mm} 
    \subfigure{\includegraphics[width=0.32\textwidth]{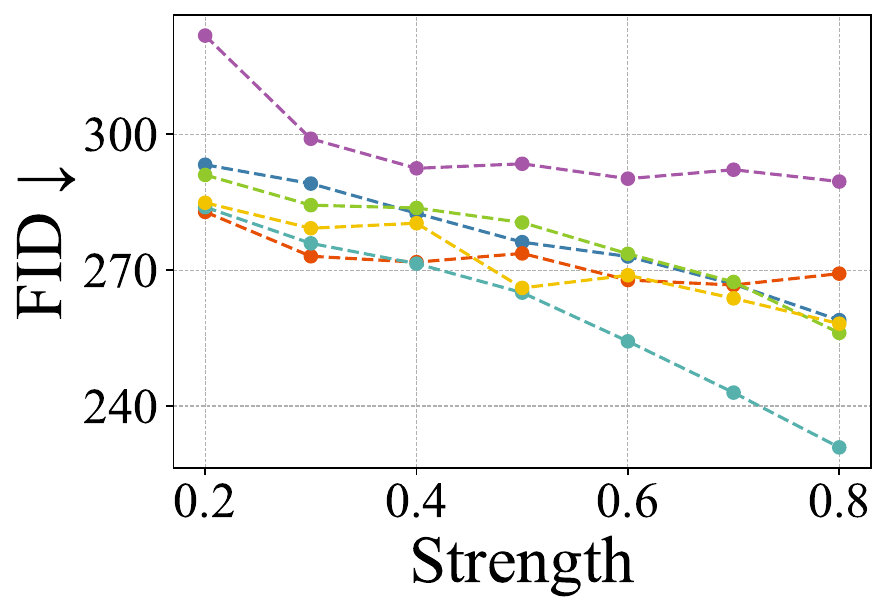}
    }\vspace{-1.5mm}  
    \subfigure{
    \centering
    \begin{minipage}{0.32\linewidth}
        \centering
        \vspace{-28mm}
        \hspace{4mm}
        \includegraphics[width=0.58\textwidth]{fig-supp/t2i-legend.pdf}
    \end{minipage}
    
    }
    \vspace{-8pt}
    \caption{Setting T2I model with different strength for CMC task in \textit{Image} mode, evaluated by 4
consistency and 4 perception metrics indicated by \colorbox{lightgray!60!white}{marked} and plain background. A strength of 0.5 can reach a balance between consistency and perception.}

    \label{fig:t2i-valid-image}
\end{figure}

\begin{figure}[t]
    \centering
    \subfigure{\includegraphics[width=0.32\textwidth]{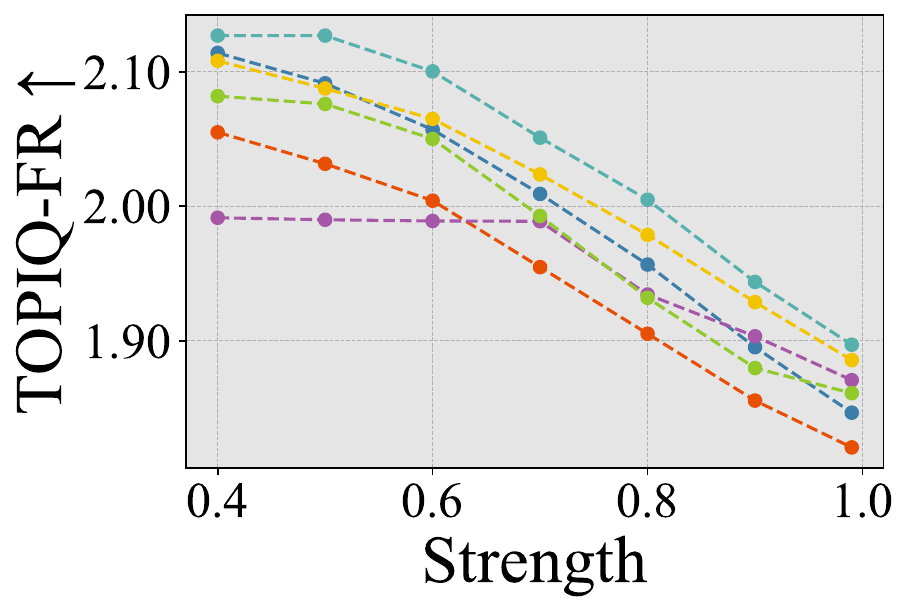}
    }\vspace{-1.5mm}
    \subfigure{\includegraphics[width=0.32\textwidth]{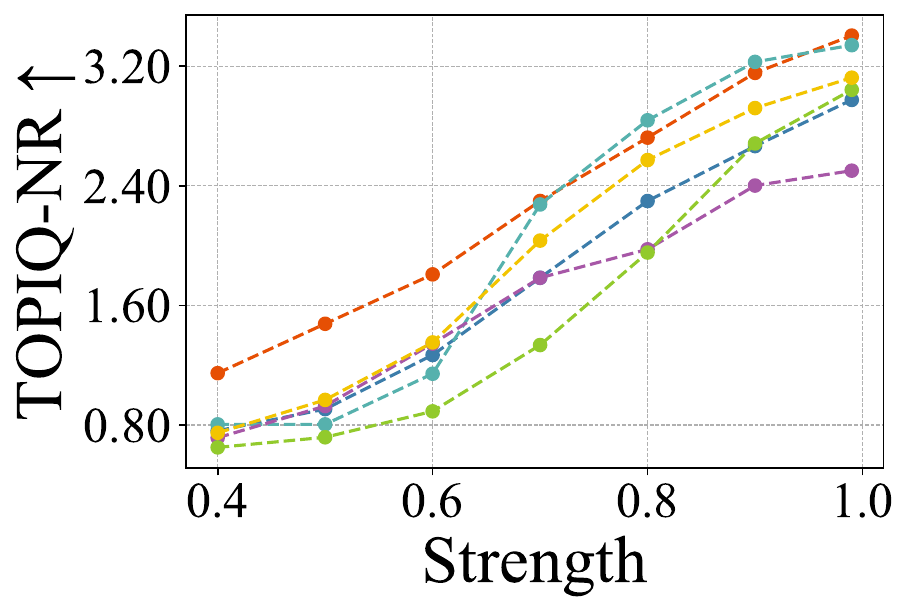}
    }\vspace{-1.5mm}
    \subfigure{\includegraphics[width=0.32\textwidth]{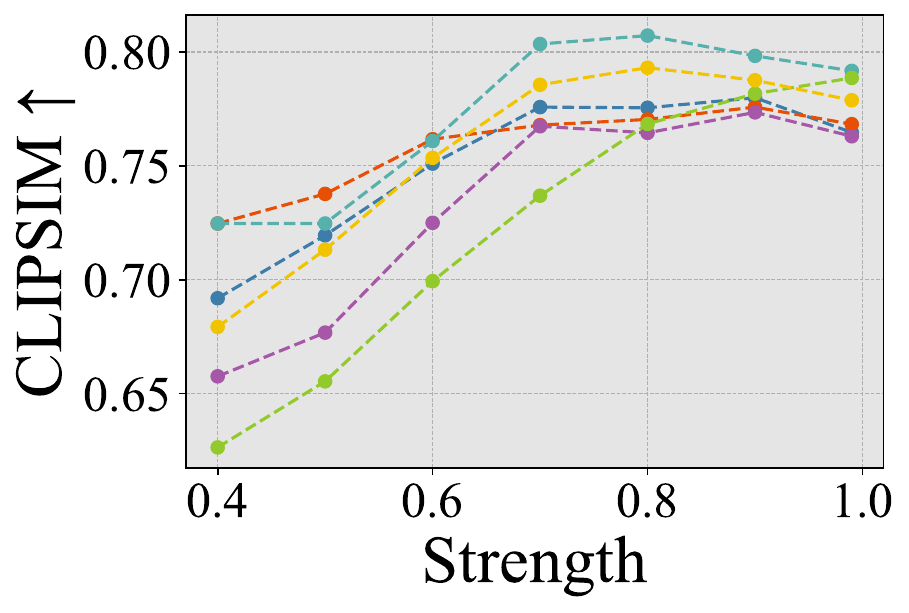}
    }\vspace{-1.5mm}
    \subfigure{\includegraphics[width=0.32\textwidth]{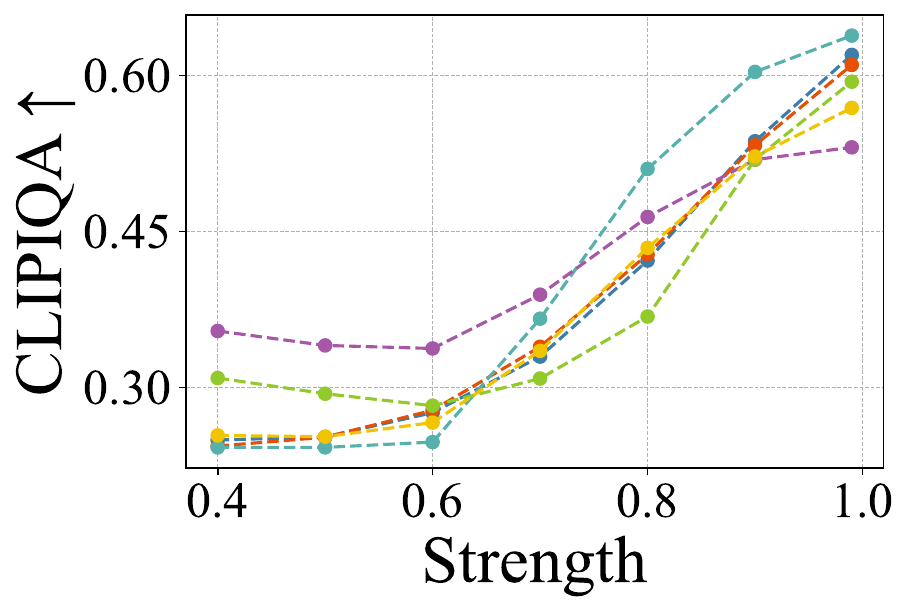}
    }\vspace{-1.5mm}
    \subfigure{\includegraphics[width=0.32\textwidth]{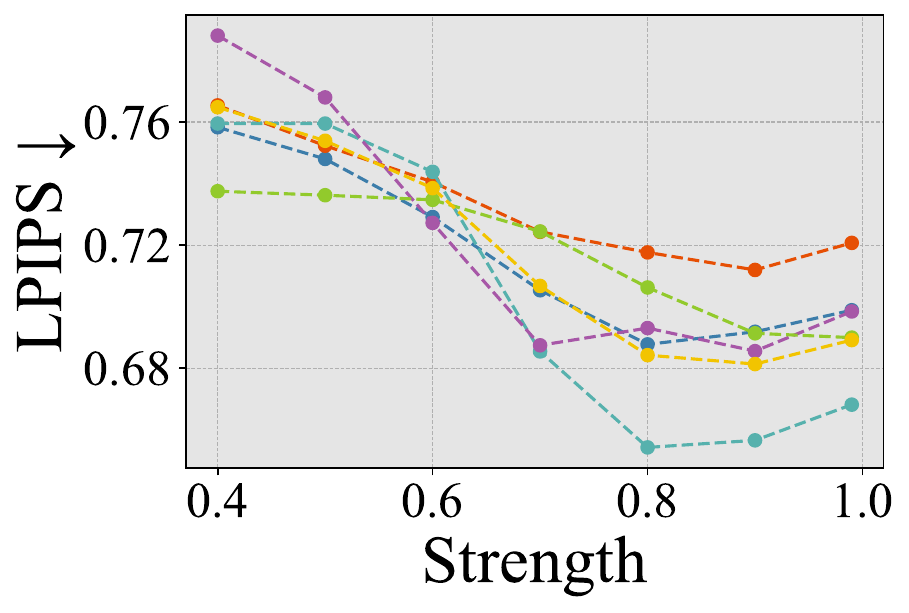}
    }\vspace{-1.5mm}
    \subfigure{\includegraphics[width=0.32\textwidth]{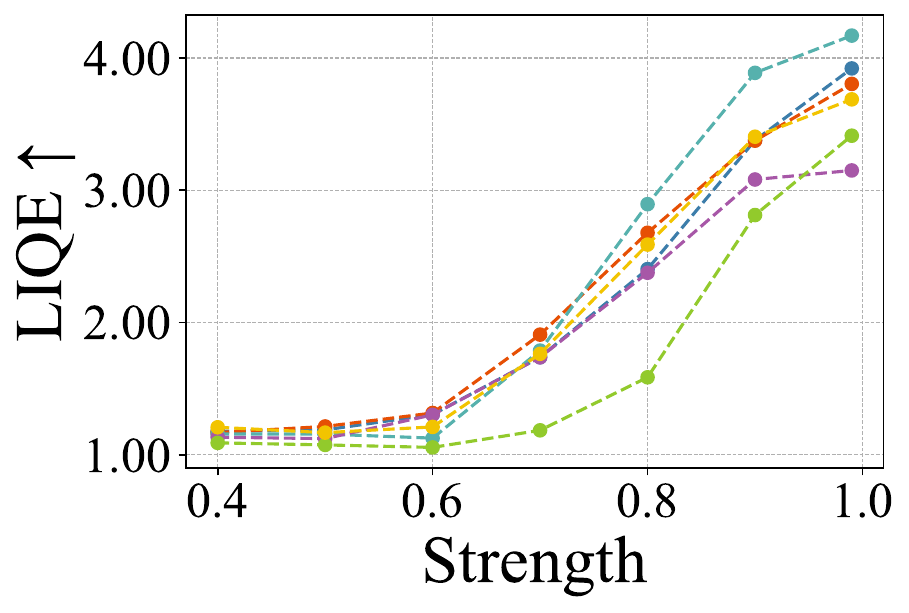}
    }\vspace{-1.5mm} 
    \subfigure{\includegraphics[width=0.32\textwidth]{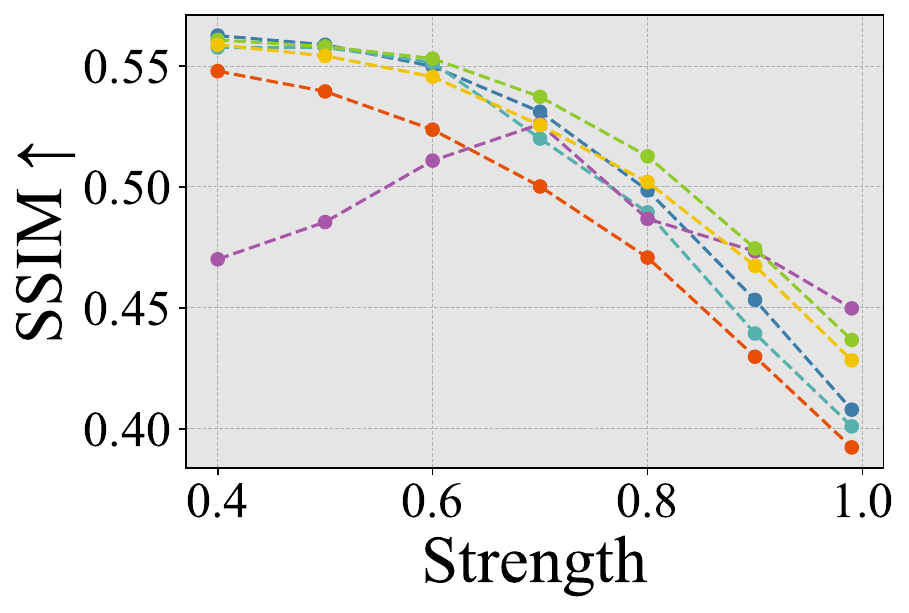}
    }\vspace{-1.5mm} 
    \subfigure{\includegraphics[width=0.32\textwidth]{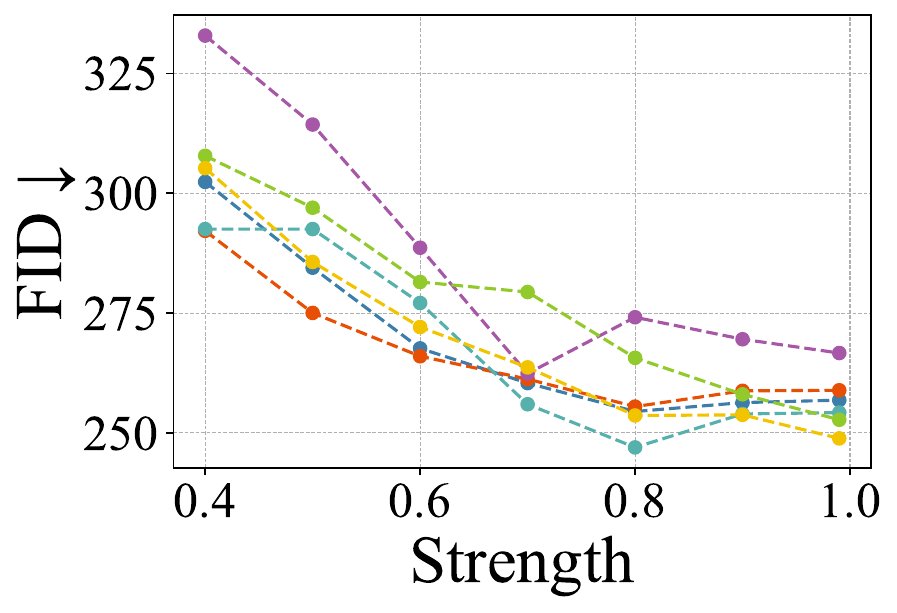}
    }\vspace{-1.5mm}  
    \subfigure{
    \centering
    \begin{minipage}{0.32\linewidth}
        \centering
        \vspace{-28mm}
        \hspace{4mm}
        \includegraphics[width=0.58\textwidth]{fig-supp/t2i-legend.pdf}
    \end{minipage}
    
    }
    \vspace{-8pt}
    \caption{Setting T2I model with different strength for CMC task in \textit{Pixel} mode, evaluated by 4
consistency and 4 perception metrics indicated by \colorbox{lightgray!60!white}{marked} and plain background. A strength of 0.8 can reach a balance between consistency and perception.}
    \label{fig:t2i-valid-pixel}
\end{figure}

The verification of \textit{Full}/\textit{Image}/\textit{Pixel} results are shown in Figure \ref{fig:t2i-valid-full}/\ref{fig:t2i-valid-image}/\ref{fig:t2i-valid-pixel} respectively, using same 4 Consistency and 4 Perception indicators. In general, as the strength increases, the Consistency index increases first and then decreases, while the Perception index continues to rise. This is because the greater the strength, the more details the T2I model adds to the image, thereby improving the Perception score. However, for Consistency, the added details at low strength can indeed make up for the unclear areas in the reference image, thereby performing restoration; but when the strength increases, the added details are inconsistent with the original image, and instead bring negative optimization to the reference image. Thus, a good strength requires a trade-off between Consistency and Perception. Taking both dimensions into consideration, we set the strength of \textit{Full} and \textit{Image} mode to 0.5 and the \textit{Pixel} mode to 0.8.

\subsection{Applicability on Different Reference Image}
\label{app:ref}

In the main text, the VVC provides the reference image with QP=53. In Table \ref{tab:ref} we further use VVC with QP=51,48,45 as the reference image for T2I model denoising to perform CMC. At higher bitrates, CMC still has an overwhelming advantage over VVC in the Perception metric. Except for SSIM, CMC achieves comprehensive optimization of all other indicators compared to traditional codecs, but the optimization range gradually decreases with the increase of bitrate. Moreover, once the QP is lower than 45, it will fall behind in the Consistency indicators. In summary, compared with traditional codecs, CMC can achieve an overall improvement in Perception and Consistency at low bitrates. However, when bpp increases to 0.1 or above, the improvement in Perception comes at the cost of Consistency. This indicates that ideal performance at higher bitrates is an important factor when using LMMs for image compression.

\begin{table}[t]
\centering
\caption{Changing reference image as different VVC QP level in \textit{Full} mode before CMC decoding, evaluated by 4
consistency and 4 perception metrics indicated by \colorbox{lightgray!60!white}{marked} and plain background.}
\label{tab:ref}
\adjustbox{max width=1.0\textwidth}{
\begin{tabular}{ll|>{\columncolor[HTML]{D0D0D0}}cc>{\columncolor[HTML]{D0D0D0}}cc>{\columncolor[HTML]{D0D0D0}}cc>{\columncolor[HTML]{D0D0D0}}cc}
\toprule
Reference                & Image    & CLIPSIM $\uparrow$ & CLIPIQA $\uparrow$ & LPIPS $\downarrow$  & LIQE $\uparrow$   & SSIM $\uparrow$   & FID $\downarrow$    & TOPIQ-FR $\uparrow$ & TOPIQ-NR $\uparrow$ \\ \midrule
\multirow{3}{*}{\begin{tabular}[c]{@{}l@{}}Extreme\\ (\textit{Full} mode)\end{tabular}} & Original & 0.825   & 0.181   & 0.523  & 1.221   & 0.677   & 300.8 & 2.546    & 1.217    \\
                         & CMC      & 0.890   & 0.670   & 0.435  & 3.530   & 0.571   & 255.8 & 2.964    & 2.553    \\
                         & Improve  & 7.88\%   & 270\% & 20.2\% & 189\% & -15.6\% & 17.5\%  & 16.4\%   & 109\%  \\ \hline
\multirow{3}{*}{QP51}    & Original & 0.874   & 0.185   & 0.445  & 1.454   & 0.722   & 286.7 & 2.886    & 1.294    \\
                         & CMC      & 0.919   & 0.696   & 0.383  & 3.749   & 0.586   & 255.2 & 3.202    & 2.810    \\
                         & Improve  & 5.15\%   & 276\% & 16.1\% & 157\% & -18.8\% & 12.3\%  & 10.9\%   & 117\%  \\ \hline
\multirow{3}{*}{QP48}    & Original & 0.912   & 0.214   & 0.372  & 1.905   & 0.764   & 281.1 & 3.241    & 1.403    \\
                         & CMC      & 0.931   & 0.704   & 0.350  & 3.884   & 0.603   & 254.2 & 3.342    & 2.981    \\
                         & Improve  & 2.08\%   & 228\% & 6.29\%  & 103\% & -21.0\% & 10.5\%  & 3.12\%    & 112\%  \\ \hline
\multirow{3}{*}{QP45}    & Original & 0.938   & 0.262   & 0.302  & 2.461   & 0.805   & 270.9 & 3.550    & 1.542    \\
                         & CMC      & 0.939   & 0.711   & 0.327  & 3.999   & 0.611   & 254.0 & 3.431    & 3.052    \\ 
                         & Improve  & 0.11\%   & 171\% & -7.65\% & 62.4\%  & -24.1\% & 6.67\%   & -3.35\%   & 97.9\%  \\ \bottomrule
\end{tabular}}
\end{table}

\subsection{Example Result Visualization}
\label{tab:refimg}
\begin{figure}[tbph]
    \centering
    \vspace{-2pt}
    \includegraphics[width=\linewidth]{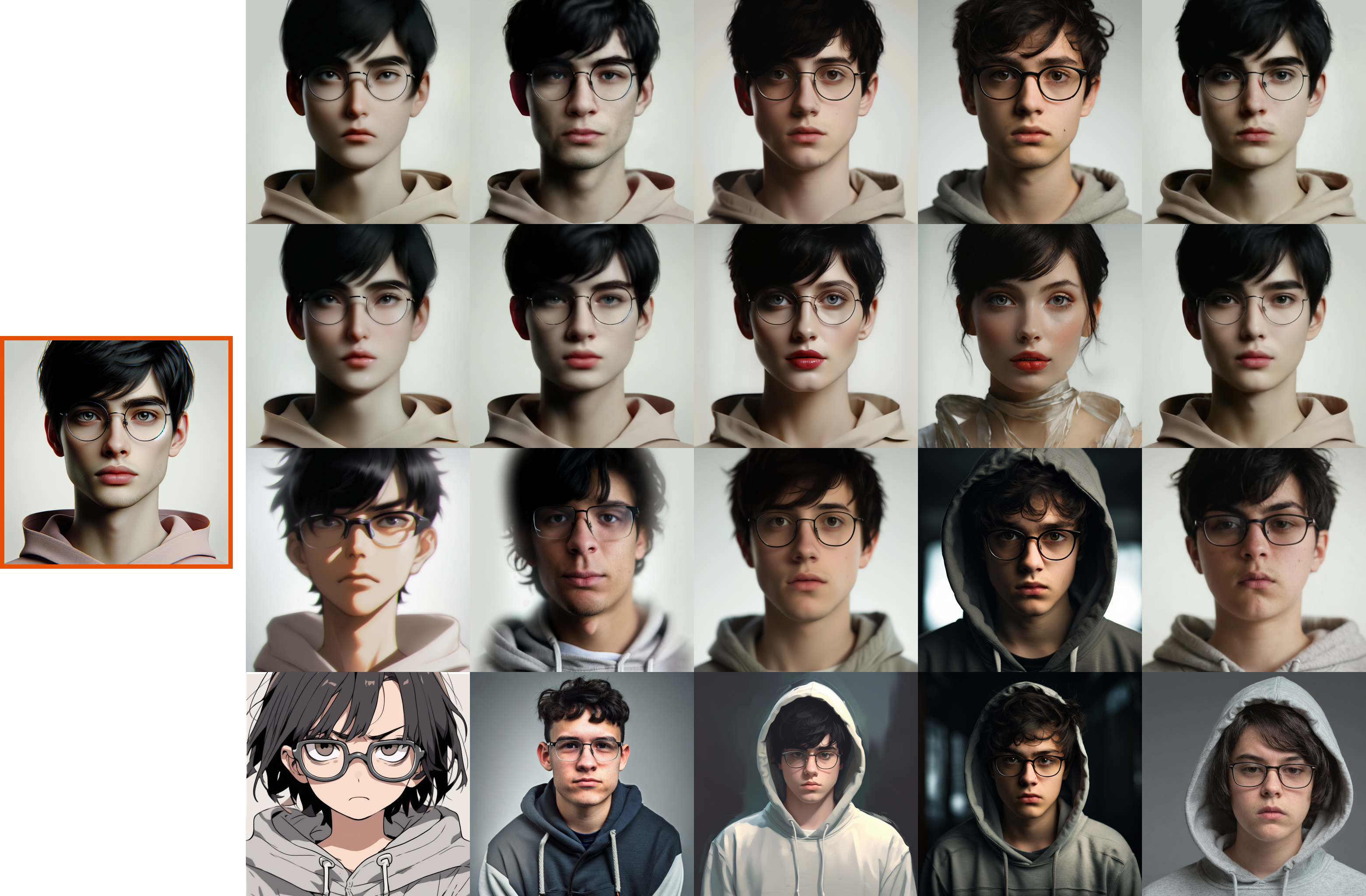}
    \vspace{-2pt}
    \caption{Visualization of an AIGI (Human) on the left after CMC. Row: \textit{Full}/\textit{Image}/\textit{Pixel}/\textit{Text} mode. Column: Animate\cite{gen:animatediff}/ Dreamlike\cite{gen:dream}/PG20\cite{gen:Playground20}/PG25\cite{gen:Playground25}/RealVis\cite{gen:RealVis} as decoder.}
    \label{fig:exp-aigi-0}
\end{figure}

\begin{figure}[tbph]
    \centering
    \vspace{-2pt}
    \includegraphics[width=\linewidth]{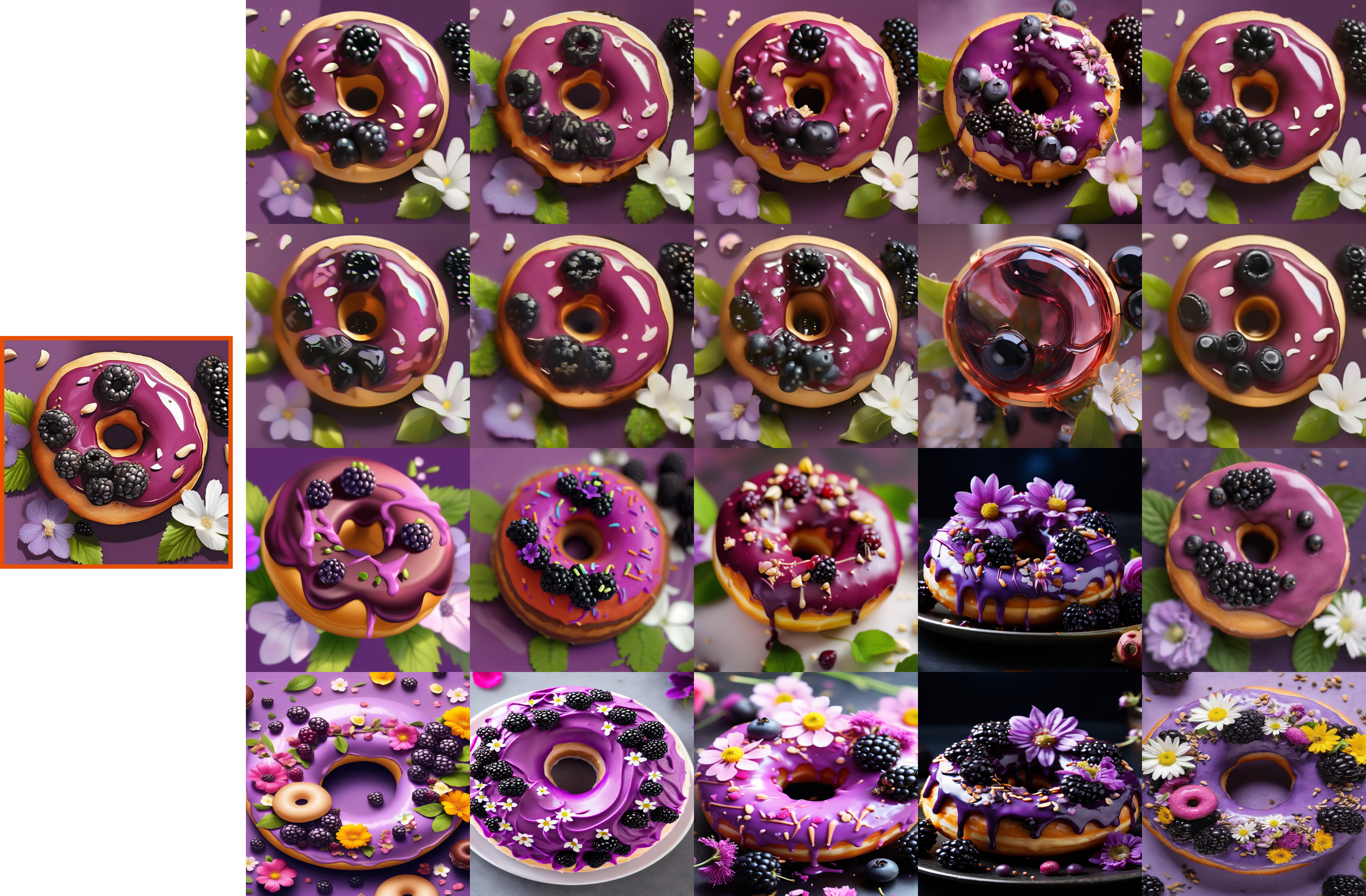}
    \vspace{-2pt}
    \caption{Visualization of an AIGI (Object) on the left after CMC. Row: \textit{Full}/\textit{Image}/\textit{Pixel}/\textit{Text} mode. Column: Animate\cite{gen:animatediff}/ Dreamlike\cite{gen:dream}/PG20\cite{gen:Playground20}/PG25\cite{gen:Playground25}/RealVis\cite{gen:RealVis} as decoder.}
    \label{fig:exp-aigi-1}
\end{figure}

\begin{figure}[tbph]
    \centering
    \vspace{-2pt}
    \includegraphics[width=\linewidth]{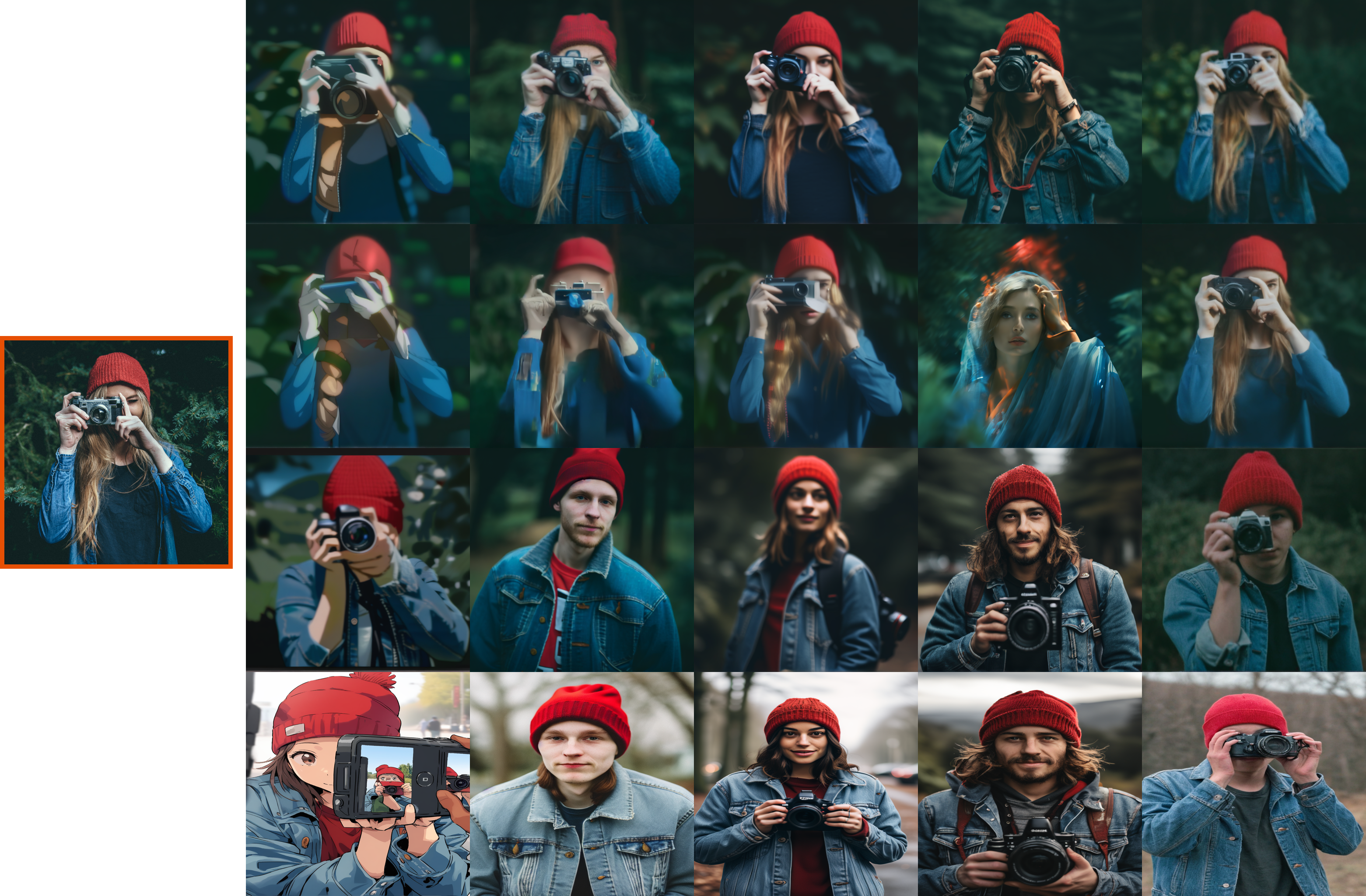}
    \vspace{-2pt}
    \caption{Visualization of an NSI (PGC) on the left after CMC. Row: \textit{Full}/\textit{Image}/\textit{Pixel}/\textit{Text} mode. Column: Animate\cite{gen:animatediff}/ Dreamlike\cite{gen:dream}/PG20\cite{gen:Playground20}/PG25\cite{gen:Playground25}/RealVis\cite{gen:RealVis} as decoder.}
    \label{fig:exp-nsi-0}
\end{figure}

\begin{figure}[tbph]
    \centering
    \vspace{-2pt}
    \includegraphics[width=\linewidth]{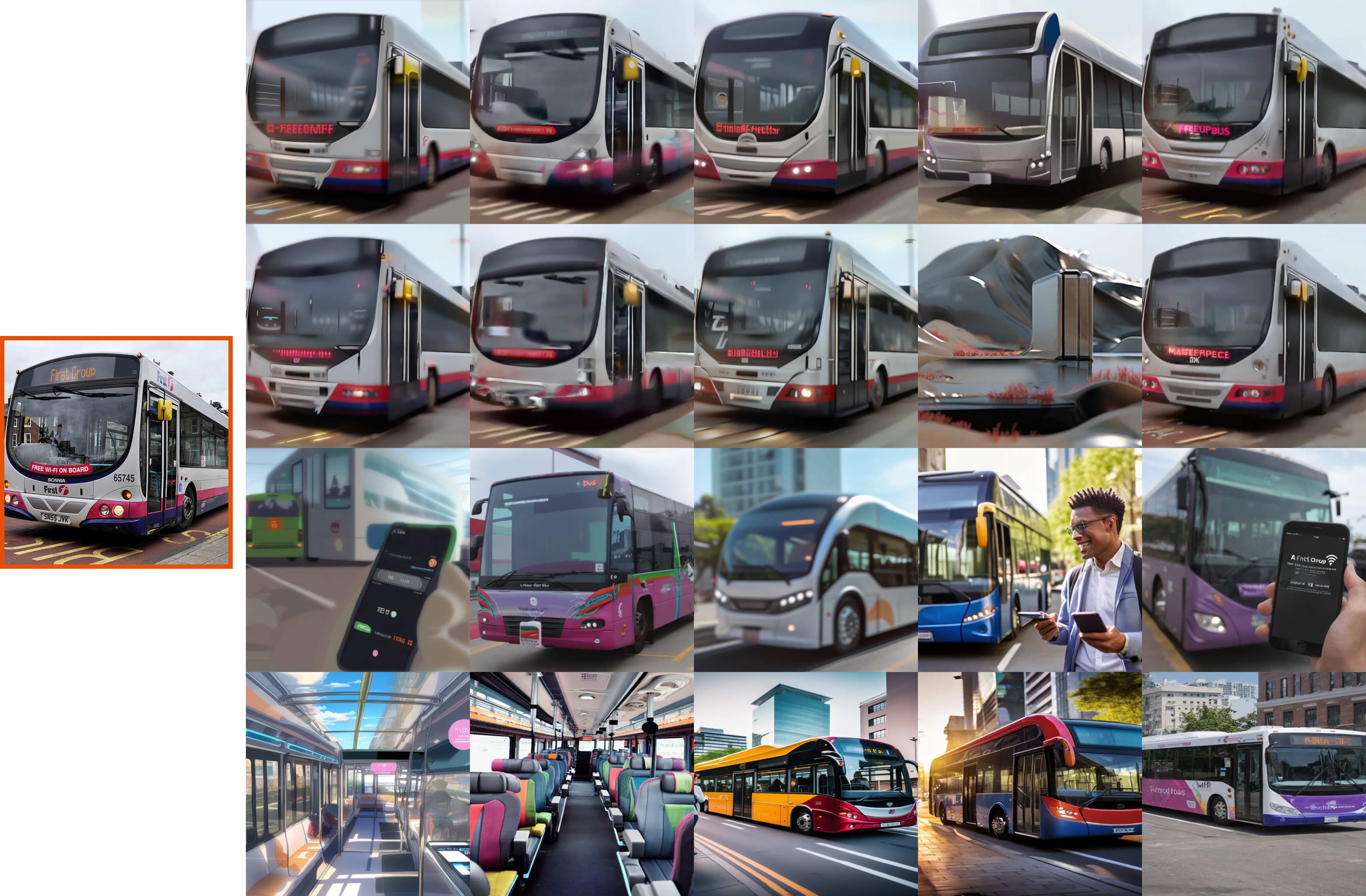}
    \vspace{-2pt}
    \caption{Visualization of an NGI (UGC) on the left after CMC. Row: \textit{Full}/\textit{Image}/\textit{Pixel}/\textit{Text} mode. Column: Animate\cite{gen:animatediff}/ Dreamlike\cite{gen:dream}/PG20\cite{gen:Playground20}/PG25\cite{gen:Playground25}/RealVis\cite{gen:RealVis} as decoder.}
    \label{fig:exp-nsi-1}
\end{figure}

\begin{figure}[tbph]
    \centering
    \vspace{-2pt}
    \includegraphics[width=\linewidth]{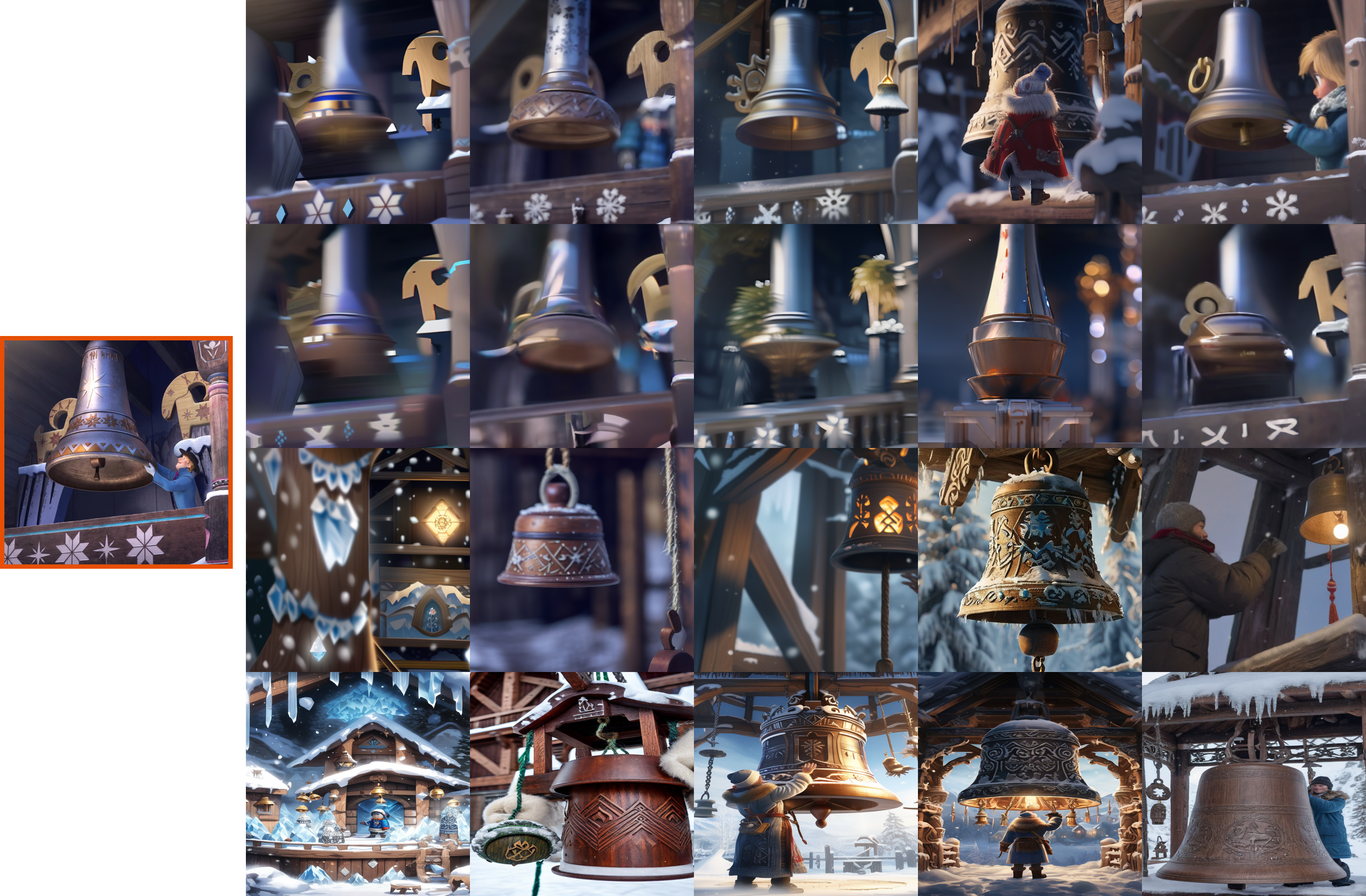}
    \vspace{-2pt}
    \caption{Visualization of an SCI (Movie) on the left after CMC. Row: \textit{Full}/\textit{Image}/\textit{Pixel}/\textit{Text} mode. Column: Animate\cite{gen:animatediff}/ Dreamlike\cite{gen:dream}/PG20\cite{gen:Playground20}/PG25\cite{gen:Playground25}/RealVis\cite{gen:RealVis} as decoder.}
    \label{fig:exp-sci-0}
\end{figure}

\begin{figure}[tbph]
    \centering
    \vspace{-2pt}
    \includegraphics[width=\linewidth]{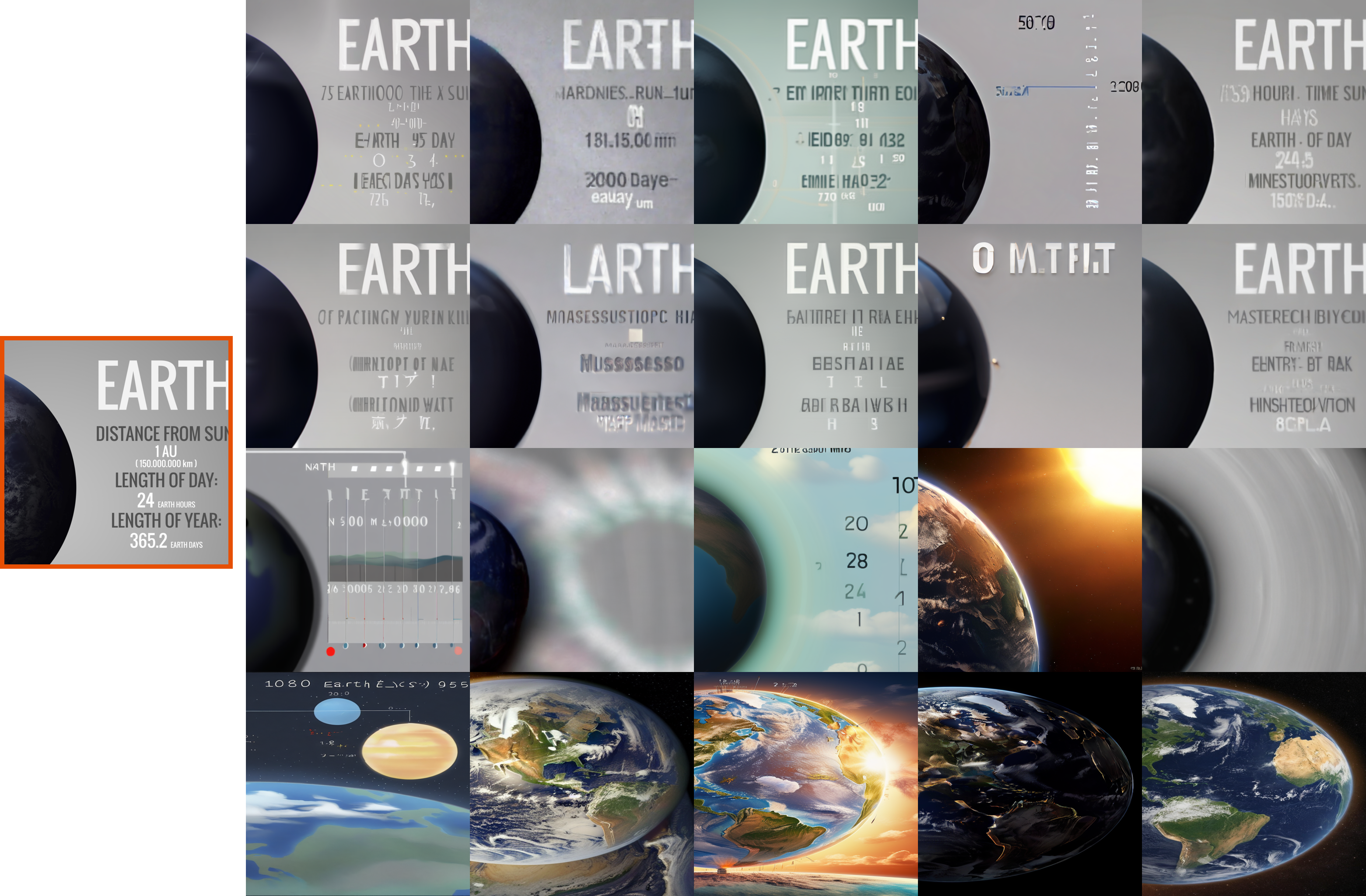}
    \vspace{-2pt}
    \caption{Visualization of an SCI (Webpage) on the left after CMC. Row: \textit{Full}/\textit{Image}/\textit{Pixel}/\textit{Text} mode. Column: Animate\cite{gen:animatediff}/ Dreamlike\cite{gen:dream}/PG20\cite{gen:Playground20}/PG25\cite{gen:Playground25}/RealVis\cite{gen:RealVis} as decoder.}
    \label{fig:exp-sci-1}
\end{figure}

The CMC result visualization is shown from Figure \ref{fig:exp-aigi-0} to \ref{fig:exp-sci-1}, all result use GPT-4o \cite{i2t:gpt4} as encoder, and Animate\cite{gen:animatediff}/ Dreamlike\cite{gen:dream}/PG20\cite{gen:Playground20}/PG25\cite{gen:Playground25}/RealVis\cite{gen:RealVis} as decoder (from left to right). Four working modes \textit{Full}/\textit{Image}/\textit{Pixel}/\textit{Text} are all included (from top to bottom). For different modes, the compression results from LMMs show that as the bitrate decreases, the decoded image is more different from the ground truth. Among them, the \textit{Full} mode can obtain results generally similar to the ground truth; the \textit{Image} mode will lose some semantic details while introducing artifacts; the \textit{Pixel} mode loses more details but ensures the consistency of the overall composition; and the result generated by \textit{Text} is significantly different from the ground truth. 

For the performance of the CMC on different contents, Figure \ref{fig:exp-aigi-0}/\ref{fig:exp-aigi-1} reveals it performs most satisfactorily on AIGIs; Figure \ref{fig:exp-nsi-0}/\ref{fig:exp-nsi-1} indicates it can also obtain results consistent with ground truth on NSIs, but it is easy to lose details such as human faces and vehicle signs; Figure \ref{fig:exp-sci-0}/\ref{fig:exp-sci-1} implies it is the least ideal on SCIs, as it misunderstands the relationship between characters in the movie or games, and cannot draw formed letters on webpages. In conclusion, CMC is a promising visual signal compression method, but to become a universal codec standard in the future, the robustness to all content types needs to be improved.

\subsection{Data Statement}
\label{app:statement}

The CMC-Bench dataset is released under the \textbf{CC BY 4.0} license. This includes all ground truth, distorted images, subjective annotations, and the weight of the Consistency/Perception evaluation model. All LMM developers can test their performance through our public scripts, and all image compression researchers can obtain the public I2T+T2I LMM pipeline. We believe these resources can inspire the next generation of visual signal codec protocols.

\end{document}